\documentclass{clv3}

\usepackage{hyperref}
\usepackage{xcolor}
\definecolor{darkblue}{rgb}{0, 0, 0.5}
\hypersetup{colorlinks=true,citecolor=darkblue, linkcolor=darkblue, urlcolor=darkblue}

\usepackage{color}
\usepackage{colortbl}

\usepackage{xcolor}
\newcommand{\myRed}[1]{\textcolor{black}{#1}}
\usepackage{microtype}

\usepackage{comment}
\usepackage{rotating}

\usepackage{multirow}
\usepackage{multicol}
\usepackage{graphicx}
\usepackage{booktabs}
\usepackage{makecell}
\usepackage{enumitem}
\usepackage{setspace}
\usepackage{longtable}
\usepackage{pdflscape}
\usepackage{tabu}
\usepackage{soul}
\usepackage{amsmath}

\usepackage{caption}
\usepackage{subcaption}
\graphicspath{ {figs/} }

\usepackage{sec/package}

\historydates{Submission received: 25 April 2021, Revised version received: 30 August 2021, Accepted for publication: 4 December 2021.}

\issue{xx}{yy}{2021}

\dochead{}

\runningtitle{Deep Learning for Text Style Transfer: A Survey}

\runningauthor{Jin, Jin, Hu, Vechtomova and
Mihalcea}

\begin{document}

\title{Deep Learning for Text Style Transfer: \\ A Survey}

\author{Di Jin\thanks{Equal contribution.}}
\affil{MIT CSAIL \\
\texttt{jindi15@mit.edu}}

\author{Zhijing Jin\textsuperscript{*}}
\affil{Max Planck Institute \& ETH Zürich\\
\texttt{zjin@tue.mpg.de}}

\author{Zhiting Hu}
\affil{UC San Diego \\
  \texttt{zhh019@ucsd.edu}}

\author{Olga Vechtomova}
\affil{University of Waterloo \\
  \texttt{ovechtom@uwaterloo.ca}}
\author{Rada Mihalcea}
\affil{University of Michigan \\
  \texttt{mihalcea@umich.edu}}

\maketitle

\begin{abstract}
Text style transfer is an important task in natural language generation, which aims to control certain attributes in the generated text, such as politeness, emotion, humor, and many others. It has a long history in the field of natural language processing, and recently has re-gained significant attention thanks to the promising performance brought by deep neural models. In this paper, we present a systematic survey of the research on neural text style transfer, spanning over 100 representative articles since the first neural text style transfer work in 2017. We discuss the task formulation, existing datasets and subtasks, evaluation, as well as the rich methodologies in the presence of parallel and non-parallel data. We also provide discussions on a variety of important topics regarding the future development of this task.\footnote{Our curated paper list is at \url{https://github.com/zhijing-jin/Text_Style_Transfer_Survey}.}
\end{abstract}

\section{Introduction}
\label{sec:intro}

Language is situational. Every utterance fits in a specific time, place, and scenario, conveys specific characteristics of the speaker, and typically has a well-defined intent. 
For example, someone who is uncertain is more likely to use tag questions (e.g., “This is true, isn’t it?”) than declarative sentences (e.g., “This is definitely true.”). Similarly, a professional setting is more likely to include formal statements (e.g., “Please consider taking a seat.”) as compared to an informal situation (e.g., “Come and sit!”). For artificial intelligence systems to accurately understand and generate language, it is necessary to model language with style/attribute,\footnote{Note that we interchangeably use the terms \textit{style} and \textit{attribute} in this survey. \textit{Attribute} is a broader terminology that can include content preferences, e.g., sentiment, topic, and so on. This survey uses \textit{style} in the same broad way, following the common practice in recent papers (see Section~\ref{sec:define}).} which goes beyond merely verbalizing the semantics in a non-stylized way.
The values of the attributes can be drawn from a wide range of choices depending on pragmatics, such as the extent of formality, politeness, simplicity, personality, emotion,
partner effect (e.g., reader awareness), genre of writing (e.g., fiction or non-fiction), and so on. 

The goal of text style transfer (TST) is  to automatically control the style attributes of text while preserving the content. TST has a wide range of applications, as outlined by \citet{mcdonald1985computational} and \citet{hovy1987generating}. The style of language is crucial because it makes natural language processing more user-centered. TST has many immediate applications. For instance, one such application is intelligent bots for which users prefer distinct and consistent persona (e.g., empathetic) instead of emotionless or inconsistent persona.
Another application is the development of intelligent writing assistants; for example, non-expert writers often need to polish their writings to better fit their purpose, e.g., more professional, polite, objective, humorous, or other advanced writing requirements, which may take years of experience to master. Other applications include automatic text simplification (where the target style is “simple”), debiasing online text (where the target style is “objective”), fighting against offensive language (where the target style is “non-offensive”), and so on.

To formally define text style transfer, let us denote the target utterance as $\bm{x}'$ and the target discourse style attribute as $a'$. TST aims to model $p(\bm{x}'|a,\bm{x})$, where $\bm{x}$ is a given text carrying a source attribute value $a$. Consider the previous example of text expressed by two different extents of formality:
\begin{table}[ht]
    \begin{tabular}{llll}
    Source sentence $\bm{x}$: & ``\textit{Come and sit!}'' & Source attribute $a$: & \textit{Informal}
    \\
    Target sentence $\bm{x}'$: & ``\textit{Please consider taking a seat.}'' & Target attribute $a'$: & \textit{Formal}
    \\
    \end{tabular}
\end{table}

In this case, a TST model should be able to modify the formality and generate the formal sentence $\bm{x}'=$``\textit{Please consider taking a seat.}'' given the informal input $\bm{x}=$``\textit{Come and sit!}''. Note that the key difference of TST from another NLP task, style-conditioned language modeling, is that the latter is conditioned on only a style token, whereas TST takes as input both the target style attribute $a'$ and a source sentence $\bm{x}$ that constrains the content.


Crucial to the definition of style transfer is the distinction of ``style'' and ``content,'' for which there are two common practices. The first one is by linguistic definition, where non-functional linguistic features are classified into the style (e.g., formality), and the semantics are classified into the content. In contrast, the second practice is data-driven, -- given two corpora (e.g., a positive review set and a negative review set), the invariance between the two corpora is the content, whereas the variance is the style (e.g., sentiment, topic) \cite{mou-vechtomova-2020-stylized}. 

Driven by the growing needs for TST, active research in this field has emerged, from the traditional linguistic approaches, to the more recent neural network-based approaches. Traditional approaches rely on term replacement and templates. For example, early work in NLG for weather forecasts builds domain-specific templates to \myRed{express different types of weather with different levels of uncertainty for different users \cite{sripada2004lessons,reiter2005choosing,belz2008automatic,gkatzia2017data}}. Research that more explicitly focuses on TST starts from the frame language-based systems \cite{mcdonald1985computational}, and schema-based NLG systems \cite{hovy1987generating,hovy1990pragmatics} which generate text with pragmatic constraints such as formality under small-scale well-defined schema. Most of this earlier work required domain-specific templates, hand-featured phrase sets that express a certain attribute (e.g., friendly), and sometimes a look-up table of expressions with the same meaning but multiple different attributes \cite{bateman1989phrasing,stamatatos1997user,power2003generating,reiter2003lessons,sheikha2011formal,mairesse2011controling}.

With the success of deep learning in the last decade, a variety of neural methods have been recently proposed for TST. If parallel data are provided, standard sequence-to-sequence models are often directly applied~\cite{rao-tetreault-2018-dear} (see Section~\ref{sec:sup}). However, most use cases do not have parallel data, so TST on non-parallel corpora has become a prolific research area (see Section~\ref{sec:unsup}).
The first line of approaches \textit{disentangle} text into its content and attribute in the latent space, and apply generative modeling \cite{Hu2017TowardCG,shen2017style}.
This trend was then joined by another distinctive line of approach, \textit{prototype editing} \cite{li-etal-2018-delete} which extracts a sentence template and its attribute markers to generate the text. Another paradigm soon followed, i.e., \textit{pseudo-parallel corpus construction} to train the model as if in a supervised way with the pseudo-parallel data \cite{zhang2018style,jin2019imat}. These three directions, (1) disentanglement, (2) prototype editing, and (3) pseudo-parallel corpus construction, are further advanced with the emergence of Transformer-based models \cite{Sudhakar2019TransformingDR,malmi2020unsupervised}.

Given the advances in TST methodologies, it now starts to expand its impact to downstream applications, such as persona-based dialog generation \cite{niu-bansal-2018-polite,huang-etal-2018-automatic}, stylistic summarization \cite{jin-etal-2020-hooks}, stylized language modeling to imitate specific authors \cite{Syed2020AdaptingLM}, online text debiasing \cite{pryzant2020automatically,ma2020powertransformer}, simile generation \cite{chakrabarty2020generating}, and many others.

\begin{table}[ht]
    \small
    \caption{Overview of the survey.}
    \label{tab:overview}
    \centering
    \setlength\tabcolsep{0pt}
    \setlength\extrarowheight{2pt}
    \begin{tabular*}{\columnwidth}{p{0.13\columnwidth}@{\extracolsep{\fill}}p{0.08\columnwidth}@{\extracolsep{\fill}}p{0.12\columnwidth}@{\extracolsep{\fill}}p{0.21\columnwidth}@{\extracolsep{\fill}}p{0.21\columnwidth}}
    \toprule
    \multicolumn{1}{c}{\textbf{Motivation}} & \multicolumn{2}{c}{\textbf{Data}} & \multicolumn{1}{c}{\textbf{Method}} & \multicolumn{1}{c}{\textbf{Extended Applications}} \\ 
    \cline{1-1} 
    \cline{2-3}
    \cline{4-4}
    \cline{5-5}
    \begin{minipage}[t]{1.3\linewidth}
    \begingroup
    \fontsize{7.5pt}{9pt}\selectfont
    \begin{itemize}[nosep, wide=0pt, leftmargin=*, after=\strut]
        \item Artistic writing
        \item Communication
        \item Mitigating social issues
    \end{itemize}
    \endgroup
    \end{minipage}
    
    &
    \begin{minipage}[t]{1.5\linewidth}
    \begingroup
    \fontsize{7.5pt}{9pt}\selectfont
    \textbf{Tasks}
    \begin{itemize}[nosep, wide=0pt, leftmargin=*, after=\strut]
        \item Formality
        \item Politeness
        \item Gender
        \item Humor
        \item Romance
        \item Biasedness
    \end{itemize}
    \endgroup
    \end{minipage}
    
    \begin{minipage}[t]{4\linewidth}
    \begingroup
    \fontsize{7.5pt}{9pt}\selectfont
    \textbf{Key Properties}
    \begin{itemize}[nosep, wide=0pt, leftmargin=*, after=\strut]
        \item Parallel vs. non-parallel
        \item Uni- vs. bi-directional
        \item Dataset size
        \item Large vs. small word overlap
    \end{itemize}
    \endgroup
    \end{minipage}
    &
    
    \begin{minipage}[t]{1.5\linewidth}
    \begingroup
    \fontsize{7.5pt}{9pt}\selectfont
    { \quad } 
    \begin{itemize}[nosep, wide=0pt, leftmargin=*, after=\strut]
        \item Toxicity
        \item Authorship
        \item Simplicity
        \item Sentiment
        \item Topic
        \item Political slant
    \end{itemize}
    \endgroup
    \end{minipage}
    &
    \begin{minipage}[t]{1.5\linewidth}
    \begingroup
    \fontsize{7.5pt}{9pt}\selectfont
    \textbf{On Parallel Data}
    \begin{itemize}[nosep, wide=0pt, leftmargin=*, after=\strut]
        \item Multi-tasking
        \item Inference techniques
        \item Data augmentation
    \end{itemize}
    \endgroup
    
    \begingroup
    \fontsize{7.5pt}{9pt}\selectfont
    { \quad }
    \newline
    \textbf{On Non-Parallel Data}
    \begin{itemize}[nosep, wide=0pt, leftmargin=*, after=\strut]
        \item Disentanglement
        \item Prototype editing
        \item Pseudo data construction
    \end{itemize}
    \endgroup
    \end{minipage}
    & 
    \begin{minipage}[t]{1.2\linewidth}
    \begingroup
    \fontsize{7.5pt}{9pt}\selectfont
    \textbf{Helping Other NLP Tasks}
    \endgroup
    \begingroup
    \fontsize{7.5pt}{9pt}\selectfont
    \begin{itemize}[nosep, wide=0pt, leftmargin=*, after=\strut]
        \item Paraphrasing
        \item Data augmentation
        \item Adversarial robustness
        \item Persona-consistent dialog
        \item \myRed{Anonymization}
        \item Summarization
        \item Style-specific MT
    \end{itemize}
    \endgroup
    \end{minipage}
    \\
    \bottomrule
\end{tabular*}
\end{table}

\paragraph{Motivation of a Survey on TST.}
The increasing interest in modeling the style of text can be regarded as a trend reflecting the fact that NLP researchers start to focus more on user-centeredness and personalization. However, despite the growing interest in TST, the existing literature shows a large diversity in the selection of benchmark datasets, methodological frameworks, and evaluation metrics. Thus, the aim of this survey is to provide summaries and potential standardizations on some important aspects of TST, such as the terminology, problem definition, benchmark datasets, and evaluation metrics.
We also aim to provide different perspectives on the methodology of TST, and suggest some potential cross-cutting research questions for our proposed research agenda of the field.
As shown in Table~\ref{tab:overview}, the key contributions targeted by this survey are as follows:

\begin{enumerate}
    \item We conduct the first comprehensive review that covers most existing works (more than 100 papers) on deep learning-based TST.
    \item We provide an overview of the task setting, terminology definition, benchmark datasets (Section~\ref{sec:task-overview}), and evaluation metrics for which we proposed standard practices that can be helpful for future works (Section~\ref{sec:eval}).
    \item We categorize the existing approaches on parallel data (Section~\ref{sec:sup}) and non-parallel data (Section~\ref{sec:unsup}) for which we distill some unified methodological frameworks.
    \item We discuss a potential research agenda for TST (Section~\ref{sec:discussion}), including expanding the scope of styles, improving the methodology, loosening dataset assumptions, and improving evaluation metrics.
    \item We provide a vision for how to broaden the impact of TST (Section~\ref{sec:impact}), including connecting to more NLP tasks, and more specialized downstream applications, as well as considering some important ethical impacts.
\end{enumerate}

\paragraph{Paper Selection.} The neural TST papers reviewed in this survey are mainly from top conferences in NLP and artificial intelligence (AI), including ACL, EMNLP, NAACL, COLING, CoNLL, NeurIPS, ICML, ICLR, AAAI, and IJCAI. Other than conference papers, we also include some non-peer-reviewed preprint papers that can offer some insightful information about the field. The major factors for selecting non-peer-reviewed preprint papers include novelty and completeness, among others. 


\section{What Is Text Style Transfer?}
\label{sec:task-overview}

This section  provides an overview of the style transfer task. Section~\ref{sec:define}  goes through the definition of styles and the scope of this survey. Section~\ref{sec:formulation}  gives a task formulation and introduces the notations that will be used across the survey. Finally, Section~\ref{sec:subtasks} lists all the common subtasks for neural text style transfer which can save the literature review efforts for future researchers.

\subsection{How to Define Style?}\label{sec:define}
\paragraph{\myRed{Linguistic Definition of Style.}}
An intuitive notion of style refers to the manner in which the semantics is expressed \cite{mcdonald1985computational}. Just as everyone has their own signatures, style originates as the characteristics inherent to every person's utterance, which can be expressed through the use of certain stylistic devices such as metaphors, as well as choice of words, syntactic structures, and so on. Style can also go beyond the sentence level to the discourse level, such as the stylistic structure of the entire piece of the work, e.g., stream of consciousness, or flashbacks.

Beyond the intrinsic personal styles, for pragmatic uses, style further becomes a protocol to regularize the manner of communication. For example, for academic writing, the protocol requires formality and professionalism. \citet{hovy1987generating} defines style by its pragmatic aspects, including both personal (e.g., personality, gender) and interpersonal (e.g., humor, romance) aspects. Most existing literature also takes these well-defined categories of styles.

\begin{figure}[t]
    \centering
    \includegraphics[width=0.9\textwidth]{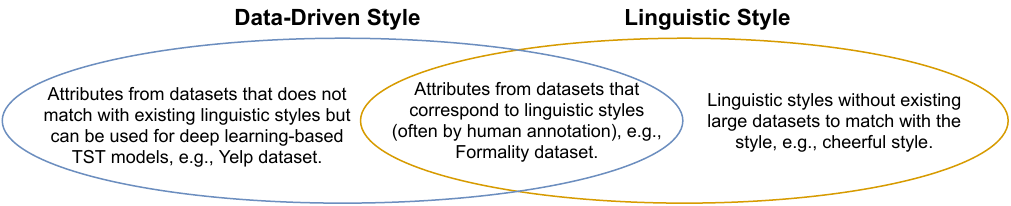}
    \caption{\myRed{Venn diagram of the linguistic definition of style and data-driven definition of style.}}
    \label{fig:defi}
\end{figure}
\paragraph{\myRed{Data-Driven Definition of Style as the Scope of this Survey.}}\label{sec:assumptions}
This survey aims to provide an overview of existing neural text style transfer approaches. To be concise, we will limit the scope to the most common settings of existing literature. Specifically, most deep learning work on TST adopts a \textbf{data-driven definition of style}, and the scope of this survey covers the styles in currently available TST datasets.
The data-driven definition of style is different from the linguistic or rule-based definition of style, which theoretically constrains what constitutes a style and what not, such as a style guide \citep[e.g.,][]{american1983publication} that requires that formal text not include any contraction, e.g., ``isn't.'' \myRed{The distinction of the two defintions of style is shown in Figure~\ref{fig:defi}.}

\myRed{With the rise of deep learning methods of TST, the data-driven definition of style extends the linguistic style to a broader concept -- the general attributes in text. It regards ``style'' as the attributes that vary across datasets, as opposed to the characteristics that stay invariant \cite{mou-vechtomova-2020-stylized}. The reason is that deep learning models, which are the focus of this survey, need large corpora to learn the style from, but not all styles have well-matched large corpora. 
Therefore, apart from the very few manually-annotated
datasets with linguistic style definitions, such as
formality \cite{rao-tetreault-2018-dear} and humor \& romance \cite{Gan2017StyleNetGA},
many recent dataset collection works automatically look for meta-information to link a corpus to a certain attribute. A typical example is the widely used Yelp review dataset \cite{shen2017style}, where reviews with low ratings are put into the negative corpus, and reviews with high ratings are put into the positive corpus, although the negative vs. positive opinion is not a style that belongs to the linguistic definition, but more of a content-related attribute.}


Most methods mentioned in this survey can be applied to scenarios that follow this data-driven definition of style. As a double-sided sword, the prerequisite for most methods is that there \textit{exist} style-specific corpora for each style of interest, either parallel or non-parallel. Note that there can be future works that do not take such an assumption, which will be discussed in Section~\ref{sec:loosening}.

\paragraph{Comparison of the Two Definitions.}
\myRed{There are two phenomena rising from the data-driven definition of style as opposed to the linguistic style. One is that the data-driven definition of style can include a broader range of attributes including content and topic preferences of the text. The other is that data-driven styles, if collected through automatic classification by meta-information such as ratings, user information, and source of text, can be more ambiguous than the linguistically defined styles. As shown in \citet[Section 4.1.1]{jin2019imat}, some automatically collected datasets have a concerningly high undecideable rate and inter-annotator disagreement rate when the annotators are asked to associate the dataset with human-defined styles such as political slant and gender-specific tones.} 

\myRed{The advantage of the data-driven style is that it can marry well with deep learning methods because most neural models learn the concept of style by learning to distinguish the multiple style corpora. For the (non-data-driven) linguistic style, although it is under-explored in the existing deep learning works of TST, we provide in Section~\ref{sec:loosening} a discussion of how potential future works can learn TST of linguistics styles with no matched data.}

\subsection{Task Formulation}\label{sec:formulation}
We define the main notations used in this survey in Table~\ref{tab:notation}.
\begin{table}[h]
    \small 
    \caption{Notation of each variable and its corresponding meaning.}
    \label{tab:notation}
    \centering
    \begin{tabular}{lcl}
    \toprule
    \textbf{Category} & \textbf{Notation} & \textbf{Meaning} \\ \midrule
    \multirow{4}{*}{Attribute} & $a$ & An attribute value, e.g., the formal style \\
    & ${a}'$ & An attribute value different from $a$ \\
    & $\mathbb{A}$ & A predefined set of attribute values \\
    & $a_i$ & The $i$-th attribute value in $\mathbb{A}$ \\ 
    \hline
    \multirow{5}{*}{Sentence} & $\bm{x}$ & A sentence with attribute value $a$\\
    & $\bm{x}'$ & A sentence with attribute value ${a}'$ \\
    & $\bm{X}_i$ & A corpus of sentences with attribute value $a_i$ \\
    & $\bm{x}_i$ & A sentence from the corpus $\bm{X}_i$ \\ 
    & $\widehat{\bm{x}'}$ & Attribute-transferred sentence of $\bm{x}$ learned by the model \\
    \hline
    \multirow{6}{*}{Model} & $E$ & Encoder of a TST model \\
    & $G$ & Generator of a TST model \\
    & $f_c$ & Attribute classifier \\
    & $\bm{\theta}_{\mathrm{E}}$ & Parameters of the encoder \\
    & $\bm{\theta}_{\mathrm{G}}$ & Parameters of the generator \\
    & $\bm{\theta}_{f_c}$ & Parameters of the attribute classifier \\
    \hline
    \multirow{2}{*}{Embedding} & $\bm{z}$ & Latent representation of text, i.e., $\bm{z} \overset{\Delta}{=} E(\bm{x})$ \\
    & $\bm{a}$ & Latent representation of the attribute value in text \\
    \bottomrule
    \end{tabular}
\end{table}

As mentioned previously in Section~\ref{sec:assumptions}, most neural approaches assume a given set of attribute values $\mathbb{A}$, and each attribute value has its own corpus. For example, if the task is about formality transfer, then for the attribute of text formality, there are two attribute values, $a$ = ``formal'' and $a'$ = ``informal,'' corresponding to a corpus $\bm{X}_1$ of formal sentences and another corpus $\bm{X}_2$ of informal sentences. The style corpora can be parallel or non-parallel. Parallel data means that each sentence with the attribute $a$ is paired with a counterpart sentence with another attribute ${a}'$. In contrast, non-parallel data only assumes mono-style corpora.


\subsection{Existing Subtasks with Datasets}\label{sec:subtasks}

\begin{table*}[h]
\small
\caption{List of common subtasks of TST and their corresponding attribute values and datasets. For datasets with multiple attribute-specific corpora, we report their sizes by the number of sentences of the smallest of all corpora. We also report whether the dataset is parallel (\textit{Pa?}).}
  \label{tab:task_type}
  \centering
  \begingroup
  \fontsize{7.5pt}{9pt}\selectfont
    
    \begin{tabular}{p{1.4cm} p{3cm} p{5.8cm} >{\centering\arraybackslash}p{0.6cm} >{\centering\arraybackslash}p{0.2cm}}
    \toprule
     \multicolumn{1}{l}{\textbf{Task}} & \multicolumn{1}{l}{\textbf{Attribute Values}} & \multicolumn{1}{l}{\textbf{Datasets}} &  \multicolumn{1}{c}{\textbf{Size}} & \multicolumn{1}{l}{\textbf{Pa?}} \\
    \midrule
    \multicolumn{3}{l}{\textbf{\textit{Style Features}}} \\
    \multirow{2}{*}{Formality} & \multirow{2}{*}{Informal$\leftrightarrow$Formal} & GYAFC\footnote{GYAFC data: \url{https://github.com/raosudha89/GYAFC-corpus}} \citep{rao-tetreault-2018-dear} & 50K & {\cmark}
    \\ 
    &  & XFORMAL\footnote{GYAFC data: \url{https://github.com/Elbria/xformal-FoST}} \citep{briakou2021ola} & 1K & {\cmark}
    \\ \hline
    Politeness & Impolite$\rightarrow$Polite & Politeness\footnote{Politeness data: \url{https://github.com/tag-and-generate/politeness-dataset}} \citep{madaan2020politeness} & 1M & \xmark
    \\ \hline
    Gender & Masculine$\leftrightarrow$Feminine & Yelp Gender\footnote{The Yelp Gender dataset is from the Yelp Challenge \url{https://www.yelp.com/dataset} and its preprocessing needs to follow \citet{prabhumoye-etal-2018-style}.} \citep{prabhumoye-etal-2018-style} & 2.5M & \xmark
    \\ \hline
    Humor\&\newline Romance & Factual$\leftrightarrow$Humorous$\leftrightarrow$\newline Romantic & FlickrStyle\footnote{FlickrStyle data: \url{https://github.com/lijuncen/Sentiment-and-Style-Transfer/tree/master/data/imagecaption}} \citep{Gan2017StyleNetGA} & 5K & \cmark
    \\ \hline
    Biasedness & Biased$\rightarrow$Neutral & Wiki Neutrality\footnote{Wiki Neutrality data: \url{http://bit.ly/bias-corpus}} \citep{pryzant2020automatically} & 181K & \cmark \\ \hline
    \multirow{3}{*}{Toxicity} & \multirow{3}{*}{Offensive$\rightarrow$Non-offensive} & Twitter~\cite{nogueira-dos-santos-etal-2018-fighting} & 58K & \multirow{3}{*}{\xmark}
    \\
    & & Reddit \cite{nogueira-dos-santos-etal-2018-fighting} & 224K
    \\
    & & Reddit Politics \cite{tran2020towards}& 350K \\ \hline
    \multirow{2}{*}{Authorship} & Shakespearean$\leftrightarrow$Modern & Shakespeare \cite{xu2012paraphrasing} & 18K & \multirow{2}{*} {\cmark} \\
    & Different Bible translators & Bible\footnote{Bible data: \url{https://github.com/keithecarlson/StyleTransferBibleData}} \citep{carlson2018evaluating} & 28M &
    \\ \hline
    \multirow{4}{*}{Simplicity} & \multirow{4}{*}{Complicated$\rightarrow$Simple} & PWKP \cite{zhu2010monolingual} & 108K & \cmark \\
    & & Expert \citep{bercken2019evaluating} & 2.2K & \cmark \\
    & & MIMIC-III\footnote{MIMIC-III data: Request access at \url{https://mimic.physionet.org/gettingstarted/access/} and follow the preprocessing of \citet{weng2019unsupervised}} \cite{weng2019unsupervised} & 59K & \xmark \\
    & & MSD\footnote{MSD data: \url{https://srhthu.github.io/expertise-style-transfer/}} \citep{cao-etal-2020-expertise} & 114K & \cmark \\ \hline
    \multirow{2}{*}{Engagingness} & \multirow{2}{*}{Plain$\rightarrow$Attractive} & Math\footnote{Math data: \url{https://gitlab.cs.washington.edu/kedzior/Rewriter/}} \cite{koncel-kedziorski2016theme} & $<$1K & {\cmark} \\
    & & TitleStylist\footnote{TitleStylist data: \url{https://github.com/jind11/TitleStylist}} \citep{jin-etal-2020-hooks} & 146K & \xmark
    \\ \hline
    \multicolumn{4}{l}{\textbf{\textit{Content Preferences}}} \\
    \multirow{2}{*}{Sentiment} & \multirow{2}{*}{Positive$\leftrightarrow$Negative} & Yelp\footnote{Yelp data: \url{https://github.com/shentianxiao/language-style-transfer}} \citep{shen2017style} & 250K & \multirow{2}{*}{\xmark} \\
    & & Amazon\footnote{Amazon data: \url{https://github.com/lijuncen/Sentiment-and-Style-Transfer/tree/master/data/amazon}} \citep{he2016ups} & 277K & \\ \hline
    Topic & Entertainment$\leftrightarrow$Politics & Yahoo! Answers\footnote{Yahoo! Answers data: \url{https://webscope.sandbox.yahoo.com/catalog.php?datatype=l&did=11}} \cite{huang2020cycle} & 153K & \xmark \\ \hline
    Politics & Democratic$\leftrightarrow$Republican & Political\footnote{Political data: \url{https://nlp.stanford.edu/robvoigt/rtgender/}} \citep{voigt2018rtgender} & 540K & \xmark \\
    \bottomrule
    \end{tabular}%
  \endgroup
\end{table*}

We list the common subtasks and corresponding datasets for neural TST in Table~\ref{tab:task_type}. The attributes of interest vary from style features (e.g., formality and politeness) to content preferences (e.g., sentiment and topics). Each task of which will be elaborated below.





\paragraph{Formality.}
Adjusting the extent of formality in text was first proposed by \citet{hovy1987generating}. It is one of the most distinctive stylistic aspects that can be observed through many linguistic phenomena, such as more full names (e.g., ``television'') instead of abbreviations (e.g., ``TV''), and more nouns (e.g., ``solicitation'') instead of verbs (e.g., ``request''). 
The formality dataset, Grammarly’s Yahoo Answers Formality Corpus (GYAFC) \cite{rao-tetreault-2018-dear},
contains 50K formal-informal pairs retrieved by first getting 50K informal sentences from the Yahoo Answers corpus, and then recruiting crowdsource workers to rewrite them in a formal way. \myRed{\citet{briakou2021ola} extend the formality dataset to a multilingual version with three more languages, Brazilian Portuguese, French, and Italian.}

\paragraph{Politeness.}
Politeness transfer \cite{madaan2020politeness} aims to control the politeness in text. For example, ``Could you please send me the
data?'' is a more polite expression than ``send
me the data!''. \citet{madaan2020politeness} compiled a dataset of 1.39 million automatically labeled instances from the raw Enron corpus \cite{shetty2004enron}. As politeness is culture-dependent, this dataset mainly focuses on politeness in North American English.

\paragraph{Gender.} 
Linguistic phenomena related to gender is a heated research area
\cite{trudgill1972sex,lakoff1973language,tannen1990gender,argamon2003gender,knight2005quantitative}. The gender-related TST dataset is proposed by \cite{prabhumoye-etal-2018-style} who compiled 2.5M reviews from Yelp Dataset Challenge that are labeled with the gender of the user.

\paragraph{Humor\&Romance.}
Humor and romance are some artistic attributes that can provide readers with joy. \citet{li-etal-2018-delete} first propose to borrow the FlickrStyle stylized caption dataset \cite{Gan2017StyleNetGA} from the computer vision domain. In the FlickrStyle image caption dataset, each image has three captions, with a factual, a humorous, and a romantic style, respectively. By keeping only the captions of the three styles, \citet{li-etal-2018-delete} created a subset of the FlickrStyle dataset of 5K parallel (factual, humorous, romantic) triplets.

\paragraph{Biasedness.}
Wiki Neutrality Corpus \cite{pryzant2020automatically} is the first corpus of biased and neutralized sentence pairs. It is collected from Wikipedia revisions that adjusted the tone of existing sentences to a more neutral voice. The types of bias in the biased corpus include framing bias, epistemological bias, and demographic bias.

\paragraph{Toxicity.}
Another important use of TST is to fight against offensive language. 
\citet{tran2020towards} collect 350K offensive sentences and 7M non-offensive sentences by crawling sentences from Reddits using a list of restricted words.

\paragraph{Authorship.}
Changing the tone of the author is an artistic use of text style transfer. \citet{xu2012paraphrasing} created an aligned corpus of 18K pairs of Shakespearean English and its modern English translation. \citet{carlson2018evaluating} collected 28M parallel data from English versions of the Bible by different translators.

\paragraph{Simplicity.}
Another important use of TST is to lower the language barrier for readers, such as translating
legalese, medical jargon, or other professional text into simple English, to avoid discrepancies between expert wordings and laymen's understanding \cite{tan2017internet}. Common tasks include converting standard English Wikipedia into Simple Wikipedia whose dataset contains 108K samples \cite{zhu2010monolingual}. Another task is to simplify medical descriptions to patient-friendly text, including a dataset with 2.2K samples \citep{bercken2019evaluating}, another non-parallel dataset with 59K free-text discharge summaries compiled from MIMIC-III \cite{weng2019unsupervised}, and a more recent parallel dataset with 114K samples compiled from from the health reference Merck Manuals (MSD) where discussions on each medical topic has one version for professionals, and the other for consumers \cite{cao-etal-2020-expertise}.

\paragraph{Sentiment.}
Sentiment modification is the most popular task in previous work on TST. It aims to change the sentiment polarity in reviews, for example from a negative review to a positive review, or vice versa. There is also work on  transferring sentiments on fine-grained review ratings, e.g., 1-5 scores. Commonly used datasets include Yelp reviews \cite{shen2017style}, and Amazon product reviews \cite{he2016ups}.

\paragraph{Topic.}
There are a few works that cover topic transfer. For example, \citet{huang2020cycle} form a two-topic corpus by compiling Yahoo! Answers under two topics, entertainment and politics, respectively. There is also a recent dataset with 21 text styles such as Sciences, Sport, Politics, and others \cite{zeng2020style}.

\paragraph{Political Slant.}
Political slant transfer proposed by \citet{prabhumoye-etal-2018-style} aims to transfer the political view in text. For example, a republican's comment can be ``defund all illegal immigrants,'' while democrats are more likely to support humanistic actions towards immigrants. The political slant dataset \cite{voigt2018rtgender} is collected from comments on Facebook posts of the United States Senate and House members. The dataset uses top-level comments directly responding to the posts of a democratic or republican Congressperson. There are 540K training, 4K development, and 56K test instances in the dataset.

\paragraph{Combined Attributes.}
\citet{Lample2019MultipleAttributeTR} propose a more challenging setting of text attribute transfer -- multi-attribute transfer. For example, the source sentence can be a positive review on an Asian restaurant written by a male reviewer, and the target sentence is a negative review on an American restaurant written by a female.
Each of their datasets has 1-3 independent categories of attributes. Their first dataset is FYelp, which is compiled from the Yelp Dataset Challenge, labeled with sentiment (positive or negative),  gender (male or female), and eatery category (American, Asian, bar, dessert, or Mexican).
Their second dataset, Amazon, which is based on the Amazon product review dataset \cite{li-etal-2018-delete}, contains the following attributes: sentiment (positive or negative), and product category (book, clothing, electronics, movies, or music). Their third dataset, Social Media Content dataset, collected from internal Facebook data which is private, contains gender (male or female), age group (18-24 or 65+), and writer-annotated feeling (relaxed or annoyed).

\section{How to Evaluate Style Transfer?}\label{sec:eval}

A successful style-transferred output not only needs to demonstrate the correct target style, but also, due to the uncontrollability of neural networks, we need to verify that it preserves the original semantics, and maintains natural language fluency. Therefore, the commonly used practice of evaluation considers the following three criteria: (1) transferred style strength, (2) semantic preservation, and (3) fluency. 

We will first introduce the practice of automatic evaluation on the three criteria, discuss the benefits and caveats of automatic evaluation, and then introduce human evaluation as a remedy for some of the intrinsic weaknesses of automatic evaluation. Finally, we will suggest some standard practice of TST evaluation for future work. The overview of evaluation methods regarding each criterion is listed in Table~\ref{tab:eval_methods}.

\begin{table*}[ht]
    \caption{Overview of evaluation methods for each criterion.}
    \label{tab:eval_methods}
    \centering
    \resizebox{\textwidth}{!}{
    \begin{tabular}{lllll}
    \toprule
        \textbf{Criterion} & \textbf{Automatic Evaluation} & \textbf{Human Evaluation} \\\hline
        Overall & BLEU with gold references & Rating or ranking \\
        \quad - Transferred Style Strength & Accuracy by a separately trained style classifier & Rating or ranking \\
        \quad - Semantic Preservation & BLEU/ROUGE/etc. with (modified) inputs & Rating or ranking \\
        \quad - Fluency & Perplexity by a separately trained LM & Rating or ranking \\
    \bottomrule
    \end{tabular}
    }
\end{table*}

\subsection{Automatic Evaluation}\label{sec:auto_eval}
Automatic evaluation provides an economic, reproducible, and scalable way to assess the quality of generation results. However, due to the complexities of natural language, each metric introduced below can address certain aspects, but also has intrinsic blind spots.

\paragraph{BLEU with Gold References.}

Similar to many text generation tasks, text style transfer also has human-written references on several datasets (e.g., Yelp, Captions, etc.), so it is common to use the BLEU score \cite{papineni-etal-2002-bleu} between the gold references and model outputs. Using BLEU to evaluate TST models has been seen across pre-deep learning works \cite{xu2012paraphrasing,jhamtani2017shakespearizing} and deep learning approaches \cite{rao-tetreault-2018-dear,li-etal-2018-delete,jin2019imat}.

There are three problems with using BLEU between the gold references and model outputs:
\begin{enumerate}[align=left, 
leftmargin=*,
label=Problem \arabic*), ref= \arabic*]
    \item It mainly evaluates content and simply copying the input can result in high BLEU scores \label{list:prob1}
    \item BLEU is shown to have low correlation with human evaluation \label{list:prob2}
    \item Some datasets do not have human-written references \label{list:prob3} 
\end{enumerate}

Problem~\ref{list:prob1}: Different from machine translation, where using BLEU only is sufficient, TST has to consider the caveat that simply copying the input sentence can achieve high BLEU scores with the gold references \myRed{on many datasets (e.g., $\sim$40 on Yelp, $\sim$20 on Humor\&Romance, $\sim$50 for informal-to-formal style transfer, and $\sim$30 for formal-to-informal style transfer). It is because most text rewrites have a large extent of n-gram overlap with the source sentence. In contrast, machine translation does not have this concern, because the vocabulary of its input and output are different, and copying the input sequence does not give high BLEU scores.} 
A possible fix to consider is to combine BLEU with PINC \cite{chen2011collecting} as in paraphrasing \cite{xu2012paraphrasing,jhamtani2017shakespearizing}. By using PINC and BLEU as a 2-dimensional metric, we can minimize the n-gram overlap with the source sentence but maximize the n-gram overlap with the reference sentences.

Problem~\ref{list:prob2}\&\ref{list:prob3}: Other problems include insufficient correlation of BLEU with human evaluations (e.g., $\leq$0.30 w.r.t. human-rated grammaticality shown in \citet{li-etal-2018-delete} and $\leq$0.45 w.r.t. human evaluations shown in \citet{mir-etal-2019-evaluating}), and the unavailability of human-written references for some datasets (e.g., gender and political datasets \cite{prabhumoye-etal-2018-style}, and the politeness dataset \cite{madaan2020politeness}). A commonly used fix is to make the evaluation more fine-grained using three different independent aspects, namely transferred style strength, semantic preservation, and fluency, which will be detailed below.

\paragraph{Transferred Style Strength.}

To automatically evaluate the transferred style strength, most works separately train a style classifier to distinguish the attributes \cite{Hu2017TowardCG,shen2017style,fu2018style,li-etal-2018-delete,prabhumoye-etal-2018-style}.\footnote{Note that this style classifier usually report 80+\% or 90+\% accuracy, and we will discuss the the problem of false positives and false negatives in last paragraph of automatic evaluation.} This classifier is used to judge whether each sample generated by the model conforms to the target attribute. The transferred style strength is calculated as $\frac{\text{\# test samples correctly classified}}{\text{\# all test samples}}$. \citet{li-etal-2018-delete} shows that the attribute classifier correlates well with human evaluation on some datasets (e.g., Yelp and Captions), but has almost no correlation with others (e.g., Amazon). The reason is that some product genres has a dominant number of positive or negative reviews.


\paragraph{Semantic Preservation.}
Many metrics can be applied to measure the similarity between the input and output sentence pairs, including BLEU~\citep{papineni-etal-2002-bleu}, ROUGE~\citep{lin-och-2004-automatic}, METEOR~\citep{banerjee-lavie-2005-meteor}, chrF~\citep{popovic-2015-chrf}, Word Mover Distance (WMD)~\citep{kusner2015word}. Recently, some additional deep-learning-based metrics are proposed, such as cosine similarity based on sentence embeddings~\citep{fu2018style}, and BERTScore~\citep{Zhang2020BERTScoreET}. There are also evaluation metrics that are specific for TST such as the Part-of-Speech distance~\citep{tian2018structured}. Another newly proposed metric is to first delete all attribute-related expressions in the text, and then apply the above similarity evaluations \cite{mir-etal-2019-evaluating}.
Among all the metrics, \citet{mir-etal-2019-evaluating,yamshchikov2020style} showed that METEOR and WMD have better correlation with human evaluation than BLEU, although, in practice, BLEU is the most widely used metric to evaluate the semantic similarity between the source sentence and style-transferred output~\cite{yang2018unsupervised,madaan2020politeness}.


\paragraph{Fluency.}
Fluency is a basic requirement for natural language outputs. To automate this evaluation, perplexity is calculated via a language model (LM) pretrained on the training data of all attributes~\citep{yang2018unsupervised}. However, the effectiveness of perplexity remains debateable, as \citet{pang2018unsupervised} showed its high correlation with human ratings of fluency, whereas \citet{mir-etal-2019-evaluating} suggested no significant
correlation between perplexity and human scores. We note that perplexity by LM can suffer from the following undesired properties:
\begin{enumerate}
    \item Biased towards shorter sentences than longer sentences.
    \item For the same meaning, less frequent words will have worse perplexity (e.g., agreeable) than frequent words (e.g., good).
    \item A sentence's own perplexity will change if the sentence prior to it changes. 
    \item LMs are not good enough yet.
    \item LMs do not necessarily handle well the domain shift between their training corpus and the style-transferred text.
    \myRed{\item Perplexity scores produced by LMs are sensitive to the training corpora, LM architecture and configuration, as well as optimization configuration. Therefore, different models' outputs must be evaluated by exactly the same LM for fair comparison, which adds more difficulty to benchmarking. }
\end{enumerate}
Such properties will bias against certain models, which is not desired for an evaluation metric. 
As a potential remedy, future researchers can try grammaticality checker to score the generated text.


\paragraph{Task-Specific Criteria.}
As TST can serve as a component for other downstream applications, some task-specific criteria are also proposed to evaluate the quality of generated text. For example, \citet{reiter2003lessons} evaluated the effect of their tailored text on reducing smokers' intent to smoke through clinical trials. \citet{jin-etal-2020-hooks} applied TST to generate eye-catchy headlines so they have an attractive score, and future works in this direction can also test the click-through rates. \citet{Hu2017TowardCG} evaluated how the generated text as augmented data can improve the downstream attribute classification accuracy.

\paragraph{Tips for Automatic Metrics.}

For the evaluation metrics that rely on the pretrained models, namely the style classifier and LM, we need to beware of the following:
\begin{enumerate}
    \item The pretrained models for automatic evaluation should be separate from the proposed TST model
    \item Machine learning models can be imperfect, so we should be aware of the potential false positives and false negatives
    \item The pretrained models are imperfect in the sense that they will favor towards a certain type of methods
\end{enumerate}
For the first point, it is important to not use the same style classifier or LM in the proposed TST approach, otherwise it can overfit or hack the metrics.

For the second point, we need to understand what can be the false positives and false negatives of the generated outputs. An illustrative example is that if the style classifier only reports 80+\% performance (e.g., on the gender dataset \cite{prabhumoye-etal-2018-style} and Amazon dataset \cite{li-etal-2018-delete}), even perfect style rewrites can only score 80+\%, but maybe an imperfect model can score 90\% because it can resemble the imperfect style classification model more and makes advantage of the \textit{false positives}.
Other reasons for false positives can be adversarial attacks. \citet{jin2020bert} showed that merely paraphrasing by synonyms can drop the performance of high-accuracy classification models from TextCNN \cite{kim2014convolutional} to BERT \cite{devlin2019bert} by 90+\%. Therefore, higher scores by the style classifier does not necessarily indicate more successful transfer.
Moreover, the style classifier can produce \textit{false negatives} if there is a distribution shift between the training data and style-transferred outputs. For example, in the training corpus, a product may appear often with the \textit{positive} attribute, and in the style-transferred outputs, this product co-occurs with the opposite, \textit{negative} attribute. Such false negatives are observed on the Amazon product review dataset \cite{li-etal-2018-delete}. On the other hand, the biases of the LM correlate with sentence length, synonym replacement, and prior context. 

The third point is a direct result implied by the second point, so in practice, we need to keep in mind and check whether the proposed model takes advantage of the evaluation metrics or makes improvements that are generalizable.

\subsection{Human Evaluation}

Compared to the pros and cons of the automatic evaluation metrics mentioned above, human evaluation stands out for its flexibility and comprehensiveness. For example, when asking humans to evaluate the fluency, we do not need to worry for the bias towards shorter sentences as in the LM. We can also design criteria that are not computationally easy such as comparing and ranking the outputs of multiple models. There are several ways to conduct human evaluation. In terms of evaluation types, there are pointwise scoring, namely asking humans to provide absolute scores of the model outputs, and pairwise comparison, namely asking humans to judge which of the two outputs is better, or providing a ranking for multiple outputs.
In terms of the criteria, humans can provide overall evaluation, or separate scores for transferred style strength, semantic preservation, and fluency.

However, the well-known limitations of human evaluation are cost and  irreproducibility.
Performing human evaluations can be time consuming, which may result in significant time and financial costs. Moreover, the human evaluation results in two studies are often not directly comparable, because human evaluation results tend to be subjective and not easily irreproducible \cite{belz2020reprogen}. Moreover, some styles are very difficult to evaluate without expertise and extensive reading experience.

As a remedy, we encourage future researchers to report inter-rater agreement scores such as the Cohen's kappa \cite{cohen1960coefficient} and Krippendorff's alpha \cite{krippendorff2018content}. \myRed{\citet{briakou2021review} also recommends to standardize and describe evaluation protocols (e.g., linguistic background of the annotators, compensation, detailed annotation instructions for each evaluation aspect), and release annotations.}

\paragraph{Tips for Human Evaluation.}
As common practice, most works use 100 outputs for each style transfer direction (e.g., 100 outputs for formal$\rightarrow$informal, and 100 outputs for informal$\rightarrow$formal), and two human annotators for each task~\citep{shen2017style,fu2018style,li-etal-2018-delete}.

\subsection{Suggested Evaluation Settings for Future Work}

Currently, the experiments of various TST works do not adopt the same setting, making it difficult to do head-to-head comparison among the empirical results of multiple works. Although it is reasonable to customize the experimental settings according to the needs of a certain work, it is suggested to at least use the standard setting in at least one of the many reported experiments, to make it easy to compare with previous and future works. For example, at least (1) experiment on at least one commonly used dataset, (2) list up-to-date best-performing previous models as baselines, (3) report on a superset of the most commonly used metrics, and (4) release system outputs.

For (1), we suggest that future works use at least one of the most commonly used benchmark datasets, such as the Yelp data prepreocessed by \citet{shen2017style} and its five human references provided by \citet{jin2019imat}, Amazon data preprocessed by \citet{li-etal-2018-delete}, and formality data provided by \citet{rao-tetreault-2018-dear}.

For (2), we suggest that future works actively check the latest style transfer papers curated at \url{https://github.com/fuzhenxin/Style-Transfer-in-Text} and our repository \url{https://github.com/zhijing-jin/Text_Style_Transfer_Survey}, and compare with the state-of-the-art performances instead of older ones. We also call for more reproducibility in the community, including source codes and evaluation codes, because, for example, there are several different scripts to evaluate the BLEU scores.

For (3), since no single evaluation metric is perfect and comprehensive enough for TST, it is strongly suggested to use both human and automatic evaluation on three criteria. In evaluation, apart from customized use of metrics, we suggest that most future works to include at least the following evaluation practices:
\begin{itemize}[nolistsep]
    \item Human evaluation: rate at least two state-of-the-art models according to the curated paper lists
    \item Automatic evaluation: at least report the BLEU score with all available references if there exist human-written references (e.g., the five references for the Yelp dataset provided by \citet{jin2019imat}), and BLEU with the input only when there are no human-written references.
\end{itemize}

\myRed{For (4), it will also be very helpful to provide system outputs for each TST paper, so that future works can better reproduce both human and automatic evaluation results. Note that releasing system outputs can help future works' comparison of automatic evaluation results, because there can be different scripts to evaluate the BLEU scores, as well as different style classifiers and LM. It will be a great addition to the TST community if future work can establish an online leaderboard, let existing works upload their output files, and automatically evaluate the model outputs using a standard set of automatic evaluation scripts.}

\section{Methods on Parallel Data}\label{sec:sup}

Over the last several years, various methods have been proposed for text style transfer. In general, they can be categorized based on whether the dataset has parallel text with different styles or several non-parallel mono-style corpora. The rightmost column ``Pa?'' in Table~\ref{tab:task_type} shows whether there exist parallel data for each TST subtask. 
In this section, we will cover TST methods on parallel datasets, and in Section~\ref{sec:unsup} we will detail the approaches on non-parallel datasets.
To ease the understanding for the readers, we will in most cases explain TST on one attribute between two values, such as transferring the formality between informal and formal tones, which can potentially be extended to multiple attributes.

Most methods adopt the standard neural sequence-to-sequence (seq2seq) model with the encoder-decoder architecture, which was initially developed for neural machine translation (NMT)~\citep{sutskever2014sequence,bahdanau2014neural,cho2014properties} and extensively seen on text generation tasks such as summarization \cite{rush2015neural} and many others \cite{song2019mass}. The encoder-decoder seq2seq model can be implemented by either LSTM as in ~\citet{rao-tetreault-2018-dear,shang-etal-2019-semi} or Transformer~\cite{vaswani2017attention} as in~\citet{xu2019formality}. Copy mechanism~\cite{gulcehre2016pointing,see2017get} is also added to better handle stretches of text which should not be changed (e.g., some proper nouns and rare words)~\citep{gu-etal-2016-incorporating,Merity2017PointerSM}. Based on this architecture, recent works have developed multiple directions of improvement: \textbf{multi-tasking}, \textbf{inference techniques}, and \textbf{data augmentation}, which will be introduced below.

\paragraph{Multi-Tasking.}
In addition to the seq2seq learning on paired attributed-text, \citet{xu2019formality} propose adding three other loss functions: (1) classifier-guided loss, which is calculated using a well-trained attribute classifier and encourages the model to generate sentences conforming to the target attribute, (2) self-reconstruction loss, which encourages the seq2seq model to reconstruct the input itself by specifying the desired style the same as the input style, and (3) cycle loss, which first transfers the input sentence to the target attribute and then transfers the output back to its original attribute. Each of the three losses can gain performance improvement of 1-5 BLEU points with the human references \cite{xu2019formality}. Another type of multi-tasking is to jointly learn TST and machine translation from French to English, which improves the performance by 1 BLEU score with human-written references \cite{niu-etal-2018-multi}. 
Specific for formality transfer, \citet{zhang-etal-2020-parallel} multi-task TST and grammar error correction (GEC) so that knowledge from GEC data can be transferred to the informal-to-formal style transfer task. 

Apart from the additional loss designs, using the pretrained language model GPT-2~\citep{radford2019language} can lead to improvement by at least 7 BLEU scores with human references \citet{wang-etal-2019-harnessing}. 

\paragraph{Inference Techniques.}
To avoid the model copying too many parts of the input sentence and not performing sufficient edits to flip the attribute, \citet{kajiwara-2019-negative} first identify words in the source sentence requiring replacement, and then change the words by negative lexically constrained decoding~\citep{post-vilar-2018-fast} that avoids naive copying. Since this method only changes the beam search process for model inference, it can be applied to any text style transfer model without model re-training.

\paragraph{Data Augmentation.}
Since style transfer data is expensive to annotate, there are not as many parallel datasets as in machine translation. Hence, various methods have been proposed for data augmentation to enrich the data. For example, \citet{rao-tetreault-2018-dear} first train a phrase-based machine translation (PBMT) model on a given parallel dataset and then use back-translation~\citep{sennrich-etal-2016-edinburgh} to construct a pseudo-parallel dataset as additional training data, which leads to an improvement of around 9.7 BLEU scores with respect to human written references.


Most recently, \citet{zhang-etal-2020-parallel} use a data augmentation technique by making use of largely available online text. They scrape informal text from online forums and generate back-translations, i.e., informal English $\rightarrow$ a pivot language such as French $\rightarrow$ formal English, where the formality of the back-translated English text is ensured with  a formality classifier that is used to only keep text that is classified as formal text. 



\section{Methods on Non-Parallel Data}\label{sec:unsup}

Parallel data for TST is difficult to obtain, and for some styles impossible to crowd-source (e.g., Mark Twain novels rewritten in Hemmingway's style). Hence, the majority TST methods assume only non-parallel mono-style corpora, and investigate how to build deep learning models based on this constraint.
In this section, we will introduce three main branches of TST methods: disentanglement (Section~\ref{sec:disentanglement}), prototype editing (Section~\ref{sec:proto_nlg}), and pseudo-parallel corpus construction (Section~\ref{sec:pseudo_construct}).

\subsection{Disentanglement}\label{sec:disentanglement}
Disentanglement-based models usually perform the following three actions:
\begin{itemize}
    \item Encode the text $\bm{x}$ with attribute $a$ into a latent representation $\bm{z}$ (i.e., $\bm{x} \rightarrow \bm{z}$) \label{item:disent_encode}
    \item Manipulate the latent representation $\bm{z}$ to remove the source attribute (i.e., $\bm{z} \rightarrow \bm{z}'$) \label{item:disent_manipulate}
    \item Decode into text $\bm{x}'$ with the target attribute $a'$ (i.e., $\bm{z}' \rightarrow \bm{x}'$)
    \label{item:disent_decode}
\end{itemize}

To build such models, the common workflow in disentanglement papers consists of the following three steps:
\begin{enumerate}[align=left, leftmargin=*,label=Step \arabic*), ref=\arabic*]
    \item Select a model as the backbone for the encoder-decoder learning (Section~\ref{sec:enc_dec}) \label{item:select_enc_dec}
    \item Select a manipulation method of the latent representation (Section~\ref{sec:manipulation}) \label{item:select_mani}
    \item For the manipulation method chosen above, select (multiple) appropriate loss functions (Section~\ref{sec:losses}) \label{item:select_loss}
\end{enumerate}

The organization of this section starts with Section~\ref{sec:enc_dec} which introduces the encoder-decoder training objectives that is used for Step~\ref{item:select_enc_dec}. Next, Section~\ref{sec:manipulation} overviews three main approaches to manipulate the latent representation for Step~\ref{item:select_mani}, and Section~\ref{sec:losses} goes through a plethora of training objectives for Step~\ref{item:select_loss}.
Table~\ref{table:disentanglement} provides an overview of existing models and their corresponding configurations. To give a rough idea of the effectiveness of each model, we show their performance on the Yelp dataset.
\begin{table*}[ht]
\caption{Summary of existing disentanglement-based methods and the setting they adopted, with a reference of their performance on the Yelp dataset. For the settings, we include the encoder-decoder training method (Enc-Dec) in Section~\ref{sec:enc_dec}, the disentanglement method (Disen.) in Section~\ref{sec:manipulation}, and the loss types used to control style (Style Control) and content (Content Control) in Section~\ref{sec:losses}. For the model performance, we report automatic evaluation scores including BLEU with the one human reference (BL-Ref) provided by \citet{li-etal-2018-delete}, accuracy (Acc.), BLEU with the input (BL-Inp) and perplexity (PPL). $^*$ marks numbers reported by \citet{liu2019revision}. \myRed{Readers can refer to \citet{Hu2020TextST} for more complete performance results on Yelp.}}
\label{table:disentanglement}
\resizebox{\textwidth}{!}{
\begin{tabular}{lcccc|cccc}
\toprule
 & \multicolumn{4}{c}{\textbf{Settings}}                     & \multicolumn{3}{|c}{\textbf{Performance on Yelp}} \\
 & \textbf{Enc-Dec} & \textbf{Disen.} & \textbf{Style Control} & \textbf{Content Control} & \textbf{BL-Ref} & \textbf{Acc. (\%)} & \textbf{BL-Inp} & \textbf{PPL$\downarrow$} \\ \midrule
\citet{mueller2017sequence} & VAE & LRE & -- & -- & -- & -- & -- & -- \\
\citet{Hu2017TowardCG} & VAE & ACC & ACO & -- & 22.3 & 86.7 & 58.4 & -- \\
\citet{shen2017style} & AE\&GAN & ACC & AdvR$\parallel$AdvO & -- & 7.8 & 73.9 & 20.7 & 72$^*$\\
\citet{fu2018style} & AE & ACC & AdvR & -- & 12.9 & 46.9 & 40.1 & 166.5$^*$\\
\citet{prabhumoye-etal-2018-style} & AE & ACC & ACO & -- & 6.8 & 87.2 & -- & 32.8$^*$\\
\citet{Zhao2018AdversariallyRA} & GAN & ACC & AdvR & -- & -- & 73.4 & 31.2 & 29.7 \\
\citet{yang2018unsupervised} & AE & ACC & LMO & -- & -- & 91.2 & 57.8 & 47.0\&60.9 \\
\citet{logeswaran2018content} & AE & ACC & AdvO & Cycle & -- & 90.5 & -- & 133 \\
\citet{tian2018structured} & AE & ACC & AdvO & Noun & 24.9 & 92.7 & 63.3 & -- \\
\citet{liao-etal-2018-quase} & VAE & LRE & -- & -- & -- & 88.3 & -- & -- \\
\citet{romanov-etal-2019-adversarial} & AE & LRS & ACR\&AdvR & -- & -- & -- & -- & -- \\
\citet{john-etal-2019-disentangled} & AE\&VAE & LRS & ACR\&AdvR & BoW\&AdvBoW & -- & 93.4 & -- & -- \\
\citet{bao-etal-2019-generating} & VAE & LRS & ACR\&AdvR & BoW\&AdvBoW & -- & -- & -- & -- \\
\citet{dai-etal-2019-style} & AE & ACC & ACO & Cycle & 20.3 & 87.7 & 54.9 & 73 \\
\citet{wang2019controllable} & AE & LRE & -- & -- & 24.6 & 95.4 & -- & 46.2 \\
\citet{Li2020ComplementaryAC} & GAN & ACC & ACO\&AdvR & -- & -- & 95.5 & 53.3 & -- \\
\citet{liu2019revision} & VAE & LRE & -- & -- & 18.8 & 92.3 & -- & 18.3 \\
\citet{Yi2020TextST} & VAE & ACC & ACO & Cycle & 26.0 & 90.8 & -- & 109 \\
\citet{jin-etal-2020-hooks} & AE & LRE & -- & -- & -- & -- & -- & -- \\
\bottomrule
\end{tabular}}
\end{table*}

\subsubsection{Encoder-Decoder Training Method}\label{sec:enc_dec}
There are three model choices to obtain the latent representation $\bm{z}$ from the discrete text $\bm{x}$ and then decode it into the new text $\bm{x}'$ via reconstruction training: auto-encoder (AE), variational auto-encoder (VAE), and generative adversarial networks (GANs).
\paragraph{Auto-Encoder (AE).}
Auto-encoding  is a commonly used method to learn the latent representation $\bm{z}$, which first encodes the input sentence $\bm{x}$ into a latent vector $\bm{z}$ and then reconstructs a sentence as similar to the input sentence as possible.
AE is used in many TST works \cite[e.g.,][]{shen2017style,Hu2017TowardCG,fu2018style,Zhao2018AdversariallyRA,prabhumoye-etal-2018-style,yang2018unsupervised}. 
To avoid auto-encoding from blindly copying all the elements from the input, \citet{hill2016learning} adopt denoising auto-encoding (DAE) \cite{vincent2010stacked} to replace AE in NLP tasks. Specifically, DAE first passes the input sentence $\bm{x}$ through a noise model to randomly drop, shuffle, or mask some words, and then reconstructs the original sentence from this corrupted sentence. This idea is used in later TST works, e.g., \citet{Lample2019MultipleAttributeTR,jin-etal-2020-hooks}. \myRed{As pre-trained models became prevalent in recent years, the DAE training method has increased in popularity relative to its counterparts such as GAN and VAE, because pre-training over large corpora can grant models better performance in terms of semantic preservation and fluency~\citep{Lai2021ThankYB,Riley2021TextSETTRFT}.} 

\paragraph{Variational Auto-Encoder (VAE).}
Instead of reconstructing data based on the deterministic latent representations by AE, a variational auto-encoder (VAE)~\citep{Kingma2014AutoEncodingVB,Rezende2014vae} reconstructs data based on the sampled latent vector from its posterior, and use the regularization by Kullback–Leibler divergence. VAE is also commonly used in TST works \cite{mueller2017sequence,Hu2017TowardCG,liu2019revision,liao-etal-2018-quase,Yi2020TextST,tikhonov-etal-2019-style}.
The VAE loss is formulated as
\begin{equation}
\begin{split}
     \mathcal{L}_{\mathrm{VAE}}(\bm{\theta}_{\mathrm{E}},\bm{\theta}_{\mathrm{G}})
     = -\mathbb{E}_{q_{\mathrm{E}}(\bm{z}|\bm{x})}\log p_{\mathrm{G}}(\bm{x}|\bm{z}) 
     + \lambda \mathrm{KL}\Big{[}q_{\mathrm{E}}(\bm{z}|\bm{x})||p(\bm{z})\Big{]},
\end{split}
\end{equation}
where $\lambda$ is the hyper-parameter to balance the reconstruction loss and the KL term, $p(\bm{z})$ is the prior drawn from the standard normal distribution of $\mathcal{N}(\bm{0}, \bm{I})$, and $q_{\mathrm{E}}(\bm{z}|\bm{x})$ is the posterior in the form of $\mathcal{N}(\bm{\mu}, \bm{\sigma})$, where $\bm{\mu}$ and $\bm{\sigma}$ are predicted by the encoder.
\begin{figure*}[h]
     \centering
     \begin{subfigure}[b]{0.33\textwidth}
         \centering
         \includegraphics[width=\textwidth]{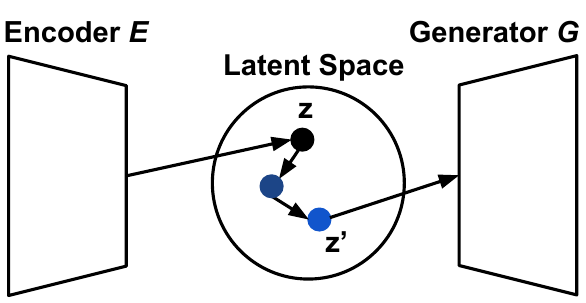}
         \caption{Latent Representation Editing}
         \label{figure:disentanglement-type-1}
     \end{subfigure}
     \hfill
     \begin{subfigure}[b]{0.32\textwidth}
         \centering
         \includegraphics[width=\textwidth]{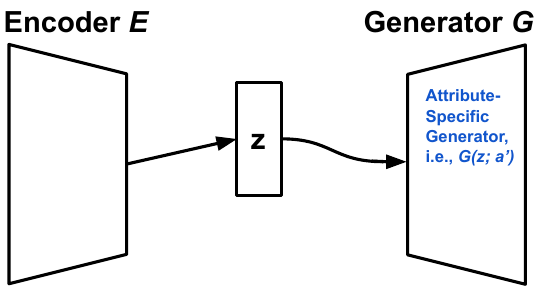}
         \caption{Attribute Code Control}
         \label{figure:disentanglement-type-2}
     \end{subfigure}
     \hfill
     \begin{subfigure}[b]{0.33\textwidth}
         \centering
         \includegraphics[width=\textwidth]{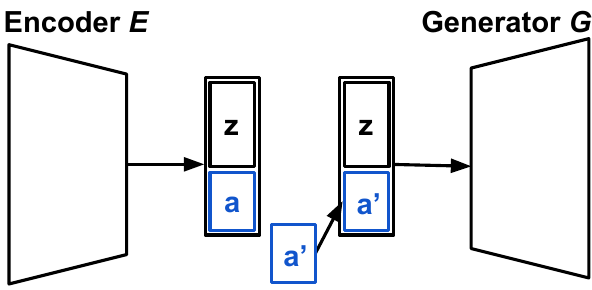}
         \caption{Latent Representation Splitting}
         \label{figure:disentanglement-type-3}
     \end{subfigure}
        \caption{Three methods to manipulate the latent space based on disentanglement for text style transfer.}
        \label{figure:disentanglement-three-types}
\end{figure*}

\paragraph{Generative Adversarial Networks (GANs).}
GANs~\citep{Goodfellow2014GenerativeAN} can also be applied to TST \cite{shen2017style,Zhao2018AdversariallyRA,Li2020ComplementaryAC}. The way GANs work is to first approximate the samples drawn from a true distribution $\bm{z}$ by employing a noise sample $\bm{s}$ and a generator function $G$ to produce $\widehat{\bm{z}} = G(\bm{s})$. Next, a critic/discriminator $f_c(\bm{z})$ is used to distinguish real data and generated samples. The critic is trained to distinguish the real samples from generated samples, and the generator is trained to fool the critic. Formally, the training process is expressed as a min-max game played among the encoder $E$, generator $G$, and the critic $f_c$:
\begin{equation}
    \max_{c}\min_{E,G}\mathcal{L}_{\mathrm{GAN}}
    = -\mathbb{E}_{p(\bm{z})}\log p_{\mathrm{G}}(\bm{x}|\bm{z}) +\mathbb{E}_{p(\bm{z})}f_c(\bm{z})-\mathbb{E}_{p(\widehat{\bm{z}})}f_c(\widehat{\bm{z}}).
\end{equation}

\subsubsection{Latent Representation Manipulation}\label{sec:manipulation}

Based on the general encoder and decoder training method, the core element of disentanglement is the manipulation of latent representation $\bm{z}$. Figure \ref{figure:disentanglement-three-types} illustrates three main methods: latent representation editing, attribute code control, and latent representation splitting. In addition, the ``Disen.'' column of Table~\ref{table:disentanglement} shows the type of latent representation manipulation for each work in disentanglement.

The first approach, \textit{Latent Representation Editing} (LRE), shown in Figure~\ref{figure:disentanglement-type-1}, is achieved by ensuring two properties of the latent representation $\bm{z}$. The first property is that $\bm{z}$ should be able to serve as the latent representation for auto-encoding, namely aligning $f_c(\bm{z})$ with the input $\bm{x}$, where $\bm{z} \overset{\Delta}{=} E(\bm{x})$.
The second property is that $\bm{z}$ should be learned such that it incorporates the new attribute value of interest $a'$. To achieve this, the common practice is to first learn an attribute classifier $f_c$, e.g., a multilayer perceptron (MLP) that takes the latent representation $\bm{z}$ as input, and then iteratively update $\bm{z}$ within the constrained space by the first property and in the same time maximize the prediction confidence score regarding $a'$ by this attribute classifier 
~\citep{mueller2017sequence,liao-etal-2018-quase,wang2019controllable,liu2019revision}. An alternative way to achieve the second property is to multi-task by another auto-encoding task on the corpus with the attribute $a'$ and share most layers of the transformer except the query transformation and layer normalization layers \cite{jin-etal-2020-hooks}.


The second approach, \textit{Attribute Code Control} (ACC), as shown in Figure~\ref{figure:disentanglement-type-2}, first enforces the latent representation $\bm{z}$ of the sentence $\bm{x}$ to contain all information except its attribute value $a$ via adversarial learning, and then the transferred output is decoded based on the combination of $\bm{z}$ and a structured attribute code $\bm{a}$ corresponding to the attribute value $a$. During the decoding process, the attribute code vector $\bm{a}$ controls the attribute of generated text by acting as either the initial state~\citep{shen2017style,Yi2020TextST} or the embedding~\citep{fu2018style,dai-etal-2019-style}. 

The third approach, \textit{Latent Representation Splitting} (LRS), as illustrated in Figure~\ref{figure:disentanglement-type-3}, first disentangles the input text into two parts: the latent attribute representation $\bm{a}$, and semantic representation $\bm{z}$ that captures attribute-independent information. We then replace the source attribute $\bm{a}$ with the target attribute $\bm{a}'$, and the final transferred text is generated using the combination of $\bm{z}$ and ${a}'$ \cite{john-etal-2019-disentangled,romanov-etal-2019-adversarial}. 

\subsubsection{Training Objectives}\label{sec:losses}
When disentangling the attribute information $a$ and the attribute-independent semantic information $\bm{z}$, we need to achieve two aims:
\begin{enumerate}[label=Aim~\arabic*), ref=Aim~\arabic*,leftmargin=*,align=left]
    \item The target attribute is \textit{fully} and \textit{exclusively} controlled by $\bm{a}$ (and not $\bm{z}$). We typically use style-oriented losses to achieve this aim (Section~\ref{sec:aim1}). \label{list:aim1}
    \item The attribute-independent information is \textit{fully} and \textit{exclusively} captured by $\bm{z}$ (and not $\bm{a}$). Content-oriented losses are more often used for this aim (Section~\ref{sec:aim2}). \label{list:aim2}
\end{enumerate}
We describe the various style-oriented and content-oriented losses  below.
\subsubsubsection{Style-Oriented Losses} \label{sec:aim1}

To achieve \ref{list:aim1}, many different \textit{style-oriented losses} have been proposed, to nudge the model to learn a more clearly disentangled $\bm{a}$ and exclude the attribute information from $\bm{z}$.

\paragraph{Attribute Classifier on Outputs (ACO).}
ACO aims to make sentences generated by the generator ${G}$ carry the target attribute $a'$ according to a pre-trained attribute classifier $f_c$~\citep{Hu2017TowardCG,prabhumoye-etal-2018-style,yamshchikov-etal-2019-decomposing}. The generator $G$ takes as input the learned attribute vector $\widehat{\bm{a}'}$, which can be either an attribute code vector trained from scratch (as in the ACC approach) or the attribute representation disentangled from text (by the LRS approach). We denote the generation process to obtain the transferred sentence $\widehat{\bm{x}'}$ as $\widehat{\bm{x}'} \overset{\Delta}{=} G(E(\bm{x}); \bm{a}')$. Correspondingly, ACO minimizes the following learning objective:
\begin{equation}
    \mathcal{L}_{\mathrm{ACO}}(\bm{\theta}_{\mathrm{G}}, \bm{a}')=-\mathbb{E}_{p(\bm{x})}\log f_c(\bm{x}')
    ~.
    \label{eq:aco}
\end{equation}
In training, ACO can be trained in two ways: either a normal loss function trained by Gumbel-softmax distribution to approximate the discrete training~\citep{jang2016categorical},
or a negative reward for reinforcement learning by policy gradient training \cite{williams1992simple} as in \citet{Luo19DualRL}.

\paragraph{Attribute Classifier on Representations (ACR).} 
Different from the previous ACO objective whose training signal is from the the output sentence $\widehat{\bm{x}'}$, ACR directly enforces the disentangled attribute representation $\bm{a}$ to be correctly classified by the attribute classifier,
by the following objective \cite{john-etal-2019-disentangled,romanov-etal-2019-adversarial}:
\begin{equation}
    \mathcal{L}_{\mathrm{ACR}}(\bm{\theta}_{\mathrm{E}},\bm{\theta}_{f_c})=-\mathbb{E}_{p(\bm{a})}\log f_c(\bm{a})~.
\end{equation}

\paragraph{Adversarial Learning on Representations (AdvR).} 
As the previous ACR explicitly requires the latent $\bm{a}$ to be classified by $f_c$, AdvR trains from another perspective -- enforcing that no attribute-related information is contained in $\bm{z}$ \cite{fu2018style,Zhao2018AdversariallyRA,romanov-etal-2019-adversarial,john-etal-2019-disentangled,tikhonov-etal-2019-style,Li2020ComplementaryAC}. Note that by combining ACR and AdvR, we can make attribute information captured \textit{fully} and \textit{exclusively} in $\bm{a}$.
To achieve AdvR, the encoder $E$ is trained to generate the latent representation $\bm{z} \overset{\Delta}{=} E(\bm{x})$ so that $\bm{z}$ cannot be discriminated by the attribute classifier $f_c$, which is expressed by the following learning objective:
\begin{equation}
    \max_{E} \min_{f_c} \mathcal{L}_{\mathrm{AdvR}}(\bm{\theta}_{\mathrm{E}},\bm{\theta}_{f_c})=-\mathbb{E}_{p(\bm{x})}\log f_c(E(\bm{x}))
    ~.
\end{equation}


Since AdvR can be imbalanced if the number of samples of each attribute value differs largely, an extension of AdvR is to treat different attribute values with equal weight \cite{shen2017style}:
\begin{equation}
\begin{split}
    \max_{E} \min_{f_c} \mathcal{L}_{\mathrm{AAE}}(\bm{\theta}_{\mathrm{E}},\bm{\theta}_{f_c})&=- \mathbb{E}_{p(\bm{x})}\Big{[}\log f_c(E(\bm{x}))\Big{]} \\
    &- \mathbb{E}_{p(\bm{x}')}\Big{[}\log(1-f_c(E(\bm{x}')))\Big{]}~.
\end{split}
\label{equation:alignment-auto-encoder}
\end{equation}
Note that $p(\bm{x})$ is the distribution of sentences of one attribute, and $p(\bm{x}')$ is the distribution of sentences of the other attribute.

\paragraph{Adversarial Learning on Outputs (AdvO).} 
Apart from AdvR that adversarially learn the latent representations, we can also use AdvO to perform adversarial training on the outputs, to make them undistinguishable from the real data \cite{shen2017style,logeswaran2018content,tian2018structured}. Specifically, for each attribute $a_i$, we train a classifier $f_c^{(i)}$ to distinguish between true $\bm{x}_i$ from the mono-style corpus of attribute $a_i$, and the generated sentence $\widehat{\bm{x}_i} \overset{\Delta}{=} G(E(\bm{x}_k); \bm{a}_i)$, where $k \neq i$, which aims to have the attribute $a_i$. The loss function is
\begin{equation}
\begin{split}
    \max_{E,G} \min_{f_c^{(i)}} \mathcal{L}_{\mathrm{AdvO}}^{(i)}(\bm{\theta}_{\mathrm{E}},\bm{\theta}_{\mathrm{G}},\bm{\theta}_{f_c^{(i)}}) 
    & = -\mathbb{E}_{p(\bm{x}_i)}\Big{[}\log f_c^{(i)}(\bm{x}_i)\Big{]} \\
    & - \mathbb{E}_{p(\bm{x}_k)}\Big{[}\log(1-f_c^{(i)}(G(E(\bm{x}_k); a_i)))\Big{]}
    ~.
\end{split}
\label{equation:cross-alignment-auto-encoder}
\end{equation}
In the training process, usually we first optimize all attribute classifiers $f_c^{(i)}$, and then train the encoder, generator, and the attribute classifiers together by optimizing the sum the all AdvO training losses:
\begin{equation}
    \max_{E,G} \sum_i^{|\mathbb{A}|} \min_{f_c^{(i)}} \mathcal{L}_{\mathrm{AdvO}}^{(i)}(\bm{\theta}_{\mathrm{E}},\bm{\theta}_{\mathrm{G}},\bm{\theta}_{f_c^{(i)}})~.
\end{equation}
Note that in order to propagate the gradients, it is feasible to use the sequence of hidden states in the generator instead of discrete text for $G(E(\bm{x}_k); a_i)$ \cite{shen2017style}.

\paragraph{Language Modeling on Outputs (LMO).} 
The above AdvO learns classifiers to distinguish between true samples and generated samples. Such discriminative classification can be alternatively achieved by generative language modeling, namely $\mathrm{LM}_i$ for each mono-style corpus with the attribute $a_i$~\citep{yang2018unsupervised}. Specifically, the training objective for each attribute is 
\begin{equation}
\begin{split}
    \mathcal{L}_{\mathrm{LMO}}^{(i)}(\bm{\theta}_{\mathrm{E}},\bm{\theta}_{\mathrm{G}},\bm{\theta}_{\mathrm{LM}_i})
    =-\mathbb{E}_{p(\bm{x_i})}\Big{[}\log p_{\mathrm{LM}_i}(\bm{x}_i)\Big{]} 
    +\gamma \mathbb{E}_{p(\bm{z}_k)}\Big{[}\log p_{\mathrm{LM}_i}(G(E(\bm{x}_k); a_i))\Big{]}~,
\end{split}
\end{equation}
where $\gamma$ is a hyperparameter to weight the two terms.
The total training objective sums over the losses of all attributes:
\begin{equation}
    \max_{E,G} \sum_i^{|\mathbb{A}|} \min_{\mathrm{LM}^{(i)}} \mathcal{L}_{\mathrm{LMO}}^{(i)}(\bm{\theta}_{\mathrm{E}},\bm{\theta}_{\mathrm{G}},\bm{\theta}_{\mathrm{LM}_i})~.
\end{equation}

\subsubsubsection{Content-Oriented Losses} 
\label{sec:aim2}
The style-oriented losses introduced above ensures the attribute information to be contained in $\bm{a}$, but not necessarily putting constraints on the style-independent semantics $\bm{z}$.
To learn the attribute-independent information fully and exclusively in $\bm{z}$, the following \textit{content-oriented losses} are proposed:
\paragraph{Cycle Reconstruction (Cycle).} 
The cycle reconstruction loss
~\citep{nogueira-dos-santos-etal-2018-fighting,logeswaran2018content,Luo19DualRL,dai-etal-2019-style,Yi2020TextST,huang2020cycle} first encodes a sentence $\bm{x}$ to its latent representation $\bm{z} \overset{\Delta}{=} E(\bm{x})$, and then feed $\bm{z}$ to the generator $G$ to obtain the generated sentence $G(\bm{z})$. Since the alignment of the input and the generated sentence is to preserve attribute-independent semantic information, the generator can be conditioned on any attribute, namely $\bm{a}$ or $\bm{a}'$. The cycle loss constrains the output $\widehat{\bm{x}'}$ to align with the input $\bm{x}$ (and, similarly, the output $\widehat{\bm{x}}$ to align with the input $\bm{x}'$) so that the content information can be preserved:
\begin{equation}
   \mathcal{L}_{\mathrm{Cycle}}(\bm{\theta}_{E},\bm{\theta}_{G})=- \mathbb{E}_{p(\bm{x})}\Big{[}\log p_{\mathrm{G}}(\bm{x}|E(\bm{x}))\Big{]}
   -\mathbb{E}_{p(\bm{x}')}\Big{[}\log p_{\mathrm{G}}(\bm{x}'|E(\bm{x}'))\Big{]}
   ~. \label{eq:cycle_loss}
\end{equation}


One way to train the above cycle loss is by reinforcement learning as done by \citet{Luo19DualRL} who use the loss function as a negative for content preservation. 


\paragraph{Bag-of-Words Overlap (BoW).} 
To approximately measure content preservation, bag-of-words (BoW) features are used by \citet{john-etal-2019-disentangled,bao-etal-2019-generating}.
To focus on content information only, \citet{john-etal-2019-disentangled} exclude
stopwords and style-specific words.

Let us denote the vocabulary set as $\mathbb{V}$. We first predict the distribution of BoW features $q_{\mathrm{BoW}}(\bm{z})$ of the latent representation $\bm{z}$ using softmax on the $1 \times |\mathbb{V}|$ BoW features. We then calculate the cross entropy loss of this BoW distribution $q_{\mathrm{BoW}}(\bm{z})$ against the ground-truth BoW distribution $p_{\mathrm{BoW}}(\bm{x})$ in the input sentence $\bm{x}$.
The BoW loss is formulated as follows:
\begin{equation}
\label{equation:BoW}
    \mathcal{L}_{\mathrm{BoW}}(\bm{\theta}_{\mathrm{E}},\bm{\theta}_{q_{\mathrm{BoW}}})=-p_{\mathrm{BoW}}(\bm{x}) \log q_{\mathrm{BoW}}(\bm{z})
    ~.
\end{equation}

\paragraph{Adversarial BoW Overlap (AdvBoW).} 
BoW ensures the content to be \textit{fully} captured in $\bm{z}$. As a further step, we want to ensure that the content information is exclusively captured in $\bm{z}$, namely not contained in $\bm{a}$ at all, via the following AdvBow loss on $\bm{a}$ \cite{john-etal-2019-disentangled,bao-etal-2019-generating}.

When disentangling $\bm{z}$ and $\bm{a}$ in the LRS framework, 
we train an adversarial classifier $q_{\mathrm{BoW}}(\bm{a})$ to predict the BoW features given $\bm{a}$ by aligning it with the ground-truth BoW distribution $p_{\mathrm{BoW}}(\bm{x})$, namely minimizing
\begin{equation}
\label{equation:AdvBoW}
    \mathcal{L}_{\mathrm{AdvBoW}}(\bm{\theta}_{\mathrm{E}},\bm{\theta}_{q_{\mathrm{BoW}}})=-p_{\mathrm{BoW}}(\bm{x}) \log q_{\mathrm{BoW}}(\bm{z})
    ~.
\end{equation}
The final min-max objective is
\begin{equation}
    \max_{E} \min_{q_{\mathrm{BoW}}} \mathcal{L}_{\mathrm{AdvBoW}}(\bm{\theta}_{\mathrm{E}},\bm{\theta}_{q_{\mathrm{BoW}}}).
\end{equation}


\paragraph{Other Losses/Rewards.}
There are also other losses/rewards in recent work such as the noun overlap loss (Noun) \cite{tian2018structured}, as well as rewards for semantics and fluency~\citep{xu-etal-2018-unpaired,gong-etal-2019-reinforcement,sancheti2020reinforced}. We do not discuss them in much detail because they do not directly operate on the disentanglement of latent representations.



\subsection{Prototype Editing}\label{sec:proto_nlg}
Despite a plethora of models that use end-to-end training of neural networks, the prototype-based text editing approach still attracts lots of attention, since the proposal of a pipeline method called {\it delete, retrieve}, and {\it generate} \cite{li-etal-2018-delete}. 

Prototype editing is reminiscent of early word replacement methods used for TST, such as synonym matching using a style dictionary \cite{sheikha2011formal}, WordNet \cite{khosmood2010automatic,mansoorizadeh2016author}, hand-crafted rules \cite{khosmood2008automatic,castro2017author}, or using hypernyms and definitions to replace the style-carrying words \cite{karadzhov2017case}.

Featuring more controllability and interpretability, prototype editing builds an explicit pipeline for text style transfer from $\bm{x}$ with attribute $a$ to its counterpart $\bm{x}'$ with attribute ${a}'$:
\begin{enumerate}[align=left, 
leftmargin=*,
label=Step \arabic*), ref=\arabic*]
    \item Detect attribute markers of $a$ in the input sentence $\bm{x}$, and delete them, resulting in a content-only sentence (Section~\ref{sec:attr_marker_detect}); \label{item:proto_marker}
    \item Retrieve candidate attribute markers carrying the desired attribute ${a}'$ (Section~\ref{sec:tar_attr_retr}); \label{item:proto_template}
    \item Infill the sentence by adding new attribute markers and make sure the generated sentence is fluent (Section~\ref{sec:gen_from_proto}). \label{item:proto_infill}
\end{enumerate}


\subsubsection{Attribute Marker Detection}\label{sec:attr_marker_detect}

Extracting attribute markers is a non-trivial NLP task. Traditional ways to do it involve first using tagging, parsing and morphological analysis to select features, and then filtering by mutual information and Chi-square testing. In recent deep learning pipelines,
there are three major types of approaches to identify attribute markers: frequency-ratio methods, attention-based methods, and fusion methods.

{\bf Frequency-ratio methods} calculate some statistics for each n-gram in the corpora.
For example, \citet{li-etal-2018-delete} detect the attribute markers by calculating its relative frequency of co-occurrence with attribute $a$ versus ${a}'$, and those with frequencies higher than a threshold are considered the markers of $a$. Using a similar approach, \citet{madaan2020politeness} first calculate the ratio of mean TF-IDF between the two attribute corpora for each n-gram, then normalize this ratio across all possible n-grams, and finally mark those n-grams with a normalized ratio $p$ higher than a pre-set threshold as attribute markers.

{\bf Attention-based methods} train an attribute classifier using the attention mechanism \cite{bahdanau2014neural}, and consider words with attention weights higher than average as markers \cite{xu-etal-2018-unpaired}.
For the architecture of the classifier, \citet{zhang2018learning} use LSTM, and \citet{Sudhakar2019TransformingDR} use a BERT classifier, \myRed{where the BERT classifier has shown higher detection accuracy for the attribute markers.}

{\bf Fusion methods} combine the advantages of the above two methods. For example, \citet{wu2019mask} prioritize the attribute markers predicted by frequency-ratio methods, and use attention-based methods as an auxiliary back up. One use case is when
frequency-ratio methods fail to identify any attribute markers in a given sentence, they will use the attention-based methods as a secondary choice to generate attribute markers.
Another case is to reduce false positives. To reduce the number of attribute markers that are wrongly recognized, \citet{wu2019mask} set a threshold to filter out
low-quality attribute markers by frequency-ratio methods, and in cases where all attribute markers are deleted, they use the markers predicted by attention-based methods.

There are still remaining limitations of the previous methods, such as
imperfect accuracy of the attribute classifier, and unclear relation between attribute and attention scores. Hence, \citet{lee2020stable} propose word importance scoring, similar to what is used by \citet{jin2020bert} for adversarial paraphrasing, to measure how important a token is to the attribute by the difference in the attribute probability of the original sentence and that after deleting a token.



\subsubsection{Target Attribute Retriever}\label{sec:tar_attr_retr}
After deleting the attribute markers $\mathrm{Marker}_a(\bm{x})$ of the sentence $\bm{x}$ with attribute $a$, we need to find a counterpart attribute marker $\mathrm{Marker}_{a'}(\bm{x}')$ from another sentence $\bm{x}'$ carrying a different attribute ${a}'$. Denote the sentence template with all attribute markers deleted as $\mathrm{Template}(\bm{x}) \overset{\Delta}{=} \bm{x} \backslash \mathrm{Marker}_a(\bm{x})$. Similarly, the template of the sentence $\bm{x}'$ is $\mathrm{Template}(\bm{x}') \overset{\Delta}{=} \bm{x}' \backslash \mathrm{Marker}_{a'}(\bm{x}')$.
A common approach is to find the counterpart attribute marker by its context, because the templates of the original attribute and its counter attribute marker should be similar. Specifically, we first match a template $\mathrm{Template}(\bm{x})$ with the most similar template $\mathrm{Template}(\bm{x}')$ in the opposite attribute corpus, and then identify the attribute markers $\mathrm{Marker}_a(\bm{x})$ and $\mathrm{Marker}_{a'}(\bm{x}')$ as counterparts of each other.
To match templates with their counterparts, most previous works find the nearest neighbors by the cosine similarity of sentence embeddings. Commonly used sentence embeddings include TF-IDF as used in \cite{li-etal-2018-delete,Sudhakar2019TransformingDR}, averaged GloVe embedding distance used in \cite{li-etal-2018-delete,Sudhakar2019TransformingDR}, and Universal Sentence Encoder \cite{cer2018universal} used in \cite{Sudhakar2019TransformingDR}. Apart from sentence embeddings, \citet{tran2020towards} use Part-of-Speech templates to match several candidates in the opposite corpus, and conduct an exhaustive search to fill parts of the candidate sentences into the masked positions of the original attribute markers.

\subsubsection{Generation from Prototypes}\label{sec:gen_from_proto}

\citet{li-etal-2018-delete} and \citet{Sudhakar2019TransformingDR} feed the content-only sentence template and new attribute markers into a pretrained language model that rearranges them into a natural sentence.
This infilling process can naturally be achieved by a masked language model (MLM)~\citep{malmi-etal-2020-unsupervised}. For example, \citet{wu2019mask} use MLM of the template conditioned on the target attribute, and this MLM is trained on an additional attribute classification loss using the model output and a fixed pre-trained attribute classifier.
Since these generation practices are complicated,
\citet{madaan2020politeness} propose a simpler way. They skip Step 2 that explicitly retrieves attribute candidates, and, instead, directly learn a generation model that only takes attribute-masked sentences as inputs. This generation model is trained on data where the attribute-carrying sentences $\bm{x}$ are paired with their templates $\mathrm{Template}(\bm{x})$.
Training on the pairs of $( \mathrm{Template}(\bm{x}), \bm{x})$ constructed in the above way can make the model learn how to fill the masked sentence template with the target attribute $a$.

\subsection{Pseudo-Parallel Corpus Construction}\label{sec:pseudo_construct}
To provide more signals for training, it is also helpful to generate pseudo-parallel data for TST. Two major approaches are retrieval-based and generation-based methods.

\paragraph{Retrieval-Based Corpora Construction.}
One common way to construct pseudo-parallel data is through retrieval, namely extracting aligned sentence pairs from two mono-style corpora. \citet{jin2019imat} empirically observe that semantically similar sentences in the two mono-style corpora tend to be the attribute-transferred counterparts of each other. Hence, they construct the initial pseudo corpora by matching sentence pairs in the two attributed corpora according to the cosine similarity of pretrained sentence embeddings.
Formally, for each sentence $\bm{x}$, its pseudo counterpart $\widehat{\bm{x}'}$ is its most similar sentence in the other attribute corpus $\bm{X}'$, namely $\widehat{\bm{x}'}=\argmax_{\bm{x}'\in \bm{X}'} \mathrm{Similarity}(\bm{x}, \bm{x}')$. This approach is extended by \citet{nikolov2019large} who use large-scale hierarchical alignment to extract pseudo-parallel style transfer pairs. 
Such retrieval-based pseudo-parallel data construction is also useful for machine translation \cite{munteanu2005improving,uszkoreit2010large,marie2017efficient,gregoire2018extracting,ren2020retrieve}.

\paragraph{Generation-Based Corpora Construction.}
Another way is through generation, such as iterative back-translation (IBT) \cite{hoang2018iterative}. IBT is a widely used method in machine translation \cite{Artetxe2018unsupervised,lample2018unsupervised,lample2018phrase,dou2020dynamic} which adopts an iterative process to generate pseudo-parallel corpora. 

Before starting the iterative process, IBT needs to first initialize two style transfer models, $M_{a \rightarrow a'}$ which transfers from the attribute $a$ to the other attribute $a'$ and $M_{a' \rightarrow a}$ which transfers from $a'$ to $a$. 
Then, in each iteration, it executes the following steps:
\begin{enumerate}[align=left, 
leftmargin=*,
label=Step \arabic*), ref=\arabic*]
    \item Use the models to generate pseudo-parallel corpora. Specifically, $M_{a \rightarrow a'}(\bm{x})$ generates pseudo pairs $(\bm{x}, \widehat{\bm{x}'})$  for all $\bm{x} \in \bm{X}$,
    and $M_{a' \rightarrow a}(\bm{x}')$ generates pairs of $(\widehat{\bm{x}}, \bm{x}')$ for all $\bm{x}' \in \bm{X}'$;
    \label{list:ibt1}
    \item Re-train these two style transfer models on the datasets generated by \ref{list:ibt1}, i.e., re-train $M_{a \rightarrow a'}(\bm{x})$ on $(  \widehat{\bm{x}}, \bm{x}')$ pairs and $M_{a' \rightarrow a}(\bm{x}')$ on $(\widehat{\bm{x}'}, \bm{x})$ pairs.
    \label{list:ibt2}
\end{enumerate}

For Step~\ref{list:ibt1}, in order to generate the initial pseudo-parallel corpora, a simple baseline is to randomly initialize the two models $M_{a \rightarrow a'}$ and
$M_{a' \rightarrow a}$, and use them to translate the attribute of each sentence in $\bm{x} \in \bm{X}$ and $\bm{x}' \in \bm{X}'$. However, this simple initialization is subject to randomness and may not bootstrap well. Another way adopted by \citet{zhang2018style} borrows the idea from unsupervised machine translation \cite{lample2018unsupervised} that first learns an unsupervised word-to-word translation table between attribute $a$ and $a'$, and uses it to generate an initial
pseudo-parallel corpora. Based on such initial corpora, they train initial style transfer models and bootstrap the IBT process.
Another model, Iterative Matching and Translation (IMaT) \cite{jin2019imat}, does not learn the word translation table, and instead trains the initial style transfer models on a retrieval-based pseudo-parallel corpora introduced in the \textit{retrieval-based corpora construction} above.

For Step~\ref{list:ibt2}, during the iterative process, it is possible to encounter divergence, as there is no constraint to ensure that each iteration will produce better pseudo-parallel corpora than the previous iteration.
One way to enhance the convergence of IBT is to add additional losses. For example, \citet{zhang2018style} use the attribute classification loss ACO, as in Eq.~\eqref{eq:aco}, to check whether the generated sentence by back-translation fits the desired attribute according to a pre-trained style classifier.
Alternatively, IMaT \cite{jin2019imat} uses a checking mechanism instead of additional losses. At the end of each iteration, IMaT looks at all candidate pseudo-pairs of an original sentence, and uses Word Mover Distance \cite{kusner2015word} to select the sentence that has the desired attribute and is the closest to the original sentence.

\section{Research Agenda}\label{sec:discussion}
In this section, we will propose some potential  directions for future TST research, including expanding the scope of styles (Section~\ref{sec:expanding_scope_of_style}), improving the methodology (Section~\ref{sec:improving_method}), loosening the style-specific data assumptions (Section~\ref{sec:loosening}), and improving evaluation metrics (Section~\ref{sec:improving_eval}).

\subsection{Expanding the Scope of Styles}\label{sec:expanding_scope_of_style}

\paragraph{More Styles.}
Extending the list of styles for TST is one popular research direction. Existing research originally focused on styles such as simplification \cite{zhu2010monolingual}, formality \cite{sheikha2011formal}, and sentiment transfer \cite{shen2017style}, while the recent two years have seen a richer set of styles such as politeness  \citet{madaan2020politeness}, biasedness  \citet{pryzant2020automatically},   medical text simplification  \citet{cao-etal-2020-expertise}, and so on.

Such extension of styles is driven by the advancement of TST methods, and also various downstream needs, such as persona-based dialog generation, customized text rewriting applications, and moderation of online text.
Apart from the styles that have been researched as listed in Table~\ref{tab:task_type}, there are also many other new styles that can be interesting to conduct new research on, including but not limited to the following:
\begin{itemize}
    \item Factual-to-empathetic transfer, to improve counseling dialogs (after the first version of this survey in 2020, we gladly found that this direction has now a preliminary exploration by \citet{sharma2021towards});
    \item Non-native-to-native transfer (i.e., reformulating grammatical error correction with TST);
    \item Sentence disambiguation, to resolve nuance in text.
\end{itemize}
\paragraph{More Difficult Forms of Style.}
Another direction is to explore more complicated forms of styles. As covered by this survey, the early work on deep learning-based TST explores relatively simple styles, such as  verb tenses \cite{Hu2017TowardCG} and positive-vs-negative Yelp reviews \cite{shen2017style}. In these tasks, each data point is one sentence with a clear, categorized style, and the entire dataset is in the same domain. Moreover, the existing datasets can decouple style and style-independent contents relatively well.

We propose that TST can potentially be extended into the following settings:
\begin{itemize}
    \item Aspect-based style transfer (e.g., transferring the sentiment on one aspect but not the other aspects on aspect-based sentiment analysis data)
    \item Authorship transfer (which has tightly coupled style and content)
    \item Document-level style transfer (which includes discourse planning)
    \item Domain adaptive style transfer (which is preceded by \citet{li2019domain})
\end{itemize}

\paragraph{Style Interwoven with Semantics.}
In some cases, it can be difficult or impossible to separate attributes from meaning, namely the subject matter or the argument that the author wants to convey. One reason is that the subject that the author is going to write about can influence the choice of writing style. For example, science fiction writing can use the first person voice and fancy, flowery tone when describing a place. Another reason is that many stylistic devices such as allusion depend on content words. 

Currently, it is a simplification of the problem setting to limit it to scenarios where the attribute and semantics can be approximately separated. For evaluation, so far researchers have allowed the human judges decide the scores of transferred style strength and the content preservation.

In future work, it will be an interesting direction to address the more challenging scenarios where the style and semantics are interwoven.

\subsection{Improving the Methodology on Non-Parallel Data}\label{sec:improving_method}
Since the majority of TST research focuses on non-parallel data, we discuss below its strengths and limitations.

\subsubsection{Understanding the Strengths and Limitations of Existing Methods}
To come up with improvement directions for TST methods, it is important to first investigate the strengths and limitations of existing methods. We analyze the three major streams of approaches for unsupervised TST in Table~\ref{tab:pros_cons}, including their strengths, weaknesses, and future directions. 

\begin{table*}[h]
    \small
    \caption{The strengths ($+$), weaknesses ($-$), and improvement directions ($?$) of the three mainstreams of TST methods on non-parallel data.}
    \label{tab:pros_cons}
    \centering
    \begin{tabular}{p{0.18\textwidth}p{0.75\textwidth} }
    \toprule
    \textbf{Method} & \textbf{Strengths \& Weaknesses} \\ \hline
    \multirow{4}{*}{Disentanglement} & $+$ More profound in theoretical analysis, e.g., disentangled representation learning \\
    & $-$ Difficulties of training deep generative models (VAEs, GANs) for text \\
    & $-$ Hard to represent all styles as latent code \\
    & $-$ Computational cost rises with the number of styles to model \\
    \hline
    \multirow{9}{*}{Prototype Editing} & $+$ High BLEU scores due to large word preservation \\
    & $-$ Attribute marker detection step can fail if the style and semantics are confounded \\
    & $-$ The step target attribute retrieval by templates can fail if there are large rewrites for styles, e.g., Shakespearean English vs. modern English \\
    & $-$ Target attribute retrieval step has large complexity (quadratic to the number of sentences) \\
    & $-$ Large computational cost if there are many styles, each of which needs a pre-trained LM for the generation step \\
    & $?$ \space Future work can enable matchings for syntactic variation \\
    & $?$ \space Future work can use grammatical error correction to post-edit the output \\
    \hline
    \multirow{9}{*}{\parbox{\linewidth}{Pseudo-Parallel Corpus Construction}} & $+$~Performance can approximate supervised model performance, if the pseudo-parallel data are of good quality \\
    & $-$ May fail for small corpora \\
    & $-$ May fail if the mono-style corpora do not have many samples with similar contents \\
    & $-$ For IBT, divergence is possible, and sometimes needs special designs to prevent it \\
    & $-$ For IBT, time complexity is high (due to iterative pseudo data generation) \\
    & $?$ \space Improve the convergence of the IBT \\
    \bottomrule
    \end{tabular}
\end{table*}
\paragraph{Challenges for Disentanglement.}
Theoretically, although disentanglement is impossible without inductive biases or other forms of supervision \cite{locatello2019challenging}, disentanglement is achievable with some weak signals, such as only knowing how many factors have changed, but not which ones \cite{locatello2020weakly}. 

In practice, some big challenges for disentanglement-based methods include, for example, the difficulty to train deep text generative models such as VAEs and GANs. Also, it is not easy to represent all styles as latent code. Moreover, if targeting multiple styles, the computational complexity linearly increases with the number of styles to model.

\paragraph{Challenges for Prototype Editing.}
Prototype-editing approaches usually result in relatively high BLEU scores, partly because the output text largely overlaps with the input text. This line of methods is likely to perform well on tasks such as sentiment modification, for which it is easy to identify ``attribute markers,'' and the input and output sentences share an attribute-independent template.

However, prototype editing cannot be applied  to all types of style transfer tasks. The first step, attribute marker retrieval, might not work if the datasets have confounded style and contents, because they may lead to wrong extraction of attribute markers, such as some content words or artifacts which can also be used to distinguish the style-specific data.

The second step, target attribute retrieval by templates, will fail if there is too little word overlap between a sentence and its counterpart carrying another style.
An example is the TST task to ``Shakespearize'' modern English. There is little lexical overlap between a Shakespearean sentence written in early modern English and its corresponding modern English expression. In such cases, the retrieval step is likely to fail, because there is a large number of rewrites between the two styles, and the template might be almost hollow. Moreover, this step is also computationally expensive, if there are a large number of sentences in the data (e.g., all Wikipedia text), since this step needs to calculate the pair-wise similarity among all available sentences across style-specific corpora.

The third step, generation from prototype, requires a separate pretrained LM for each style corpus. When there are multiple styles of interest (e.g., multiple persona), this will induce a large computational cost.

The last limitation of prototype editing is that it amplifies the intrinsic problem of using BLEU to evaluate TST (Problem~\ref{list:prob1}, namely the fact that simply copying the input can result in a high BLEU score) as introduced in Section~\ref{sec:auto_eval}).
For the retrieval-based method, some can argue that there is some performance gain because this method in practice copies more expressions in the input sentence than other lines of methods.

As future study, there can be many interesting directions to explore, for example, investigating the performance of existing prototype editing models under a challenging dataset that reveals the above shortcomings, proposing new models to improve this line of approaches, and better evaluation methods for prototype editing models.

\paragraph{Challenges for Pseudo-Parallel Corpus Construction.}

The method to construct pseudo-parallel data can be effective, especially when the pseudo-parallel corpora resemble supervised data. The challenge is that this approach may not work if the non-parallel corpora do not have enough samples that can be matched to create the pseudo-parallel corpora, or when the IBT cannot bootstrap well or fails to converge. The time complexity for training IBT is also very high because it needs to iteratively generate pseudo-parallel corpus and re-train models. Interesting future directions can be reducing the computational cost, designing more effective bootstrapping, and improving the convergence of IBT.

\subsubsection{Understanding the Evolution from Traditional NLG to Deep Learning Methods}
Despite the exciting methodological revolution led by deep learning recently, we are also interested in the merging point of traditional computational linguistics and the deep learning techniques \cite{henderson2020unstoppable}. Specific to the context of TST, we will introduce the traditional NLG framework, and its impact on the current TST approaches, especially the prototype editing method.

\paragraph{Traditional NLG Framework.}
The traditional NLG framework stages sentence generation into the following steps \cite{DBLP:journals/nle/ReiterD97}:
\begin{enumerate}[nolistsep]
\item Content determination (not applicable)
\item Discourse planning (not applicable)
\item Sentence aggregation
\item Lexicalization
\item Referring expression generation
\item Linguistic realization
\end{enumerate}
The first two steps, content determination and discourse planning are not applicable to most datasets because the current focus of TST is sentence-level and not discourse-level.

Among  Steps 3 to 6, 
\textit{sentence aggregation} groups necessary information into a single sentence, \textit{lexicalization} chooses the right word to express the concepts generated by sentence aggregation, \textit{referring expression generation} produces surface linguistic forms for domain entities,
and \textit{linguistic realization} edits the text so that it conforms to grammar, including syntax, morphology, and orthography. This framework is widely applied to NLG tasks \cite[e.g.,][]{zue2000conversational,mani2001automatic,mctear2002spoken,gatt2009simplenlg,androutsopoulos2010survey}.

\paragraph{Re-Viewing Prototype-Based TST.}
Among the approaches introduced so far, the most relevant for the traditional NLG is the prototype-based text editing, which has been introduced in Section~\ref{sec:proto_nlg}.

Using the language of the traditional NLG framework, the prototype-based techniques can be viewed as a combination of \textit{sentence aggregation}, \textit{lexicalization}, and \textit{linguistic realization}.
Specifically, prototype-based techniques first prepare an attribute-free sentence template, and supply it with candidate attribute markers that carry the desired attribute, both of which are \textit{sentence aggregation}. Then, using language models to infill the prototype with the correct expressions corresponds to \textit{lexicalization} and \textit{linguistic realization}.
Note that the existing TST systems do not explicitly deal with \textit{referring expression generation} (e.g., generating co-references), leaving it to be handled by language models.

\paragraph{Meeting Point of Traditional and New Methods.}
Viewing prototype-based editing as a merging point where traditional, controllable framework meets deep learning models, we can see that it takes advantage of the powerful deep learning models and the interpretable pipeline of the traditional NLG.
There are several advantages in merging the traditional NLG with the deep learning models. First, sentence planning-like steps make the generated contents more controllable. For example, the template of the original sentence is saved, and the counterpart attributes can also be explicitly retrieved, as a preparation for the final rewriting. Such a controllable, white-box approach can be easy to tune, debug, and improve. The accuracy of attribute marker extraction, for example, is constantly improving across literature \cite{Sudhakar2019TransformingDR} and different ways to extract attribute markers can be easily fused \cite{wu2019mask}.
Second, sentence planning-like steps ensure the truthfulness of information. As most content words are kept and no additional information is hallucinated by the black-box neural networks, we can better ensure that the information of the attribute-transferred output is consistent with the original input.


\subsubsection{Inspiration from Tasks with Similar Nature} \label{sec:related_tasks}

An additional perspective that can inspire new methodological innovation is insights from other tasks that share a similar nature as TST. We will introduce in this section several closely-related tasks, including machine translation, image style transfer, style-conditioned language modeling, counterfactual story rewriting, contrastive text generation, and prototype-based text editing.

\paragraph{Machine Translation.}
The problem settings of machine translation and text style transfer share much in common: the source and target language in machine translation is analogous to the original and desired attribute, $a$ and $a'$, respectively. The major difference is that in NMT, the source and target corpora are in completely different languages, which have almost disjoint word vocabulary, whereas in text style transfer, the input and output are in the same language, and the model is usually encouraged to copy most content words from input such as the BoW loss introduced in Section~\ref{sec:aim2}. Some TST works have been inspired by MT, such as the pseudo-parallel construction \cite{nikolov2019large,zhang2018style}, and in the future there may be more interesting intersections.

\myRed{\paragraph{Data-to-Text Generation.}
Data-to-text generation is another potential domain that can draw inspiration from and to TST. The data-to-text generation task is to generate textual descriptions from structured data such as tables \cite{wiseman2017challenges,parikh2020totto}, meaning representations \cite{,novikova2017e2e}, or Resource Description Framework (RDF) triples \cite{gardent2017webnlg,ferreira20202020}.
With the recent rise of pretrained seq2seq models for transfer learning \cite{raffel2020exploring}, it is common to formulate data-to-text as a seq2seq task by serializing the structured data into a sequence \cite{kale2020text,ribeiro2020investigating,p22020guo}. Then data-to-text generation can be seen as a special form of TST from structured information to text. This potential connection has not yet been investigated but worth exploring.}

\paragraph{Neural Style Transfer.}

Neural style transfer first originates in image style transfer \cite{gatys2016image}, and its disentanglement ideas inspired some early TST researcg \cite{shen2017style}. The difference between image style transfer and TST is that, for images, it is feasible to disentangle the explicit representation of the image texture as the gram matrix of image neural feature vectors, but for text, styles do not have such an explicit representation, but more abstract attributes. Besides this difference, many other aspects of style transfer research can have shared nature. Note that there are style transfer works across different modalities, including images \cite{gatys2016image,zhu2017unpaired,chen2017stylebank}, text, voice \cite{gao2018voice,qian2019autovc,yuan2021improving}, handwriting \cite{azadi2018multi,zhang2013writer}, and videos \cite{ruder2016artistic,chen2017coherent}.
Many new advances in one style transfer field can inspire another style transfer field. For example, image style transfer has been used as a way for data augmentation \cite{zheng2019stada,jackson2019style} and adversarial attack \cite{xu2020towardsfeature}, but TST has not yet been applied for such usage.


\paragraph{Style-Conditioned Language Modeling.}
Different from language modeling that learns how to generate general natural language text, conditional language modeling learns how to generate text given a condition, such as some context, or a control code \cite{pfaff1979constraints,poplack2000sometimes}. Recent advances of conditional language models \cite{keskar2019ctrl,dathathri2020plug} also include text generation conditioned on a style token, such as positive or negative. 
Possible conditions include author style \cite{Syed2020AdaptingLM}, speaker identity, persona and emotion \cite{li2016persona}, genre, attributes derived from text, topics, and sentiment \cite{ficler2017controlling}.
They are currently limited to a small set of pre-defined ``condition'' tokens and can only generate from scratch a sentence, but not yet able to be conditioned on an original sentence for style rewriting. The interesting finding in this research direction is that it can make good use of a pretrained LM and just do some light-weight inference techniques to generate style-conditioned text, so perhaps such approaches can inspire future TST methods and reduce the carbon footprints of training TST models from scratch.

\paragraph{Counterfactual Story Rewriting.}
Counterfactual story rewriting aims to learn a new event sequence in the presence of a perturbation of a previous event (i.e., counterfactual condition) \cite{goodman1947problem,starr2019counterfactuals}. \citet{qin2019counterfactual} propose the first dataset, each sample of which takes an originally five-sentence story, and changes the event in the second sentence to a new, counterfactual event. The task is to generate the last three sentences of the story based on the newly altered second sentence that initiates the story. The criteria of the counterfactual story rewriting include relevance with the first two sentences, and minimal edits from the original story ending. This line of research is relatively difficult to directly apply to TST, because its motivation and dataset nature is different from the general text style transfer, and more importantly, this task is not conditioned on a predefined categorized style token, but the free-form textual story beginning.

\paragraph{Contrastive Text Generation.}
As neural network-based NLP models more easily learn spurious statistical correlations in the data rather than achieve robust understanding \cite{jia2017adversarial}, there are recent works to construct auxillary datasets composed of near-misses of the original data. For example, \citet{matt2020evaluating} ask crowdsource workers to rewrite the input of the task with minimal changes but matching a different target label. To alleviate expensive human labor, \citet{xing2020tasty} develop an automatic text editing approach to generate contrast set for aspect-based sentiment analysis. The difference between contrastive text generation and text style transfer is that the former does not require content preservation but mainly aims to construct a slightly textually different input that can result in a change of the ground-truth output, to test the model robustness. So the two tasks are not completely the same, although they have some intersections that might inspire future work, such as aspect-based style transfer suggested in Section~\ref{sec:expanding_scope_of_style}.

\paragraph{Prototype-Based Text Editing.}
Prototype editing is not unique in TST, but also widely used in other NLP tasks. Knowing the new advances in prototype editing for other tasks can potentially inspire new method innovations in TST. \citet{guu2018generating} first proposes the protype editing approach to improve LM by first sampling a lexically similar sentence prototype and then editing it using variational encoder and decoders. This prototype-and-then-edit approach can also be seen in summarization \cite{wang2019biset}, machine translation \cite{cao2018encoding,wu2019extract,gu2018search,zhang2018guiding,bulte2019neural}, conversation generation \cite{weston2018retrieve,cai2018skeleton}, code generation \cite{hashimoto2018retrieve}, and question answering \cite{lewis2020retrieval}. As an extension to the retrieve and edit steps, \citet{hossain2020simple} use an ensemble approach to retrieve a set of relevant prototypes, edit, and finally rerank to pick the best output for machine translation. Such extension can also be potentially applied to text style transfer.


\subsection{Loosening the Style-Specific Dataset Assumptions}\label{sec:loosening}
A common assumption for most deep learning-based TST works, as mentioned in Section~\ref{sec:assumptions}, is the availability of style-specific corpora for each style of interest, either parallel or non-parallel.
This assumption can potentially be loosened in two ways.
\myRed{\paragraph{Linguistic Styles with No Matched Data.}
Since there are various concerns raised by the data-driven definition of style as described in Section~\ref{sec:define}, a potentially good research direction is to bring back the linguistic definition of style, and thus remove some of the concerns associated with large datasets. Several methods can be a potential fit for this: prompt design \cite{li2020prefix,qin2021learning,scao2021how} that passes a prompt to GPT \cite{radford2019language,brown2020language} to obtain a style-transferred text; style-specific template design; or use templates to first generate synthetic data and make models learn from the synthetic data. Prompt design is not yet investigated as a direction for TST research, but it is an interesting direction to explore.}

\myRed{\paragraph{Distinguishing Styles from a Mixed Corpus.}
It might also be possible to distinguish styles direction from a mixed corpus with no style labels.
For example, \citet{riley2021textsettr} learn a style vector space from text;} \citet{xu2020variational} use unsupervised representation learning to separate the style and contents from a mixed corpus of unspecified styles; \citet{guo2020fork} use cycle training with a conditional variational auto-encoder to unsupervisedly learn to express the same semantics through different styles. Theoretically, although disentanglement is impossible without inductive biases or other forms of supervision \cite{locatello2019challenging}, disentanglement is achievable with some weak signals, such as only knowing how many factors have changed, but not which ones \cite{locatello2020weakly}. A more advanced direction can be emergent styles \cite{kang2020incorporating}, since styles can be evolving, for example across dialog turns.

\subsection{Improving Evaluation Metrics}\label{sec:improving_eval}
There has been a lot of attention to the problems of evaluation metrics of TST and potential improvements \cite{pang2018unsupervised,tikhonov2018wrong,mir-etal-2019-evaluating,fu2019rethinking,pang2020daunting,yamshchikov2020style,jafaritazehjani2020style}. 
Recently, \citet{gehrmann2021thegem} has proposed a new framework which is a live environment to evaluate NLG in a principled and reproducible manner.
Apart from the existing scoring methods, future works can also make use of linguistic rules such as a checklist to evaluate what capabilities the TST model has achieved. For example, there can be a checklist for formality transfer according to existing style guidelines such as the APA style guide \cite{american1983publication}. Such a checklist-based evaluation can make the performance of black-box deep learning models more interpretable, and also allow for more insightful error analysis.

\section{Expanding the Impact of TST}\label{sec:impact}
In this last section of this survey, we highlight several directions to expand the impact of TST. First, TST can be used to help other NLP tasks such as paraphrasing, data augmentation, and adversarial robustness probing (Section~\ref{sec:applications}). Moreover, many specialized downstream tasks can be achieved with the help of TST, such as persona-consistent dialog generation, attractive headline generation, style-specific machine translation, and anonymization (Section~\ref{sec:downstream}). Last but not the least, we overview the ethical impacts that are important to take into consideration for future development of TST (Section~\ref{sec:ethics}).

\subsection{Connecting TST to More NLP Tasks}\label{sec:applications}
Text style transfer can be applied to other important NLP tasks, such as paraphrase generation, data augmentation, and adversarial robustness probing.

\paragraph{Paraphrase Generation.}
Paraphrase generation is to express the same information in alternative ways \cite{madnani2010generating}. The nature of paraphrasing shares a lot in common with TST, which is to transfer the style of text while preserving the content. \myRed{One of the common ways of paraphrasing is syntactic variation, such as ``X wrote Y.'', ``Y was written by X.'' and ``X is the writer of Y.'' \cite{androutsopoulos2010survey}. Besides syntactic variation, it also makes sense to include stylistic variation as a form of paraphrases, which means that the linguistic style transfer (not the content preference transfer in Table~\ref{tab:task_type}) can be regarded as a subset of paraphrasing. The caution here is that if the paraphrasing is for a downstream task, researchers should first check if the downstream task is compatible with the used styles. For example, dialog generation may be sensitive to all linguistic styles, whereas summarization can allow linguistic style-varied paraphrases in the dataset.}

There are three implications of this connection of TST and paraphrase generation. First, many trained TST models can be borrowed for paraphrasing, such as formality transfer and simplification. A second connection is that the method innovations proposed in the two fields can inspire each other. For example, \citet{krishna2020reformulating} formulate style transfer as a paraphrasing task. Thirdly, the evaluation metrics of the two tasks can also inspire each other. For example, \citet{yamshchikov2020style} associate the semantic similarity metrics for two tasks.

\paragraph{Data Augmentation.}
Data augmentation generates text similar to the existing training data so that the model can have larger training data. TST is a good method for data augmentation because TST can produce text with different styles but the same meaning. Image style transfer has already been used for data augmentation \cite{zheng2019stada,jackson2019style}, so it can be interesting to see future works to also apply text style transfer for data augmentation. 

\paragraph{Adversarial Robustness Probing.}
Another use of style transferred text is adversarial robustness probing. For example, styles that are task-agnostic can be used for general adversarial attack (e.g., politeness transfer to probe sentiment classification robustness) \cite{jin2020bert}, while the styles that can change the task output can be used to construct contrast sets (e.g., sentiment transfer to probe sentiment classification robustness) \cite{xing2020tasty}.
\citet{xu2020towardsfeature} applies image style transfer to adversarial attack, and future research can also explore the use of TST in the two ways suggested above.

\subsection{Connecting TST to More Specialized Applications}\label{sec:downstream}

TST can be applied not only to other NLP tasks as introduced in the previous section, but also be helpful for specialized downstream applications. In practice, when applying NLP models, it is important to customize for some specific needs, such as generating dialog with a consistent persona, writing headlines that are attractive and engaging,
making machine translation models adapt to different styles, and anonymizing the user identity by obfuscating the style.

\paragraph{Persona-Consistent Dialog Generation.}
A useful downstream application of TST is persona-consistent dialog generation \cite{li2016persona,zhang2018personalizing,shuster2020image}. Since conversational agents directly interact with users, there is a strong demand for human-like dialog generation. Previously, this is done by encoding speaker traits into a vector and the conversation is then conditioned on this vector \cite{li2016persona}. As future work, text style transfer can also be used as part of the pipeline of persona-based dialog generation, where the persona can be categorized into distinctive style types, and then the generated text can be post-processed by a style transfer model.

\paragraph{Attractive Headline Generation.}
In journalism writing, it is crucial to generate engaging headlines.  \citet{jin-etal-2020-hooks} first use TST to generate eye-catchy headlines with three different styles, humorous, romantic, and clickbaity styles. \citet{li2020stylecontent} follow this direction and propose a disentanglement-based model to generate attractive headlines for Chinese news.

\paragraph{Style-Specific Machine Translation.}
In machine translation, it is useful to have an additional control of the style for the translated text. Commonly used styles for TST in machine translation are  politeness \cite{sennrich2016controlling} and formality \cite{niu2017study,Wu2020ADF}. For example, \citet{Wu2020ADF} translates from informal Chinese to formal English.

\paragraph{Anonymization.}
TST can also be used for anonymization, which is an important way to protect user privacy, especially since there are ongoing heated discussions of ethics in the artificial intelligence community. Many concerns have been raised towards the discriminative task of author profiling, which can mine the demographic identities of the author of a writing, even including privacy-invading properties such as gender and age \cite{schler2006effects}. As a potential solution, TST can be applied to alter the text and obfuscate the real identity of the users \cite{reddy2016obfuscating,grondahl2020effective}.

\subsection{Considering Ethical Impacts of TST}\label{sec:ethics}

Recently, there is more and more attention being paid to the ethical concerns associated with AI research. We discuss in the following two ethical considerations: (1) social impact of TST applications, and (2) data privacy problem of text style transfer.

Fields that involve human subjects or direct application to humans work under a set of core principles and guidelines \cite{beauchamp2001principles}. Before initiating a research project, responsible research bodies use these principles as a ruler to judge whether the research is ethically correct to start. NLP reserch and applications, including TST, that directly involve human users, is regulated under a central regulatory board, Institutional Review Board (IRB). We also provide several guidelines below to avoid ethical misconduct in future publications on text style transfer.

\subsubsection{Social Impact of TST Applications}

Technologies can have unintended negative consequences \cite{hovy2016social}. For example, TST can facilitate the automation of intelligent assistants with designed attributes, but can also be used to create fake text or fraud.

Thus, inventors of a technology should beware how other people very probably adopt this technology for their own incentives. 
For TST, since it has a wide range of subtasks and applications, we examine each of them with the following two questions:
\begin{itemize}
    \item Who will benefit from such a technology?
    \item Who will be harmed by such a technology?
\end{itemize}
Although many ethical issues are debatable, we try to categorize the text attribute tasks into three ethical levels: beneficial, neutral, and tasks that can be obvious double-sided swords.

\paragraph{Beneficial.}
An important direction of NLP for social good is to fight against abusive online text. Text style transfer can serve as a very helpful tool as it can be used to transfer malicious text to normal language. Shades of abusive language include hate speech, offensive language, sexist and racist language, aggression, profanity, cyberbullying, harassment, trolling, and toxic language \cite{waseem2017understanding}. There are also other negative text such as propaganda \cite{bernays2005propaganda,carey1997taking}, and others. It is widely known that malicious text is harmful to people. For example, research shows that cyberbullying victims tend to have more stress and suicidal ideation \cite{kowalski2014bullying}, and also detachment from family and offline victimization \cite{oksanen2014exposure}. There are more and more efforts put into combating toxic language, such as 30K content moderators that Facebook and Instagram employ   \cite{wired2019twitter}. Therefore, the automatic malicious-to-normal language transfer can be a helpful intelligent assistant to address such needs. Apart from purifying malicious text on social media, it can also be used on social chatbots to make sure there are no bad contents in language they generate~\citep{roller2020recipes}.

\paragraph{Neutral.} Most text style transfer tasks are neutral. For example, informal-to-formal transfer can be used as a writing assistant to help make writings more professional, and formal-to-informal transfer can tune the tone of bots to be more casual. Most applications to customized the persona of bots are also neutral with regard to their societal impact.


\paragraph{Double-Sided Sword.}
Besides positive and neutral applications, there are, unfortunately, several text style transfer tasks that are double-sided swords.
For example, one of the most popular TST tasks, sentiment modification, although it can be used to change intelligent assistants or robots from a negative to positive mood (which is unlikely to harm any parties), the vast majority of papers applies this technology to manipulate the polarity of reviews, such as Yelp \cite{shen2017style} and Amazon reviews \cite{he2016ups}.
This leads to a setting where a negative restaurant review is changed to a positive comment, or vice versa, with debatable ethics. Such a technique can be used
as a cheating method for the commercial body to polish its reviews, or harm the reputation of their competitors. Once this technology is used, it will automatically manipulate the online text to contain polarity that the model owner desires. Hence, we suggest the research community raise serious concern against the review sentiment modification task.

Another task, political slant transfer, may induce concerns within some specific context. For example, social bots (i.e., autonomous bots on social media, such as Twitter bots and Facebook bots) are a big problem in the U.S., even playing a significant role in the 2016 United States presidential election \cite{bessi2016social,shao2018spread}.
It is reported that at least 400,000 bots were responsible for about 19\% of the total Tweets. Social bots usually target to advocate certain ideas, supporting campaigns, or aggregating other sources either by acting as a "follower" and/or gathering followers itself. So the political slant transfer task, which transfers the tone and content between republican comments and democratic ones, are highly sensitive and may face the risk of being used on social bots to manipulate political views of the mass.

Some more arguable ones are male-to-female tone transfer, which can be potentially used for identity deception. The cheater can create an online account and pretend to be an attractive young lady. There is also the reversed direction (female-to-male tone transfer) which can be used for applications  such as authorship obfuscation \cite{shetty2018author} which anonymizes the author attributes by hiding the gender of a female author by re-synthesizing the text to use male.


\subsubsection{Data Privacy Issues for TST}
Another ethical concern is the use of data in the research practice. Researchers should not overmine user data, such as the demographic identities. Such data privacy widely exists in the data science community as a whole, and there have been many ethical discussions \cite{tse2015bibliometric,russell2015research}.

Although the TST task needs data containing \textit{some} attributes along with the text content. While it is acceptable to use ratings of reviews that are classified as positive or negative, but the user attributes are sensitive, including the gender of the user's account \cite{prabhumoye-etal-2018-style}, and age \cite{Lample2019MultipleAttributeTR}. \myRed{The collection and potential use of such sensitive user attributes can have implications that need to be carefully considered.}

\section{Conclusion}
\label{sec:conclusion}
This paper presented a comprehensive review of text style transfer with deep learning methods. We have surveyed recent research efforts in TST and developed schemes to categorize and distill the existing literature. This survey has covered the task formulation, evaluation metrics, and methods on parallel and non-parallel data. We also discussed several important topics in the research agenda of TST, and how to expand the impact of TST to other tasks and applications, including ethical considerations. This survey provides a reference for future researchers working on TST.

\section*{Acknowledgement}
We thank Qipeng Guo for his insightful discussions and the anonymous reviewers for their constructive suggestions.

\starttwocolumn
\bibliography{bibliography}

\begin{thebibliography}{239}
\expandafter\ifx\csname natexlab\endcsname\relax\def\natexlab#1{#1}\fi

\bibitem[{{American {P}sychological
  {A}ssociation}(1983)}]{american1983publication}
{American {P}sychological {A}ssociation}. 1983.
\newblock \emph{Publication manual}.
\newblock American Psychological Association Washington, DC.

\bibitem[{Androutsopoulos and Malakasiotis(2010)}]{androutsopoulos2010survey}
Androutsopoulos, Ion and Prodromos Malakasiotis. 2010.
\newblock A survey of paraphrasing and textual entailment methods.
\newblock \emph{Journal of Artificial Intelligence Research}, 38:135--187.

\bibitem[{Argamon et~al.(2003)Argamon, Koppel, Fine, and
  Shimoni}]{argamon2003gender}
Argamon, Shlomo, Moshe Koppel, Jonathan Fine, and Anat~Rachel Shimoni. 2003.
\newblock Gender, genre, and writing style in formal written texts.
\newblock \emph{Text \& Talk}, 23(3):321--346.

\bibitem[{Artetxe et~al.(2018)Artetxe, Labaka, Agirre, and
  Cho}]{Artetxe2018unsupervised}
Artetxe, Mikel, Gorka Labaka, Eneko Agirre, and Kyunghyun Cho. 2018.
\newblock Unsupervised neural machine translation.
\newblock In \emph{6th International Conference on Learning Representations,
  {ICLR} 2018, Vancouver, BC, Canada, April 30 - May 3, 2018, Conference Track
  Proceedings}, OpenReview.net.

\bibitem[{Azadi et~al.(2018)Azadi, Fisher, Kim, Wang, Shechtman, and
  Darrell}]{azadi2018multi}
Azadi, Samaneh, Matthew Fisher, Vladimir~G. Kim, Zhaowen Wang, Eli Shechtman,
  and Trevor Darrell. 2018.
\newblock Multi-content {GAN} for few-shot font style transfer.
\newblock In \emph{2018 {IEEE} Conference on Computer Vision and Pattern
  Recognition, {CVPR} 2018, Salt Lake City, UT, USA, June 18-22, 2018}, pages
  7564--7573, {IEEE} Computer Society.

\bibitem[{Bahdanau, Cho, and Bengio(2015)}]{bahdanau2014neural}
Bahdanau, Dzmitry, Kyunghyun Cho, and Yoshua Bengio. 2015.
\newblock Neural machine translation by jointly learning to align and
  translate.
\newblock In \emph{3rd International Conference on Learning Representations,
  {ICLR} 2015, San Diego, CA, USA, May 7-9, 2015, Conference Track
  Proceedings}.

\bibitem[{Banerjee and Lavie(2005)}]{banerjee-lavie-2005-meteor}
Banerjee, Satanjeev and Alon Lavie. 2005.
\newblock {METEOR}: {A}n automatic metric for {MT} evaluation with improved
  correlation with human judgments.
\newblock In \emph{Proceedings of the {ACL} Workshop on Intrinsic and Extrinsic
  Evaluation Measures for Machine Translation and/or Summarization}, pages
  65--72, Association for Computational Linguistics, Ann Arbor, Michigan.

\bibitem[{Bao et~al.(2019)Bao, Zhou, Huang, Li, Mou, Vechtomova, Dai, and
  Chen}]{bao-etal-2019-generating}
Bao, Yu, Hao Zhou, Shujian Huang, Lei Li, Lili Mou, Olga Vechtomova, Xin-yu
  Dai, and Jiajun Chen. 2019.
\newblock Generating sentences from disentangled syntactic and semantic spaces.
\newblock In \emph{Proceedings of the 57th Annual Meeting of the Association
  for Computational Linguistics}, pages 6008--6019, Association for
  Computational Linguistics, Florence, Italy.

\bibitem[{Bateman and Paris(1989)}]{bateman1989phrasing}
Bateman, John~A and Cecile Paris. 1989.
\newblock Phrasing a text in terms the user can understand.
\newblock In \emph{IJCAI}, pages 1511--1517.

\bibitem[{Beauchamp, Childress et~al.(2001)}]{beauchamp2001principles}
Beauchamp, Tom~L, James~F Childress, et~al. 2001.
\newblock \emph{Principles of biomedical ethics}.
\newblock Oxford University Press, USA.

\bibitem[{Belz(2008)}]{belz2008automatic}
Belz, Anja. 2008.
\newblock Automatic generation of weather forecast texts using comprehensive
  probabilistic generation-space models.
\newblock \emph{Natural Language Engineering}, 14(4):431--455.

\bibitem[{Belz et~al.(2020)Belz, Agarwal, Shimorina, and
  Reiter}]{belz2020reprogen}
Belz, Anya, Shubham Agarwal, Anastasia Shimorina, and Ehud Reiter. 2020.
\newblock {R}epro{G}en: {P}roposal for a shared task on reproducibility of
  human evaluations in {NLG}.
\newblock In \emph{Proceedings of the 13th International Conference on Natural
  Language Generation}, pages 232--236, Association for Computational
  Linguistics, Dublin, Ireland.

\bibitem[{den Bercken, Sips, and Lofi(2019)}]{bercken2019evaluating}
den Bercken, Laurens~Van, Robert{-}Jan Sips, and Christoph Lofi. 2019.
\newblock Evaluating neural text simplification in the medical domain.
\newblock In \emph{The World Wide Web Conference, {WWW} 2019, San Francisco,
  CA, USA, May 13-17, 2019}, pages 3286--3292, {ACM}.

\bibitem[{Bernays(2005)}]{bernays2005propaganda}
Bernays, Edward~L. 2005.
\newblock \emph{Propaganda}.
\newblock Ig publishing.

\bibitem[{Bessi and Ferrara(2016)}]{bessi2016social}
Bessi, Alessandro and Emilio Ferrara. 2016.
\newblock Social bots distort the 2016 us presidential election online
  discussion.
\newblock \emph{First Monday}, 21(11-7).

\bibitem[{Boulis and Ostendorf(2005)}]{knight2005quantitative}
Boulis, Constantinos and Mari Ostendorf. 2005.
\newblock A quantitative analysis of lexical differences between genders in
  telephone conversations.
\newblock In \emph{{ACL} 2005, 43rd Annual Meeting of the Association for
  Computational Linguistics, Proceedings of the Conference, 25-30 June 2005,
  University of Michigan, {USA}}, pages 435--442, The Association for Computer
  Linguistics.

\bibitem[{Briakou et~al.(2021{\natexlab{a}})Briakou, Agrawal, Zhang, Tetreault,
  and Carpuat}]{briakou2021review}
Briakou, Eleftheria, Sweta Agrawal, Ke~Zhang, Joel Tetreault, and Marine
  Carpuat. 2021{\natexlab{a}}.
\newblock A review of human evaluation for style transfer.
\newblock In \emph{Proceedings of the 1st Workshop on Natural Language
  Generation, Evaluation, and Metrics (GEM 2021)}, pages 58--67, Association
  for Computational Linguistics, Online.

\bibitem[{Briakou et~al.(2021{\natexlab{b}})Briakou, Lu, Zhang, and
  Tetreault}]{briakou2021ola}
Briakou, Eleftheria, Di~Lu, Ke~Zhang, and Joel Tetreault. 2021{\natexlab{b}}.
\newblock Ol{\'a}, bonjour, salve! {XFORMAL}: {A} benchmark for multilingual
  formality style transfer.
\newblock In \emph{Proceedings of the 2021 Conference of the North American
  Chapter of the Association for Computational Linguistics: Human Language
  Technologies}, pages 3199--3216, Association for Computational Linguistics,
  Online.

\bibitem[{Brown et~al.(2020)Brown, Mann, Ryder, Subbiah, Kaplan, Dhariwal,
  Neelakantan, Shyam, Sastry, Askell, Agarwal, Herbert{-}Voss, Krueger,
  Henighan, Child, Ramesh, Ziegler, Wu, Winter, Hesse, Chen, Sigler, Litwin,
  Gray, Chess, Clark, Berner, McCandlish, Radford, Sutskever, and
  Amodei}]{brown2020language}
Brown, Tom~B., Benjamin Mann, Nick Ryder, Melanie Subbiah, Jared Kaplan,
  Prafulla Dhariwal, Arvind Neelakantan, Pranav Shyam, Girish Sastry, Amanda
  Askell, Sandhini Agarwal, Ariel Herbert{-}Voss, Gretchen Krueger, Tom
  Henighan, Rewon Child, Aditya Ramesh, Daniel~M. Ziegler, Jeffrey Wu, Clemens
  Winter, Christopher Hesse, Mark Chen, Eric Sigler, Mateusz Litwin, Scott
  Gray, Benjamin Chess, Jack Clark, Christopher Berner, Sam McCandlish, Alec
  Radford, Ilya Sutskever, and Dario Amodei. 2020.
\newblock Language models are few-shot learners.
\newblock In \emph{Advances in Neural Information Processing Systems 33: Annual
  Conference on Neural Information Processing Systems 2020, NeurIPS 2020,
  December 6-12, 2020, virtual}.

\bibitem[{Bult{\'{e}} and Tezcan(2019)}]{bulte2019neural}
Bult{\'{e}}, Bram and Arda Tezcan. 2019.
\newblock Neural fuzzy repair: {I}ntegrating fuzzy matches into neural machine
  translation.
\newblock In \emph{Proceedings of the 57th Conference of the Association for
  Computational Linguistics, {ACL} 2019, Florence, Italy, July 28- August 2,
  2019, Volume 1: Long Papers}, pages 1800--1809, Association for Computational
  Linguistics.

\bibitem[{Cai et~al.(2019)Cai, Wang, Bi, Tu, Liu, Lam, and
  Shi}]{cai2018skeleton}
Cai, Deng, Yan Wang, Wei Bi, Zhaopeng Tu, Xiaojiang Liu, Wai Lam, and Shuming
  Shi. 2019.
\newblock Skeleton-to-response: {D}ialogue generation guided by retrieval
  memory.
\newblock In \emph{Proceedings of the 2019 Conference of the North American
  Chapter of the Association for Computational Linguistics: Human Language
  Technologies, {NAACL-HLT} 2019, Minneapolis, MN, USA, June 2-7, 2019, Volume
  1 (Long and Short Papers)}, pages 1219--1228, Association for Computational
  Linguistics.

\bibitem[{Cao and Xiong(2018)}]{cao2018encoding}
Cao, Qian and Deyi Xiong. 2018.
\newblock Encoding gated translation memory into neural machine translation.
\newblock In \emph{Proceedings of the 2018 Conference on Empirical Methods in
  Natural Language Processing, Brussels, Belgium, October 31 - November 4,
  2018}, pages 3042--3047, Association for Computational Linguistics.

\bibitem[{Cao et~al.(2020)Cao, Shui, Pan, Kan, Liu, and
  Chua}]{cao-etal-2020-expertise}
Cao, Yixin, Ruihao Shui, Liangming Pan, Min-Yen Kan, Zhiyuan Liu, and Tat-Seng
  Chua. 2020.
\newblock Expertise style transfer: {A} new task towards better communication
  between experts and laymen.
\newblock In \emph{Proceedings of the 58th Annual Meeting of the Association
  for Computational Linguistics}, pages 1061--1071, Association for
  Computational Linguistics, Online.

\bibitem[{Carey(1997)}]{carey1997taking}
Carey, Alex. 1997.
\newblock \emph{Taking the risk out of democracy: {C}orporate propaganda versus
  freedom and liberty}.
\newblock University of Illinois Press.

\bibitem[{Carlson, Riddell, and Rockmore(2018)}]{carlson2018evaluating}
Carlson, Keith, Allen Riddell, and Daniel Rockmore. 2018.
\newblock Evaluating prose style transfer with the bible.
\newblock \emph{Royal Society open science}, 5(10):171920.

\bibitem[{Castro, Ortega, and Mu{\~n}oz(2017)}]{castro2017author}
Castro, Daniel, Reynier Ortega, and Rafael Mu{\~n}oz. 2017.
\newblock Author masking by sentence transformation—notebook for pan at clef
  2017.
\newblock In \emph{CLEF 2017 Evaluation Labs and Workshop--Working Notes
  Papers}, pages 11--14.

\bibitem[{Cer et~al.(2018)Cer, Yang, Kong, Hua, Limtiaco, John, Constant,
  Guajardo{-}Cespedes, Yuan, Tar, Strope, and Kurzweil}]{cer2018universal}
Cer, Daniel, Yinfei Yang, Sheng{-}yi Kong, Nan Hua, Nicole Limtiaco, Rhomni~St.
  John, Noah Constant, Mario Guajardo{-}Cespedes, Steve Yuan, Chris Tar, Brian
  Strope, and Ray Kurzweil. 2018.
\newblock Universal sentence encoder for english.
\newblock In \emph{Proceedings of the 2018 Conference on Empirical Methods in
  Natural Language Processing, {EMNLP} 2018: System Demonstrations, Brussels,
  Belgium, October 31 - November 4, 2018}, pages 169--174, Association for
  Computational Linguistics.

\bibitem[{Chakrabarty, Muresan, and Peng(2020)}]{chakrabarty2020generating}
Chakrabarty, Tuhin, Smaranda Muresan, and Nanyun Peng. 2020.
\newblock Generating similes effortlessly like a pro: {A} style transfer
  approach for simile generation.
\newblock In \emph{Proceedings of the 2020 Conference on Empirical Methods in
  Natural Language Processing, {EMNLP} 2020, Online, November 16-20, 2020},
  pages 6455--6469, Association for Computational Linguistics.

\bibitem[{Chen and Dolan(2011)}]{chen2011collecting}
Chen, David~L. and William~B. Dolan. 2011.
\newblock Collecting highly parallel data for paraphrase evaluation.
\newblock In \emph{The 49th Annual Meeting of the Association for Computational
  Linguistics: Human Language Technologies, Proceedings of the Conference,
  19-24 June, 2011, Portland, Oregon, {USA}}, pages 190--200, The Association
  for Computer Linguistics.

\bibitem[{Chen et~al.(2017{\natexlab{a}})Chen, Liao, Yuan, Yu, and
  Hua}]{chen2017coherent}
Chen, Dongdong, Jing Liao, Lu~Yuan, Nenghai Yu, and Gang Hua.
  2017{\natexlab{a}}.
\newblock Coherent online video style transfer.
\newblock In \emph{{IEEE} International Conference on Computer Vision, {ICCV}
  2017, Venice, Italy, October 22-29, 2017}, pages 1114--1123, {IEEE} Computer
  Society.

\bibitem[{Chen et~al.(2017{\natexlab{b}})Chen, Yuan, Liao, Yu, and
  Hua}]{chen2017stylebank}
Chen, Dongdong, Lu~Yuan, Jing Liao, Nenghai Yu, and Gang Hua.
  2017{\natexlab{b}}.
\newblock Style{B}ank: {A}n explicit representation for neural image style
  transfer.
\newblock In \emph{2017 {IEEE} Conference on Computer Vision and Pattern
  Recognition, {CVPR} 2017, Honolulu, HI, USA, July 21-26, 2017}, pages
  2770--2779, {IEEE} Computer Society.

\bibitem[{Cho et~al.(2014)Cho, van Merrienboer, Bahdanau, and
  Bengio}]{cho2014properties}
Cho, Kyunghyun, Bart van Merrienboer, Dzmitry Bahdanau, and Yoshua Bengio.
  2014.
\newblock On the properties of neural machine translation: {E}ncoder-decoder
  approaches.
\newblock In \emph{Proceedings of SSST@EMNLP 2014, Eighth Workshop on Syntax,
  Semantics and Structure in Statistical Translation, Doha, Qatar, 25 October
  2014}, pages 103--111, Association for Computational Linguistics.

\bibitem[{Cohen(1960)}]{cohen1960coefficient}
Cohen, Jacob. 1960.
\newblock A coefficient of agreement for nominal scales.
\newblock \emph{Educational and psychological measurement}, 20(1):37--46.

\bibitem[{Dai et~al.(2019)Dai, Liang, Qiu, and Huang}]{dai-etal-2019-style}
Dai, Ning, Jianze Liang, Xipeng Qiu, and Xuanjing Huang. 2019.
\newblock Style transformer: {U}npaired text style transfer without
  disentangled latent representation.
\newblock In \emph{Proceedings of the 57th Annual Meeting of the Association
  for Computational Linguistics}, pages 5997--6007, Association for
  Computational Linguistics, Florence, Italy.

\bibitem[{Dathathri et~al.(2020)Dathathri, Madotto, Lan, Hung, Frank, Molino,
  Yosinski, and Liu}]{dathathri2020plug}
Dathathri, Sumanth, Andrea Madotto, Janice Lan, Jane Hung, Eric Frank, Piero
  Molino, Jason Yosinski, and Rosanne Liu. 2020.
\newblock Plug and play language models: {A} simple approach to controlled text
  generation.
\newblock In \emph{8th International Conference on Learning Representations,
  {ICLR} 2020, Addis Ababa, Ethiopia, April 26-30, 2020}, OpenReview.net.

\bibitem[{Devlin et~al.(2019)Devlin, Chang, Lee, and
  Toutanova}]{devlin2019bert}
Devlin, Jacob, Ming{-}Wei Chang, Kenton Lee, and Kristina Toutanova. 2019.
\newblock {BERT:} pre-training of deep bidirectional transformers for language
  understanding.
\newblock In \emph{Proceedings of the 2019 Conference of the North American
  Chapter of the Association for Computational Linguistics: Human Language
  Technologies, {NAACL-HLT} 2019, Minneapolis, MN, USA, June 2-7, 2019, Volume
  1 (Long and Short Papers)}, pages 4171--4186, Association for Computational
  Linguistics.

\bibitem[{Dou, Anastasopoulos, and Neubig(2020)}]{dou2020dynamic}
Dou, Zi{-}Yi, Antonios Anastasopoulos, and Graham Neubig. 2020.
\newblock Dynamic data selection and weighting for iterative back-translation.
\newblock In \emph{Proceedings of the 2020 Conference on Empirical Methods in
  Natural Language Processing, {EMNLP} 2020, Online, November 16-20, 2020},
  pages 5894--5904, Association for Computational Linguistics.

\bibitem[{Ferreira et~al.(2020)Ferreira, Gardent, Ilinykh, van~der Lee, Mille,
  Moussallem, and Shimorina}]{ferreira20202020}
Ferreira, Thiago, Claire Gardent, Nikolai Ilinykh, Chris van~der Lee, Simon
  Mille, Diego Moussallem, and Anastasia Shimorina. 2020.
\newblock The 2020 bilingual, bi-directional {WebNLG}+ shared task overview and
  evaluation results ({WebNLG}+ 2020).
\newblock In \emph{Proceedings of the 3rd International Workshop on Natural
  Language Generation from the Semantic Web (WebNLG+)}.

\bibitem[{Ficler and Goldberg(2017)}]{ficler2017controlling}
Ficler, Jessica and Yoav Goldberg. 2017.
\newblock Controlling linguistic style aspects in neural language generation.
\newblock \emph{CoRR}, abs/1707.02633.

\bibitem[{Fu et~al.(2019)Fu, Zhou, Chen, and Li}]{fu2019rethinking}
Fu, Yao, Hao Zhou, Jiaze Chen, and Lei Li. 2019.
\newblock Rethinking text attribute transfer: {A} lexical analysis.
\newblock In \emph{Proceedings of the 12th International Conference on Natural
  Language Generation, {INLG} 2019, Tokyo, Japan, October 29 - November 1,
  2019}, pages 24--33, Association for Computational Linguistics.

\bibitem[{Fu et~al.(2018)Fu, Tan, Peng, Zhao, and Yan}]{fu2018style}
Fu, Zhenxin, Xiaoye Tan, Nanyun Peng, Dongyan Zhao, and Rui Yan. 2018.
\newblock Style transfer in text: {E}xploration and evaluation.
\newblock In \emph{Proceedings of the Thirty-Second {AAAI} Conference on
  Artificial Intelligence, (AAAI-18), the 30th innovative Applications of
  Artificial Intelligence (IAAI-18), and the 8th {AAAI} Symposium on
  Educational Advances in Artificial Intelligence (EAAI-18), New Orleans,
  Louisiana, USA, February 2-7, 2018}, pages 663--670, {AAAI} Press.

\bibitem[{Gan et~al.(2017)Gan, Gan, He, Gao, and Deng}]{Gan2017StyleNetGA}
Gan, Chuang, Zhe Gan, Xiaodong He, Jianfeng Gao, and Li~Deng. 2017.
\newblock Style{N}et: {G}enerating attractive visual captions with styles.
\newblock In \emph{2017 {IEEE} Conference on Computer Vision and Pattern
  Recognition, {CVPR} 2017, Honolulu, HI, USA, July 21-26, 2017}, pages
  955--964, {IEEE} Computer Society.

\bibitem[{Gao, Singh, and Raj(2018)}]{gao2018voice}
Gao, Yang, Rita Singh, and Bhiksha Raj. 2018.
\newblock Voice impersonation using generative adversarial networks.
\newblock In \emph{2018 {IEEE} International Conference on Acoustics, Speech
  and Signal Processing, {ICASSP} 2018, Calgary, AB, Canada, April 15-20,
  2018}, pages 2506--2510, {IEEE}.

\bibitem[{Gardent et~al.(2017)Gardent, Shimorina, Narayan, and
  Perez{-}Beltrachini}]{gardent2017webnlg}
Gardent, Claire, Anastasia Shimorina, Shashi Narayan, and Laura
  Perez{-}Beltrachini. 2017.
\newblock The {WebNLG} challenge: Generating text from {RDF} data.
\newblock In \emph{Proceedings of the 10th International Conference on Natural
  Language Generation, {INLG} 2017, Santiago de Compostela, Spain, September
  4-7, 2017}, pages 124--133, Association for Computational Linguistics.

\bibitem[{Gardner et~al.(2020)Gardner, Artzi, Basmova, Berant, Bogin, Chen,
  Dasigi, Dua, Elazar, Gottumukkala, Gupta, Hajishirzi, Ilharco, Khashabi, Lin,
  Liu, Liu, Mulcaire, Ning, Singh, Smith, Subramanian, Tsarfaty, Wallace,
  Zhang, and Zhou}]{matt2020evaluating}
Gardner, Matt, Yoav Artzi, Victoria Basmova, Jonathan Berant, Ben Bogin, Sihao
  Chen, Pradeep Dasigi, Dheeru Dua, Yanai Elazar, Ananth Gottumukkala, Nitish
  Gupta, Hannaneh Hajishirzi, Gabriel Ilharco, Daniel Khashabi, Kevin Lin,
  Jiangming Liu, Nelson~F. Liu, Phoebe Mulcaire, Qiang Ning, Sameer Singh,
  Noah~A. Smith, Sanjay Subramanian, Reut Tsarfaty, Eric Wallace, Ally Zhang,
  and Ben Zhou. 2020.
\newblock Evaluating models' local decision boundaries via contrast sets.
\newblock In \emph{Proceedings of the 2020 Conference on Empirical Methods in
  Natural Language Processing: Findings, {EMNLP} 2020, Online Event, 16-20
  November 2020}, pages 1307--1323, Association for Computational Linguistics.

\bibitem[{Gatt and Reiter(2009)}]{gatt2009simplenlg}
Gatt, Albert and Ehud Reiter. 2009.
\newblock {SimpleNLG: A} realisation engine for practical applications.
\newblock In \emph{{ENLG} 2009 - Proceedings of the 12th European Workshop on
  Natural Language Generation, March 30-31, 2009, Athens, Greece}, pages
  90--93, The Association for Computer Linguistics.

\bibitem[{Gatys, Ecker, and Bethge(2016)}]{gatys2016image}
Gatys, Leon~A., Alexander~S. Ecker, and Matthias Bethge. 2016.
\newblock Image style transfer using convolutional neural networks.
\newblock In \emph{2016 {IEEE} Conference on Computer Vision and Pattern
  Recognition, {CVPR} 2016, Las Vegas, NV, USA, June 27-30, 2016}, pages
  2414--2423, {IEEE} Computer Society.

\bibitem[{Gehrmann et~al.(2021)Gehrmann, Adewumi, Aggarwal, Ammanamanchi,
  Anuoluwapo, Bosselut, Chandu, Clinciu, Das, Dhole, Du, Durmus, Dusek, Emezue,
  Gangal, Garbacea, Hashimoto, Hou, Jernite, Jhamtani, Ji, Jolly, Kumar,
  Ladhak, Madaan, Maddela, Mahajan, Mahamood, Majumder, Martins,
  McMillan{-}Major, Mille, van Miltenburg, Nadeem, Narayan, Nikolaev,
  Niyongabo, Osei, Parikh, Perez{-}Beltrachini, Rao, Raunak, Rodriguez,
  Santhanam, Sedoc, Sellam, Shaikh, Shimorina, Cabezudo, Strobelt, Subramani,
  Xu, Yang, Yerukola, and Zhou}]{gehrmann2021thegem}
Gehrmann, Sebastian, Tosin~P. Adewumi, Karmanya Aggarwal, Pawan~Sasanka
  Ammanamanchi, Aremu Anuoluwapo, Antoine Bosselut, Khyathi~Raghavi Chandu,
  Miruna{-}Adriana Clinciu, Dipanjan Das, Kaustubh~D. Dhole, Wanyu Du, Esin
  Durmus, Ondrej Dusek, Chris Emezue, Varun Gangal, Cristina Garbacea,
  Tatsunori Hashimoto, Yufang Hou, Yacine Jernite, Harsh Jhamtani, Yangfeng Ji,
  Shailza Jolly, Dhruv Kumar, Faisal Ladhak, Aman Madaan, Mounica Maddela,
  Khyati Mahajan, Saad Mahamood, Bodhisattwa~Prasad Majumder, Pedro~Henrique
  Martins, Angelina McMillan{-}Major, Simon Mille, Emiel van Miltenburg, Moin
  Nadeem, Shashi Narayan, Vitaly Nikolaev, Rubungo~Andre Niyongabo, Salomey
  Osei, Ankur~P. Parikh, Laura Perez{-}Beltrachini, Niranjan~Ramesh Rao, Vikas
  Raunak, Juan~Diego Rodriguez, Sashank Santhanam, Jo{\~{a}}o Sedoc, Thibault
  Sellam, Samira Shaikh, Anastasia Shimorina, Marco Antonio~Sobrevilla
  Cabezudo, Hendrik Strobelt, Nishant Subramani, Wei Xu, Diyi Yang, Akhila
  Yerukola, and Jiawei Zhou. 2021.
\newblock The {GEM} benchmark: {N}atural language generation, its evaluation
  and metrics.
\newblock \emph{CoRR}, abs/2102.01672.

\bibitem[{Gkatzia, Lemon, and Rieser(2017)}]{gkatzia2017data}
Gkatzia, Dimitra, Oliver Lemon, and Verena Rieser. 2017.
\newblock Data-to-text generation improves decision-making under uncertainty.
\newblock \emph{{IEEE} Computational Intelligence Magazine}, 12(3):10--17.

\bibitem[{Gong et~al.(2019)Gong, Bhat, Wu, Xiong, and
  Hwu}]{gong-etal-2019-reinforcement}
Gong, Hongyu, Suma Bhat, Lingfei Wu, JinJun Xiong, and Wen-mei Hwu. 2019.
\newblock Reinforcement learning based text style transfer without parallel
  training corpus.
\newblock In \emph{Proceedings of the 2019 Conference of the North {A}merican
  Chapter of the Association for Computational Linguistics: Human Language
  Technologies, Volume 1 (Long and Short Papers)}, pages 3168--3180,
  Association for Computational Linguistics, Minneapolis, Minnesota.

\bibitem[{Goodfellow et~al.(2014)Goodfellow, Pouget-Abadie, Mirza, Xu,
  Warde-Farley, Ozair, Courville, and Bengio}]{Goodfellow2014GenerativeAN}
Goodfellow, Ian~J., Jean Pouget-Abadie, M.~Mirza, Bing Xu, David Warde-Farley,
  Sherjil Ozair, Aaron~C. Courville, and Yoshua Bengio. 2014.
\newblock Generative adversarial nets.
\newblock In \emph{NIPS}.

\bibitem[{Goodman(1947)}]{goodman1947problem}
Goodman, Nelson. 1947.
\newblock The problem of counterfactual conditionals.
\newblock \emph{The Journal of Philosophy}, 44(5):113--128.

\bibitem[{Gr{\'{e}}goire and Langlais(2018)}]{gregoire2018extracting}
Gr{\'{e}}goire, Francis and Philippe Langlais. 2018.
\newblock Extracting parallel sentences with bidirectional recurrent neural
  networks to improve machine translation.
\newblock In \emph{Proceedings of the 27th International Conference on
  Computational Linguistics, {COLING} 2018, Santa Fe, New Mexico, USA, August
  20-26, 2018}, pages 1442--1453, Association for Computational Linguistics.

\bibitem[{Gr{\"{o}}ndahl and Asokan(2020)}]{grondahl2020effective}
Gr{\"{o}}ndahl, Tommi and N.~Asokan. 2020.
\newblock Effective writing style transfer via combinatorial paraphrasing.
\newblock \emph{Proceedings on Privacy Enhancing Technologies},
  2020(4):175--195.

\bibitem[{Gu et~al.(2016)Gu, Lu, Li, and Li}]{gu-etal-2016-incorporating}
Gu, Jiatao, Zhengdong Lu, Hang Li, and Victor~O.K. Li. 2016.
\newblock Incorporating copying mechanism in sequence-to-sequence learning.
\newblock In \emph{Proceedings of the 54th Annual Meeting of the Association
  for Computational Linguistics (Volume 1: Long Papers)}, pages 1631--1640,
  Association for Computational Linguistics, Berlin, Germany.

\bibitem[{Gu et~al.(2018)Gu, Wang, Cho, and Li}]{gu2018search}
Gu, Jiatao, Yong Wang, Kyunghyun Cho, and Victor O.~K. Li. 2018.
\newblock Search engine guided neural machine translation.
\newblock In \emph{Proceedings of the Thirty-Second {AAAI} Conference on
  Artificial Intelligence, (AAAI-18), the 30th innovative Applications of
  Artificial Intelligence (IAAI-18), and the 8th {AAAI} Symposium on
  Educational Advances in Artificial Intelligence (EAAI-18), New Orleans,
  Louisiana, USA, February 2-7, 2018}, pages 5133--5140, {AAAI} Press.

\bibitem[{G{\"{u}}l{\c{c}}ehre et~al.(2016)G{\"{u}}l{\c{c}}ehre, Ahn,
  Nallapati, Zhou, and Bengio}]{gulcehre2016pointing}
G{\"{u}}l{\c{c}}ehre, {\c{C}}aglar, Sungjin Ahn, Ramesh Nallapati, Bowen Zhou,
  and Yoshua Bengio. 2016.
\newblock Pointing the unknown words.
\newblock In \emph{Proceedings of the 54th Annual Meeting of the Association
  for Computational Linguistics, {ACL} 2016, August 7-12, 2016, Berlin,
  Germany, Volume 1: Long Papers}, The Association for Computer Linguistics.

\bibitem[{Guo et~al.(2020)Guo, Jin, Dai, Qiu, Xue, Wipf, and Zhang}]{p22020guo}
Guo, Qipeng, Zhijing Jin, Ning Dai, Xipeng Qiu, Xiangyang Xue, David Wipf, and
  Zheng Zhang. 2020.
\newblock P2: {A} plan-and-pretrain approach for knowledge graph-to-text
  generation.
\newblock In \emph{Proceedings of the 3rd WebNLG Workshop on Natural Language
  Generation from the Semantic Web (WebNLG+ 2020)}, Association for
  Computational Linguistics, Dublin, Ireland (Virtual).

\bibitem[{Guo et~al.(2021)Guo, Jin, Wang, Qiu, Zhang, Zhu, Zhang, and
  Wipf}]{guo2020fork}
Guo, Qipeng, Zhijing Jin, Ziyu Wang, Xipeng Qiu, Weinan Zhang, Jun Zhu, Zheng
  Zhang, and David Wipf. 2021.
\newblock Fork or fail: {C}ycle-consistent training with many-to-one mappings.
\newblock In \emph{The 24th International Conference on Artificial Intelligence
  and Statistics, {AISTATS} 2021, April 13-15, 2021, Virtual Event}, volume 130
  of \emph{Proceedings of Machine Learning Research}, pages 1828--1836, {PMLR}.

\bibitem[{Guu et~al.(2018)Guu, Hashimoto, Oren, and Liang}]{guu2018generating}
Guu, Kelvin, Tatsunori~B. Hashimoto, Yonatan Oren, and Percy Liang. 2018.
\newblock Generating sentences by editing prototypes.
\newblock \emph{Transactions of the Association for Computational Linguistics},
  6:437--450.

\bibitem[{Harrison(2019)}]{wired2019twitter}
Harrison, Sara. 2019.
\newblock Twitter and instagram unveil new ways to combat hate{--}again.

\bibitem[{Hashimoto et~al.(2018)Hashimoto, Guu, Oren, and
  Liang}]{hashimoto2018retrieve}
Hashimoto, Tatsunori~B., Kelvin Guu, Yonatan Oren, and Percy Liang. 2018.
\newblock A retrieve-and-edit framework for predicting structured outputs.
\newblock In \emph{Advances in Neural Information Processing Systems 31: Annual
  Conference on Neural Information Processing Systems 2018, NeurIPS 2018, 3-8
  December 2018, Montr{\'{e}}al, Canada}, pages 10073--10083.

\bibitem[{He and McAuley(2016)}]{he2016ups}
He, Ruining and Julian~J. McAuley. 2016.
\newblock Ups and downs: {M}odeling the visual evolution of fashion trends with
  one-class collaborative filtering.
\newblock In \emph{Proceedings of the 25th International Conference on World
  Wide Web, {WWW} 2016, Montreal, Canada, April 11 - 15, 2016}, pages 507--517,
  {ACM}.

\bibitem[{Henderson(2020)}]{henderson2020unstoppable}
Henderson, James. 2020.
\newblock The unstoppable rise of computational linguistics in deep learning.
\newblock In \emph{Proceedings of the 58th Annual Meeting of the Association
  for Computational Linguistics, {ACL} 2020, Online, July 5-10, 2020}, pages
  6294--6306, Association for Computational Linguistics.

\bibitem[{Hill, Cho, and Korhonen(2016)}]{hill2016learning}
Hill, Felix, Kyunghyun Cho, and Anna Korhonen. 2016.
\newblock Learning distributed representations of sentences from unlabelled
  data.
\newblock In \emph{{NAACL} {HLT} 2016, The 2016 Conference of the North
  American Chapter of the Association for Computational Linguistics: Human
  Language Technologies, San Diego California, USA, June 12-17, 2016}, pages
  1367--1377, The Association for Computational Linguistics.

\bibitem[{Hoang et~al.(2018)Hoang, Koehn, Haffari, and
  Cohn}]{hoang2018iterative}
Hoang, Vu Cong~Duy, Philipp Koehn, Gholamreza Haffari, and Trevor Cohn. 2018.
\newblock Iterative back-translation for neural machine translation.
\newblock In \emph{Proceedings of the 2nd Workshop on Neural Machine
  Translation and Generation}, pages 18--24, Association for Computational
  Linguistics, Melbourne, Australia.

\bibitem[{Hossain, Ghazvininejad, and Zettlemoyer(2020)}]{hossain2020simple}
Hossain, Nabil, Marjan Ghazvininejad, and Luke Zettlemoyer. 2020.
\newblock Simple and effective retrieve-edit-rerank text generation.
\newblock In \emph{Proceedings of the 58th Annual Meeting of the Association
  for Computational Linguistics, {ACL} 2020, Online, July 5-10, 2020}, pages
  2532--2538, Association for Computational Linguistics.

\bibitem[{Hovy and Spruit(2016)}]{hovy2016social}
Hovy, Dirk and Shannon~L. Spruit. 2016.
\newblock The social impact of natural language processing.
\newblock In \emph{Proceedings of the 54th Annual Meeting of the Association
  for Computational Linguistics, {ACL} 2016, August 7-12, 2016, Berlin,
  Germany, Volume 2: Short Papers}, The Association for Computer Linguistics.

\bibitem[{Hovy(1987)}]{hovy1987generating}
Hovy, Eduard. 1987.
\newblock Generating natural language under pragmatic constraints.
\newblock \emph{Journal of Pragmatics}, 11(6):689--719.

\bibitem[{Hovy(1990)}]{hovy1990pragmatics}
Hovy, Eduard~H. 1990.
\newblock Pragmatics and natural language generation.
\newblock \emph{Artificial Intelligence}, 43(2):153--197.

\bibitem[{Hu, Lee, and Aggarwal(2020)}]{Hu2020TextST}
Hu, Zhiqiang, R.~K. Lee, and C.~Aggarwal. 2020.
\newblock Text style transfer: {A} review and experiment evaluation.
\newblock \emph{ArXiv}, abs/2010.12742.

\bibitem[{Hu et~al.(2017)Hu, Yang, Liang, Salakhutdinov, and
  Xing}]{Hu2017TowardCG}
Hu, Zhiting, Zichao Yang, Xiaodan Liang, R.~Salakhutdinov, and E.~Xing. 2017.
\newblock Toward controlled generation of text.
\newblock In \emph{ICML}.

\bibitem[{Huang et~al.(2018)Huang, Za{\"\i}ane, Trabelsi, and
  Dziri}]{huang-etal-2018-automatic}
Huang, Chenyang, Osmar Za{\"\i}ane, Amine Trabelsi, and Nouha Dziri. 2018.
\newblock Automatic dialogue generation with expressed emotions.
\newblock In \emph{Proceedings of the 2018 Conference of the North {A}merican
  Chapter of the Association for Computational Linguistics: Human Language
  Technologies, Volume 2 (Short Papers)}, pages 49--54, Association for
  Computational Linguistics, New Orleans, Louisiana.

\bibitem[{Huang et~al.(2020)Huang, Zhu, Xiong, Zhang, Hu, and
  Xu}]{huang2020cycle}
Huang, Yufang, Wentao Zhu, Deyi Xiong, Yiye Zhang, Changjian Hu, and Feiyu Xu.
  2020.
\newblock Cycle-consistent adversarial autoencoders for unsupervised text style
  transfer.
\newblock In \emph{Proceedings of the 28th International Conference on
  Computational Linguistics, {COLING} 2020, Barcelona, Spain (Online), December
  8-13, 2020}, pages 2213--2223, International Committee on Computational
  Linguistics.

\bibitem[{Jackson et~al.(2019)Jackson, Abarghouei, Bonner, Breckon, and
  Obara}]{jackson2019style}
Jackson, Philip T.~G., Amir~Atapour Abarghouei, Stephen Bonner, Toby~P.
  Breckon, and Boguslaw Obara. 2019.
\newblock Style augmentation: {D}ata augmentation via style randomization.
\newblock In \emph{{IEEE} Conference on Computer Vision and Pattern Recognition
  Workshops, {CVPR} Workshops 2019, Long Beach, CA, USA, June 16-20, 2019},
  pages 83--92, Computer Vision Foundation / {IEEE}.

\bibitem[{Jafaritazehjani et~al.(2020)Jafaritazehjani, Lecorv{\'{e}}, Lolive,
  and Kelleher}]{jafaritazehjani2020style}
Jafaritazehjani, Somayeh, Gw{\'{e}}nol{\'{e}} Lecorv{\'{e}}, Damien Lolive, and
  John Kelleher. 2020.
\newblock Style versus content: {A} distinction without a (learnable)
  difference?
\newblock In \emph{Proceedings of the 28th International Conference on
  Computational Linguistics, {COLING} 2020, Barcelona, Spain (Online), December
  8-13, 2020}, pages 2169--2180, International Committee on Computational
  Linguistics.

\bibitem[{Jang, Gu, and Poole(2017)}]{jang2016categorical}
Jang, Eric, Shixiang Gu, and Ben Poole. 2017.
\newblock Categorical reparameterization with gumbel-softmax.
\newblock In \emph{5th International Conference on Learning Representations,
  {ICLR} 2017, Toulon, France, April 24-26, 2017, Conference Track
  Proceedings}, OpenReview.net.

\bibitem[{Jhamtani et~al.(2017)Jhamtani, Gangal, Hovy, and
  Nyberg}]{jhamtani2017shakespearizing}
Jhamtani, Harsh, Varun Gangal, Eduard~H. Hovy, and Eric Nyberg. 2017.
\newblock Shakespearizing modern language using copy-enriched
  sequence-to-sequence models.
\newblock \emph{CoRR}, abs/1707.01161.

\bibitem[{Jia and Liang(2017)}]{jia2017adversarial}
Jia, Robin and Percy Liang. 2017.
\newblock Adversarial examples for evaluating reading comprehension systems.
\newblock In \emph{Proceedings of the 2017 Conference on Empirical Methods in
  Natural Language Processing, {EMNLP} 2017, Copenhagen, Denmark, September
  9-11, 2017}, pages 2021--2031, Association for Computational Linguistics.

\bibitem[{Jin et~al.(2020{\natexlab{a}})Jin, Jin, Zhou, Orii, and
  Szolovits}]{jin-etal-2020-hooks}
Jin, Di, Zhijing Jin, Joey~Tianyi Zhou, Lisa Orii, and Peter Szolovits.
  2020{\natexlab{a}}.
\newblock Hooks in the headline: {L}earning to generate headlines with
  controlled styles.
\newblock In \emph{Proceedings of the 58th Annual Meeting of the Association
  for Computational Linguistics}, pages 5082--5093, Association for
  Computational Linguistics, Online.

\bibitem[{Jin et~al.(2020{\natexlab{b}})Jin, Jin, Zhou, and
  Szolovits}]{jin2020bert}
Jin, Di, Zhijing Jin, Joey~Tianyi Zhou, and Peter Szolovits.
  2020{\natexlab{b}}.
\newblock Is {BERT} really robust? {A} strong baseline for natural language
  attack on text classification and entailment.
\newblock In \emph{The Thirty-Fourth {AAAI} Conference on Artificial
  Intelligence, {AAAI} 2020, The Thirty-Second Innovative Applications of
  Artificial Intelligence Conference, {IAAI} 2020, The Tenth {AAAI} Symposium
  on Educational Advances in Artificial Intelligence, {EAAI} 2020, New York,
  NY, USA, February 7-12, 2020}, pages 8018--8025, {AAAI} Press.

\bibitem[{Jin et~al.(2019)Jin, Jin, Mueller, Matthews, and
  Santus}]{jin2019imat}
Jin, Zhijing, Di~Jin, Jonas Mueller, Nicholas Matthews, and Enrico Santus.
  2019.
\newblock {IMaT: U}nsupervised text attribute transfer via iterative matching
  and translation.
\newblock In \emph{Proceedings of the 2019 Conference on Empirical Methods in
  Natural Language Processing and the 9th International Joint Conference on
  Natural Language Processing, {EMNLP-IJCNLP} 2019, Hong Kong, China, November
  3-7, 2019}, pages 3095--3107, Association for Computational Linguistics.

\bibitem[{John et~al.(2019)John, Mou, Bahuleyan, and
  Vechtomova}]{john-etal-2019-disentangled}
John, Vineet, Lili Mou, Hareesh Bahuleyan, and Olga Vechtomova. 2019.
\newblock Disentangled representation learning for non-parallel text style
  transfer.
\newblock In \emph{Proceedings of the 57th Annual Meeting of the Association
  for Computational Linguistics}, pages 424--434, Association for Computational
  Linguistics, Florence, Italy.

\bibitem[{Kajiwara(2019)}]{kajiwara-2019-negative}
Kajiwara, Tomoyuki. 2019.
\newblock Negative lexically constrained decoding for paraphrase generation.
\newblock In \emph{Proceedings of the 57th Annual Meeting of the Association
  for Computational Linguistics}, pages 6047--6052, Association for
  Computational Linguistics, Florence, Italy.

\bibitem[{Kale and Rastogi(2020)}]{kale2020text}
Kale, Mihir and Abhinav Rastogi. 2020.
\newblock Text-to-text pre-training for data-to-text tasks.
\newblock In \emph{Proceedings of the 13th International Conference on Natural
  Language Generation, {INLG} 2020, Dublin, Ireland, December 15-18, 2020},
  pages 97--102, Association for Computational Linguistics.

\bibitem[{Kang, Wang, and de~Melo(2020)}]{kang2020incorporating}
Kang, Yipeng, Tonghan Wang, and Gerard de~Melo. 2020.
\newblock Incorporating pragmatic reasoning communication into emergent
  language.
\newblock In \emph{Advances in Neural Information Processing Systems 33: Annual
  Conference on Neural Information Processing Systems 2020, NeurIPS 2020,
  December 6-12, 2020, virtual}.

\bibitem[{Karadzhov et~al.(2017)Karadzhov, Mihaylova, Kiprov, Georgiev,
  Koychev, and Nakov}]{karadzhov2017case}
Karadzhov, Georgi, Tsvetomila Mihaylova, Yasen Kiprov, Georgi Georgiev, Ivan
  Koychev, and Preslav Nakov. 2017.
\newblock The case for being average: {A} mediocrity approach to style masking
  and author obfuscation - (best of the labs track at {CLEF-2017)}.
\newblock In \emph{Experimental {IR} Meets Multilinguality, Multimodality, and
  Interaction - 8th International Conference of the {CLEF} Association, {CLEF}
  2017, Dublin, Ireland, September 11-14, 2017, Proceedings}, volume 10456 of
  \emph{Lecture Notes in Computer Science}, pages 173--185, Springer.

\bibitem[{Keskar et~al.(2019)Keskar, McCann, Varshney, Xiong, and
  Socher}]{keskar2019ctrl}
Keskar, Nitish~Shirish, Bryan McCann, Lav~R. Varshney, Caiming Xiong, and
  Richard Socher. 2019.
\newblock {CTRL:} {A} conditional transformer language model for controllable
  generation.
\newblock \emph{CoRR}, abs/1909.05858.

\bibitem[{Khosmood and Levinson(2010)}]{khosmood2010automatic}
Khosmood, Foaad and Robert Levinson. 2010.
\newblock Automatic synonym and phrase replacement show promise for style
  transformation.
\newblock In \emph{The Ninth International Conference on Machine Learning and
  Applications, {ICMLA} 2010, Washington, DC, USA, 12-14 December 2010}, pages
  958--961, {IEEE} Computer Society.

\bibitem[{Khosmood and Levinson(2008)}]{khosmood2008automatic}
Khosmood, Foaad and Robert~A Levinson. 2008.
\newblock Automatic natural language style classification and transformation.
\newblock In \emph{BCS-IRSG Workshop on Corpus Profiling}, pages 1--11.

\bibitem[{Kim(2014)}]{kim2014convolutional}
Kim, Yoon. 2014.
\newblock Convolutional neural networks for sentence classification.
\newblock In \emph{Proceedings of the 2014 Conference on Empirical Methods in
  Natural Language Processing, {EMNLP} 2014, October 25-29, 2014, Doha, Qatar,
  {A} meeting of SIGDAT, a Special Interest Group of the {ACL}}, pages
  1746--1751, {ACL}.

\bibitem[{Kingma and Welling(2014)}]{Kingma2014AutoEncodingVB}
Kingma, Diederik~P. and M.~Welling. 2014.
\newblock Auto-encoding variational bayes.
\newblock \emph{CoRR}, abs/1312.6114.

\bibitem[{Koncel{-}Kedziorski et~al.(2016)Koncel{-}Kedziorski, Konstas,
  Zettlemoyer, and Hajishirzi}]{koncel-kedziorski2016theme}
Koncel{-}Kedziorski, Rik, Ioannis Konstas, Luke Zettlemoyer, and Hannaneh
  Hajishirzi. 2016.
\newblock A theme-rewriting approach for generating algebra word problems.
\newblock In \emph{Proceedings of the 2016 Conference on Empirical Methods in
  Natural Language Processing, {EMNLP} 2016, Austin, Texas, USA, November 1-4,
  2016}, pages 1617--1628, The Association for Computational Linguistics.

\bibitem[{Kowalski et~al.(2014)Kowalski, Giumetti, Schroeder, and
  Lattanner}]{kowalski2014bullying}
Kowalski, Robin~M, Gary~W Giumetti, Amber~N Schroeder, and Micah~R Lattanner.
  2014.
\newblock Bullying in the digital age: {A} critical review and meta-analysis of
  cyberbullying research among youth.
\newblock \emph{Psychological bulletin}, 140(4):1073.

\bibitem[{Krippendorff(2018)}]{krippendorff2018content}
Krippendorff, Klaus. 2018.
\newblock \emph{Content analysis: {A}n introduction to its methodology}.
\newblock Sage publications.

\bibitem[{Krishna, Wieting, and Iyyer(2020)}]{krishna2020reformulating}
Krishna, Kalpesh, John Wieting, and Mohit Iyyer. 2020.
\newblock Reformulating unsupervised style transfer as paraphrase generation.
\newblock \emph{CoRR}, abs/2010.05700.

\bibitem[{Kusner et~al.(2015)Kusner, Sun, Kolkin, and
  Weinberger}]{kusner2015word}
Kusner, Matt, Yu~Sun, Nicholas Kolkin, and Kilian Weinberger. 2015.
\newblock From word embeddings to document distances.
\newblock In \emph{International Conference on Machine Learning}, pages
  957--966.

\bibitem[{Lai, Toral, and Nissim(2021)}]{Lai2021ThankYB}
Lai, Huiyuan, Antonio Toral, and M.~Nissim. 2021.
\newblock Thank you bart! rewarding pre-trained models improves formality style
  transfer.
\newblock In \emph{ACL/IJCNLP}.

\bibitem[{Lakoff(1973)}]{lakoff1973language}
Lakoff, Robin. 1973.
\newblock Language and woman's place.
\newblock \emph{Language in society}, 2(1):45--79.

\bibitem[{Lample et~al.(2018{\natexlab{a}})Lample, Conneau, Denoyer, and
  Ranzato}]{lample2018unsupervised}
Lample, Guillaume, Alexis Conneau, Ludovic Denoyer, and Marc'Aurelio Ranzato.
  2018{\natexlab{a}}.
\newblock Unsupervised machine translation using monolingual corpora only.
\newblock In \emph{6th International Conference on Learning Representations,
  {ICLR} 2018, Vancouver, BC, Canada, April 30 - May 3, 2018, Conference Track
  Proceedings}, OpenReview.net.

\bibitem[{Lample et~al.(2018{\natexlab{b}})Lample, Ott, Conneau, Denoyer, and
  Ranzato}]{lample2018phrase}
Lample, Guillaume, Myle Ott, Alexis Conneau, Ludovic Denoyer, and Marc'Aurelio
  Ranzato. 2018{\natexlab{b}}.
\newblock Phrase-based {\&} neural unsupervised machine translation.
\newblock In \emph{Proceedings of the 2018 Conference on Empirical Methods in
  Natural Language Processing, Brussels, Belgium, October 31 - November 4,
  2018}, pages 5039--5049, Association for Computational Linguistics.

\bibitem[{Lample et~al.(2019)Lample, Subramanian, Smith, Denoyer, Ranzato, and
  Boureau}]{Lample2019MultipleAttributeTR}
Lample, Guillaume, Sandeep Subramanian, Eric~Michael Smith, Ludovic Denoyer,
  Marc'Aurelio Ranzato, and Y-Lan Boureau. 2019.
\newblock Multiple-attribute text rewriting.
\newblock In \emph{ICLR}.

\bibitem[{Lee(2020)}]{lee2020stable}
Lee, Joosung. 2020.
\newblock Stable style transformer: {D}elete and generate approach with
  encoder-decoder for text style transfer.
\newblock \emph{CoRR}, abs/2005.12086.

\bibitem[{Lewis et~al.(2020)Lewis, Perez, Piktus, Petroni, Karpukhin, Goyal,
  K{\"{u}}ttler, Lewis, Yih, Rockt{\"{a}}schel, Riedel, and
  Kiela}]{lewis2020retrieval}
Lewis, Patrick S.~H., Ethan Perez, Aleksandra Piktus, Fabio Petroni, Vladimir
  Karpukhin, Naman Goyal, Heinrich K{\"{u}}ttler, Mike Lewis, Wen{-}tau Yih,
  Tim Rockt{\"{a}}schel, Sebastian Riedel, and Douwe Kiela. 2020.
\newblock Retrieval-augmented generation for knowledge-intensive {NLP} tasks.
\newblock \emph{CoRR}, abs/2005.11401.

\bibitem[{Li et~al.(2019)Li, Zhang, Gan, Cheng, Brockett, Dolan, and
  Sun}]{li2019domain}
Li, Dianqi, Yizhe Zhang, Zhe Gan, Yu~Cheng, Chris Brockett, Bill Dolan, and
  Ming{-}Ting Sun. 2019.
\newblock Domain adaptive text style transfer.
\newblock In \emph{Proceedings of the 2019 Conference on Empirical Methods in
  Natural Language Processing and the 9th International Joint Conference on
  Natural Language Processing, {EMNLP-IJCNLP} 2019, Hong Kong, China, November
  3-7, 2019}, pages 3302--3311, Association for Computational Linguistics.

\bibitem[{Li et~al.(2016)Li, Galley, Brockett, Spithourakis, Gao, and
  Dolan}]{li2016persona}
Li, Jiwei, Michel Galley, Chris Brockett, Georgios~P. Spithourakis, Jianfeng
  Gao, and William~B. Dolan. 2016.
\newblock A persona-based neural conversation model.
\newblock In \emph{Proceedings of the 54th Annual Meeting of the Association
  for Computational Linguistics, {ACL} 2016, August 7-12, 2016, Berlin,
  Germany, Volume 1: Long Papers}, The Association for Computer Linguistics.

\bibitem[{Li et~al.(2018)Li, Jia, He, and Liang}]{li-etal-2018-delete}
Li, Juncen, Robin Jia, He~He, and Percy Liang. 2018.
\newblock Delete, retrieve, generate: {A} simple approach to sentiment and
  style transfer.
\newblock In \emph{Proceedings of the 2018 Conference of the North {A}merican
  Chapter of the Association for Computational Linguistics: Human Language
  Technologies, Volume 1 (Long Papers)}, pages 1865--1874, Association for
  Computational Linguistics, New Orleans, Louisiana.

\bibitem[{Li et~al.(2021)Li, Chen, Yang, Gao, Zhao, and
  Yan}]{li2020stylecontent}
Li, Mingzhe, Xiuying Chen, Min Yang, Shen Gao, Dongyan Zhao, and Rui Yan. 2021.
\newblock The style-content duality of attractiveness: {L}earning to write
  eye-catching headlines via disentanglement.
\newblock In \emph{Thirty-Fifth {AAAI} Conference on Artificial Intelligence,
  {AAAI} 2021, Thirty-Third Conference on Innovative Applications of Artificial
  Intelligence, {IAAI} 2021, The Eleventh Symposium on Educational Advances in
  Artificial Intelligence, {EAAI} 2021, Virtual Event, February 2-9, 2021},
  pages 13252--13260, {AAAI} Press.

\bibitem[{Li and Liang(2021)}]{li2020prefix}
Li, Xiang~Lisa and Percy Liang. 2021.
\newblock Prefix-tuning: {O}ptimizing continuous prompts for generation.
\newblock In \emph{Proceedings of the 59th Annual Meeting of the Association
  for Computational Linguistics and the 11th International Joint Conference on
  Natural Language Processing, {ACL/IJCNLP} 2021, (Volume 1: Long Papers),
  Virtual Event, August 1-6, 2021}, pages 4582--4597, Association for
  Computational Linguistics.

\bibitem[{Li et~al.(2020)Li, Li, Zhang, Li, Zheng, Carin, and
  Gao}]{Li2020ComplementaryAC}
Li, Yuan, Chunyuan Li, Yizhe Zhang, Xiujun Li, Guoqing Zheng, Lawrence Carin,
  and Jianfeng Gao. 2020.
\newblock Complementary auxiliary classifiers for label-conditional text
  generation.
\newblock In \emph{The Thirty-Fourth {AAAI} Conference on Artificial
  Intelligence, {AAAI} 2020, The Thirty-Second Innovative Applications of
  Artificial Intelligence Conference, {IAAI} 2020, The Tenth {AAAI} Symposium
  on Educational Advances in Artificial Intelligence, {EAAI} 2020, New York,
  NY, USA, February 7-12, 2020}, pages 8303--8310, {AAAI} Press.

\bibitem[{Liao et~al.(2018)Liao, Bing, Li, Shi, Lam, and
  Zhang}]{liao-etal-2018-quase}
Liao, Yi, Lidong Bing, Piji Li, Shuming Shi, Wai Lam, and Tong Zhang. 2018.
\newblock {Q}ua{SE}: {S}equence editing under quantifiable guidance.
\newblock In \emph{Proceedings of the 2018 Conference on Empirical Methods in
  Natural Language Processing}, pages 3855--3864, Association for Computational
  Linguistics, Brussels, Belgium.

\bibitem[{Lin and Och(2004)}]{lin-och-2004-automatic}
Lin, Chin-Yew and Franz~Josef Och. 2004.
\newblock Automatic evaluation of machine translation quality using longest
  common subsequence and skip-bigram statistics.
\newblock In \emph{Proceedings of the 42nd Annual Meeting of the Association
  for Computational Linguistics ({ACL}-04)}, pages 605--612, Barcelona, Spain.

\bibitem[{Liu et~al.(2020)Liu, Fu, Zhang, Pal, and Lv}]{liu2019revision}
Liu, Dayiheng, Jie Fu, Yidan Zhang, Chris Pal, and Jiancheng Lv. 2020.
\newblock Revision in continuous space: {U}nsupervised text style transfer
  without adversarial learning.
\newblock In \emph{The Thirty-Fourth {AAAI} Conference on Artificial
  Intelligence, {AAAI} 2020, The Thirty-Second Innovative Applications of
  Artificial Intelligence Conference, {IAAI} 2020, The Tenth {AAAI} Symposium
  on Educational Advances in Artificial Intelligence, {EAAI} 2020, New York,
  NY, USA, February 7-12, 2020}, pages 8376--8383, {AAAI} Press.

\bibitem[{Locatello et~al.(2019)Locatello, Bauer, Lucic, R{\"{a}}tsch, Gelly,
  Sch{\"{o}}lkopf, and Bachem}]{locatello2019challenging}
Locatello, Francesco, Stefan Bauer, Mario Lucic, Gunnar R{\"{a}}tsch, Sylvain
  Gelly, Bernhard Sch{\"{o}}lkopf, and Olivier Bachem. 2019.
\newblock Challenging common assumptions in the unsupervised learning of
  disentangled representations.
\newblock In \emph{Proceedings of the 36th International Conference on Machine
  Learning, {ICML} 2019, 9-15 June 2019, Long Beach, California, {USA}},
  volume~97 of \emph{Proceedings of Machine Learning Research}, pages
  4114--4124, {PMLR}.

\bibitem[{Locatello et~al.(2020)Locatello, Poole, R{\"{a}}tsch,
  Sch{\"{o}}lkopf, Bachem, and Tschannen}]{locatello2020weakly}
Locatello, Francesco, Ben Poole, Gunnar R{\"{a}}tsch, Bernhard Sch{\"{o}}lkopf,
  Olivier Bachem, and Michael Tschannen. 2020.
\newblock Weakly-supervised disentanglement without compromises.
\newblock In \emph{Proceedings of the 37th International Conference on Machine
  Learning, {ICML} 2020, 13-18 July 2020, Virtual Event}, volume 119 of
  \emph{Proceedings of Machine Learning Research}, pages 6348--6359, {PMLR}.

\bibitem[{Logeswaran, Lee, and Bengio(2018)}]{logeswaran2018content}
Logeswaran, Lajanugen, Honglak Lee, and Samy Bengio. 2018.
\newblock Content preserving text generation with attribute controls.
\newblock In \emph{Advances in Neural Information Processing Systems 31: Annual
  Conference on Neural Information Processing Systems 2018, NeurIPS 2018,
  December 3-8, 2018, Montr{\'{e}}al, Canada}, pages 5108--5118.

\bibitem[{Luo et~al.(2019)Luo, Li, Zhou, Yang, Chang, Sui, and
  Sun}]{Luo19DualRL}
Luo, Fuli, Peng Li, Jie Zhou, Pengcheng Yang, Baobao Chang, Zhifang Sui, and
  Xu~Sun. 2019.
\newblock A dual reinforcement learning framework for unsupervised text style
  transfer.
\newblock In \emph{Proceedings of the 28th International Joint Conference on
  Artificial Intelligence, {IJCAI} 2019}.

\bibitem[{Ma et~al.(2020)Ma, Sap, Rashkin, and Choi}]{ma2020powertransformer}
Ma, Xinyao, Maarten Sap, Hannah Rashkin, and Yejin Choi. 2020.
\newblock Power{T}ransformer: {U}nsupervised controllable revision for biased
  language correction.
\newblock In \emph{Proceedings of the 2020 Conference on Empirical Methods in
  Natural Language Processing, {EMNLP} 2020, Online, November 16-20, 2020},
  pages 7426--7441, Association for Computational Linguistics.

\bibitem[{Madaan et~al.(2020)Madaan, Setlur, Parekh, P{\'{o}}czos, Neubig,
  Yang, Salakhutdinov, Black, and Prabhumoye}]{madaan2020politeness}
Madaan, Aman, Amrith Setlur, Tanmay Parekh, Barnab{\'{a}}s P{\'{o}}czos, Graham
  Neubig, Yiming Yang, Ruslan Salakhutdinov, Alan~W. Black, and Shrimai
  Prabhumoye. 2020.
\newblock Politeness transfer: {A} tag and generate approach.
\newblock In \emph{Proceedings of the 58th Annual Meeting of the Association
  for Computational Linguistics, {ACL} 2020, Online, July 5-10, 2020}, pages
  1869--1881, Association for Computational Linguistics.

\bibitem[{Madnani and Dorr(2010)}]{madnani2010generating}
Madnani, Nitin and Bonnie~J. Dorr. 2010.
\newblock Generating phrasal and sentential paraphrases: {A} survey of
  data-driven methods.
\newblock \emph{Compututational Linguistics}, 36(3):341--387.

\bibitem[{Mairesse and Walker(2011)}]{mairesse2011controling}
Mairesse, Fran{\c{c}}ois and Marilyn~A. Walker. 2011.
\newblock Controlling user perceptions of linguistic style: {T}rainable
  generation of personality traits.
\newblock \emph{Computational Linguistics}, 37(3):455--488.

\bibitem[{Malmi, Severyn, and
  Rothe(2020{\natexlab{a}})}]{malmi2020unsupervised}
Malmi, Eric, Aliaksei Severyn, and Sascha Rothe. 2020{\natexlab{a}}.
\newblock Unsupervised text style transfer with padded masked language models.
\newblock In \emph{Proceedings of the 2020 Conference on Empirical Methods in
  Natural Language Processing, {EMNLP} 2020, Online, November 16-20, 2020},
  pages 8671--8680, Association for Computational Linguistics.

\bibitem[{Malmi, Severyn, and
  Rothe(2020{\natexlab{b}})}]{malmi-etal-2020-unsupervised}
Malmi, Eric, Aliaksei Severyn, and Sascha Rothe. 2020{\natexlab{b}}.
\newblock Unsupervised text style transfer with padded masked language models.
\newblock In \emph{Proceedings of the 2020 Conference on Empirical Methods in
  Natural Language Processing (EMNLP)}, pages 8671--8680, Association for
  Computational Linguistics, Online.

\bibitem[{Mani(2001)}]{mani2001automatic}
Mani, Inderjeet. 2001.
\newblock \emph{Automatic summarization}, volume~3.
\newblock John Benjamins Publishing.

\bibitem[{Mansoorizadeh et~al.(2016)Mansoorizadeh, Rahgooy, Aminiyan, and
  Eskandari}]{mansoorizadeh2016author}
Mansoorizadeh, Muharram, Taher Rahgooy, Mohammad Aminiyan, and Mahdy Eskandari.
  2016.
\newblock Author obfuscation using wordnet and language models—notebook for
  pan at clef 2016.
\newblock In \emph{CLEF 2016 Evaluation Labs and Workshop--Working Notes
  Papers}, pages 5--8.

\bibitem[{Marie and Fujita(2017)}]{marie2017efficient}
Marie, Benjamin and Atsushi Fujita. 2017.
\newblock Efficient extraction of pseudo-parallel sentences from raw
  monolingual data using word embeddings.
\newblock In \emph{Proceedings of the 55th Annual Meeting of the Association
  for Computational Linguistics, {ACL} 2017, Vancouver, Canada, July 30 -
  August 4, Volume 2: Short Papers}, pages 392--398, Association for
  Computational Linguistics.

\bibitem[{McDonald and Pustejovsky(1985)}]{mcdonald1985computational}
McDonald, David~D. and James Pustejovsky. 1985.
\newblock A computational theory of prose style for natural language
  generation.
\newblock In \emph{{EACL} 1985, 2nd Conference of the European Chapter of the
  Association for Computational Linguistics, March 27-29, 1985, University of
  Geneva, Geneva, Switzerland}, pages 187--193, The Association for Computer
  Linguistics.

\bibitem[{McTear(2002)}]{mctear2002spoken}
McTear, Michael~F. 2002.
\newblock Spoken dialogue technology: {E}nabling the conversational user
  interface.
\newblock \emph{{ACM} Computing Surveys}, 34(1):90--169.

\bibitem[{Merity et~al.(2017)Merity, Xiong, Bradbury, and
  Socher}]{Merity2017PointerSM}
Merity, Stephen, Caiming Xiong, James Bradbury, and R.~Socher. 2017.
\newblock Pointer sentinel mixture models.
\newblock \emph{ArXiv}, abs/1609.07843.

\bibitem[{Mir et~al.(2019)Mir, Felbo, Obradovich, and
  Rahwan}]{mir-etal-2019-evaluating}
Mir, Remi, Bjarke Felbo, Nick Obradovich, and Iyad Rahwan. 2019.
\newblock Evaluating style transfer for text.
\newblock In \emph{Proceedings of the 2019 Conference of the North {A}merican
  Chapter of the Association for Computational Linguistics: Human Language
  Technologies, Volume 1 (Long and Short Papers)}, pages 495--504, Association
  for Computational Linguistics, Minneapolis, Minnesota.

\bibitem[{Mou and Vechtomova(2020)}]{mou-vechtomova-2020-stylized}
Mou, Lili and Olga Vechtomova. 2020.
\newblock Stylized text generation: {A}pproaches and applications.
\newblock In \emph{Proceedings of the 58th Annual Meeting of the Association
  for Computational Linguistics: Tutorial Abstracts}, pages 19--22, Association
  for Computational Linguistics, Online.

\bibitem[{Mueller, Gifford, and Jaakkola(2017)}]{mueller2017sequence}
Mueller, Jonas, David~K. Gifford, and Tommi~S. Jaakkola. 2017.
\newblock Sequence to better sequence: {C}ontinuous revision of combinatorial
  structures.
\newblock In \emph{Proceedings of the 34th International Conference on Machine
  Learning, {ICML} 2017, Sydney, NSW, Australia, 6-11 August 2017}, volume~70
  of \emph{Proceedings of Machine Learning Research}, pages 2536--2544, {PMLR}.

\bibitem[{Munteanu and Marcu(2005)}]{munteanu2005improving}
Munteanu, Dragos~Stefan and Daniel Marcu. 2005.
\newblock Improving machine translation performance by exploiting non-parallel
  corpora.
\newblock \emph{Computational Linguistics}, 31(4):477--504.

\bibitem[{Nikolov and Hahnloser(2019)}]{nikolov2019large}
Nikolov, Nikola~I. and Richard H.~R. Hahnloser. 2019.
\newblock Large-scale hierarchical alignment for data-driven text rewriting.
\newblock In \emph{Proceedings of the International Conference on Recent
  Advances in Natural Language Processing, {RANLP} 2019, Varna, Bulgaria,
  September 2-4, 2019}, pages 844--853, {INCOMA} Ltd.

\bibitem[{Niu and Bansal(2018)}]{niu-bansal-2018-polite}
Niu, Tong and Mohit Bansal. 2018.
\newblock Polite dialogue generation without parallel data.
\newblock \emph{Transactions of the Association for Computational Linguistics},
  6:373--389.

\bibitem[{Niu, Martindale, and Carpuat(2017)}]{niu2017study}
Niu, Xing, Marianna~J. Martindale, and Marine Carpuat. 2017.
\newblock A study of style in machine translation: {C}ontrolling the formality
  of machine translation output.
\newblock In \emph{Proceedings of the 2017 Conference on Empirical Methods in
  Natural Language Processing, {EMNLP} 2017, Copenhagen, Denmark, September
  9-11, 2017}, pages 2814--2819, Association for Computational Linguistics.

\bibitem[{Niu, Rao, and Carpuat(2018)}]{niu-etal-2018-multi}
Niu, Xing, Sudha Rao, and Marine Carpuat. 2018.
\newblock Multi-task neural models for translating between styles within and
  across languages.
\newblock In \emph{Proceedings of the 27th International Conference on
  Computational Linguistics}, pages 1008--1021, Association for Computational
  Linguistics, Santa Fe, New Mexico, USA.

\bibitem[{Novikova, Dusek, and Rieser(2017)}]{novikova2017e2e}
Novikova, Jekaterina, Ondrej Dusek, and Verena Rieser. 2017.
\newblock The {E2E} dataset: {N}ew challenges for end-to-end generation.
\newblock In \emph{Proceedings of the 18th Annual SIGdial Meeting on Discourse
  and Dialogue, Saarbr{\"{u}}cken, Germany, August 15-17, 2017}, pages
  201--206, Association for Computational Linguistics.

\bibitem[{Oksanen et~al.(2014)Oksanen, Hawdon, Holkeri, N{\"a}si, and
  R{\"a}s{\"a}nen}]{oksanen2014exposure}
Oksanen, Atte, James Hawdon, Emma Holkeri, Matti N{\"a}si, and Pekka
  R{\"a}s{\"a}nen. 2014.
\newblock Exposure to online hate among young social media users.
\newblock In \emph{Soul of society: A focus on the lives of children \& youth}.
  Emerald Group Publishing Limited.

\bibitem[{Pang(2019)}]{pang2020daunting}
Pang, Richard~Yuanzhe. 2019.
\newblock The daunting task of real-world textual style transfer
  auto-evaluation.
\newblock \emph{CoRR}, abs/1910.03747.

\bibitem[{Pang and Gimpel(2019)}]{pang2018unsupervised}
Pang, Richard~Yuanzhe and Kevin Gimpel. 2019.
\newblock Unsupervised evaluation metrics and learning criteria for
  non-parallel textual transfer.
\newblock In \emph{Proceedings of the 3rd Workshop on Neural Generation and
  Translation@EMNLP-IJCNLP 2019, Hong Kong, November 4, 2019}, pages 138--147,
  Association for Computational Linguistics.

\bibitem[{Papineni et~al.(2002)Papineni, Roukos, Ward, and
  Zhu}]{papineni-etal-2002-bleu}
Papineni, Kishore, Salim Roukos, Todd Ward, and Wei-Jing Zhu. 2002.
\newblock {BLEU}: {A} method for automatic evaluation of machine translation.
\newblock In \emph{Proceedings of the 40th Annual Meeting of the Association
  for Computational Linguistics}, pages 311--318, Association for Computational
  Linguistics, Philadelphia, Pennsylvania, USA.

\bibitem[{Parikh et~al.(2020)Parikh, Wang, Gehrmann, Faruqui, Dhingra, Yang,
  and Das}]{parikh2020totto}
Parikh, Ankur~P., Xuezhi Wang, Sebastian Gehrmann, Manaal Faruqui, Bhuwan
  Dhingra, Diyi Yang, and Dipanjan Das. 2020.
\newblock {ToTTo}: {A} controlled table-to-text generation dataset.
\newblock In \emph{Proceedings of the 2020 Conference on Empirical Methods in
  Natural Language Processing, {EMNLP} 2020, Online, November 16-20, 2020},
  pages 1173--1186, Association for Computational Linguistics.

\bibitem[{Pfaff(1979)}]{pfaff1979constraints}
Pfaff, Carol~W. 1979.
\newblock Constraints on language mixing: {I}ntrasentential code-switching and
  borrowing in {S}panish/{E}nglish.
\newblock \emph{Language}, pages 291--318.

\bibitem[{Poplack(2000)}]{poplack2000sometimes}
Poplack, Shana. 2000.
\newblock Sometimes {I}’ll start a sentence in {S}panish y termino en
  espa{\~n}ol: {T}oward a typology of code-switching.
\newblock \emph{The bilingualism reader}, 18(2):221--256.

\bibitem[{Popovi{\'c}(2015)}]{popovic-2015-chrf}
Popovi{\'c}, Maja. 2015.
\newblock chr{F}: {C}haracter n-gram {F}-score for automatic {MT} evaluation.
\newblock In \emph{Proceedings of the Tenth Workshop on Statistical Machine
  Translation}, pages 392--395, Association for Computational Linguistics,
  Lisbon, Portugal.

\bibitem[{Post and Vilar(2018)}]{post-vilar-2018-fast}
Post, Matt and David Vilar. 2018.
\newblock Fast lexically constrained decoding with dynamic beam allocation for
  neural machine translation.
\newblock In \emph{Proceedings of the 2018 Conference of the North {A}merican
  Chapter of the Association for Computational Linguistics: Human Language
  Technologies, Volume 1 (Long Papers)}, pages 1314--1324, Association for
  Computational Linguistics, New Orleans, Louisiana.

\bibitem[{Power, Scott, and Bouayad-Agha(2003)}]{power2003generating}
Power, Richard, Donia Scott, and Nadjet Bouayad-Agha. 2003.
\newblock Generating texts with style.
\newblock In \emph{International Conference on Intelligent Text Processing and
  Computational Linguistics}, pages 444--452, Springer.

\bibitem[{Prabhumoye et~al.(2018)Prabhumoye, Tsvetkov, Salakhutdinov, and
  Black}]{prabhumoye-etal-2018-style}
Prabhumoye, Shrimai, Yulia Tsvetkov, Ruslan Salakhutdinov, and Alan~W Black.
  2018.
\newblock Style transfer through back-translation.
\newblock In \emph{Proceedings of the 56th Annual Meeting of the Association
  for Computational Linguistics (Volume 1: Long Papers)}, pages 866--876,
  Association for Computational Linguistics, Melbourne, Australia.

\bibitem[{Pryzant et~al.(2020)Pryzant, Martinez, Dass, Kurohashi, Jurafsky, and
  Yang}]{pryzant2020automatically}
Pryzant, Reid, Richard~Diehl Martinez, Nathan Dass, Sadao Kurohashi, Dan
  Jurafsky, and Diyi Yang. 2020.
\newblock Automatically neutralizing subjective bias in text.
\newblock In \emph{The Thirty-Fourth {AAAI} Conference on Artificial
  Intelligence, {AAAI} 2020, The Thirty-Second Innovative Applications of
  Artificial Intelligence Conference, {IAAI} 2020, The Tenth {AAAI} Symposium
  on Educational Advances in Artificial Intelligence, {EAAI} 2020, New York,
  NY, USA, February 7-12, 2020}, pages 480--489, {AAAI} Press.

\bibitem[{Qian et~al.(2019)Qian, Zhang, Chang, Yang, and
  Hasegawa{-}Johnson}]{qian2019autovc}
Qian, Kaizhi, Yang Zhang, Shiyu Chang, Xuesong Yang, and Mark
  Hasegawa{-}Johnson. 2019.
\newblock Auto{VC}: {Z}ero-shot voice style transfer with only autoencoder
  loss.
\newblock In \emph{Proceedings of the 36th International Conference on Machine
  Learning, {ICML} 2019, 9-15 June 2019, Long Beach, California, {USA}},
  volume~97 of \emph{Proceedings of Machine Learning Research}, pages
  5210--5219, {PMLR}.

\bibitem[{Qin and Eisner(2021)}]{qin2021learning}
Qin, Guanghui and Jason Eisner. 2021.
\newblock Learning how to ask: {Q}uerying {LM}s with mixtures of soft prompts.
\newblock In \emph{Proceedings of the 2021 Conference of the North American
  Chapter of the Association for Computational Linguistics: Human Language
  Technologies, {NAACL-HLT} 2021, Online, June 6-11, 2021}, pages 5203--5212,
  Association for Computational Linguistics.

\bibitem[{Qin et~al.(2019)Qin, Bosselut, Holtzman, Bhagavatula, Clark, and
  Choi}]{qin2019counterfactual}
Qin, Lianhui, Antoine Bosselut, Ari Holtzman, Chandra Bhagavatula, Elizabeth
  Clark, and Yejin Choi. 2019.
\newblock Counterfactual story reasoning and generation.
\newblock In \emph{Proceedings of the 2019 Conference on Empirical Methods in
  Natural Language Processing and the 9th International Joint Conference on
  Natural Language Processing, {EMNLP-IJCNLP} 2019, Hong Kong, China, November
  3-7, 2019}, pages 5042--5052, Association for Computational Linguistics.

\bibitem[{Radford et~al.(2019)Radford, Wu, Child, Luan, Amodei, and
  Sutskever}]{radford2019language}
Radford, Alec, Jeffrey Wu, Rewon Child, David Luan, Dario Amodei, and Ilya
  Sutskever. 2019.
\newblock Language models are unsupervised multitask learners.
\newblock \emph{OpenAI Blog}, 1(8):9.

\bibitem[{Raffel et~al.(2020)Raffel, Shazeer, Roberts, Lee, Narang, Matena,
  Zhou, Li, and Liu}]{raffel2020exploring}
Raffel, Colin, Noam Shazeer, Adam Roberts, Katherine Lee, Sharan Narang,
  Michael Matena, Yanqi Zhou, Wei Li, and Peter~J. Liu. 2020.
\newblock Exploring the limits of transfer learning with a unified text-to-text
  transformer.
\newblock \emph{Journal of Machine Learning Research}, 21:140:1--140:67.

\bibitem[{Rao and Tetreault(2018)}]{rao-tetreault-2018-dear}
Rao, Sudha and Joel Tetreault. 2018.
\newblock Dear sir or madam, may {I} introduce the {GYAFC} dataset: {C}orpus,
  benchmarks and metrics for formality style transfer.
\newblock In \emph{Proceedings of the 2018 Conference of the North {A}merican
  Chapter of the Association for Computational Linguistics: Human Language
  Technologies, Volume 1 (Long Papers)}, pages 129--140, Association for
  Computational Linguistics, New Orleans, Louisiana.

\bibitem[{Reddy and Knight(2016)}]{reddy2016obfuscating}
Reddy, Sravana and Kevin Knight. 2016.
\newblock Obfuscating gender in social media writing.
\newblock In \emph{Proceedings of the First Workshop on {NLP} and Computational
  Social Science, NLP+CSS@EMNLP 2016, Austin, TX, USA, November 5, 2016}, pages
  17--26, Association for Computational Linguistics.

\bibitem[{Reiter and Dale(1997)}]{DBLP:journals/nle/ReiterD97}
Reiter, Ehud and Robert Dale. 1997.
\newblock Building applied natural language generation systems.
\newblock \emph{Natural Language Engineering}, 3(1):57--87.

\bibitem[{Reiter, Robertson, and Osman(2003)}]{reiter2003lessons}
Reiter, Ehud, Roma Robertson, and Liesl~M Osman. 2003.
\newblock Lessons from a failure: {G}enerating tailored smoking cessation
  letters.
\newblock \emph{Artificial Intelligence}, 144(1-2):41--58.

\bibitem[{Reiter et~al.(2005)Reiter, Sripada, Hunter, Yu, and
  Davy}]{reiter2005choosing}
Reiter, Ehud, Somayajulu Sripada, Jim Hunter, Jin Yu, and Ian Davy. 2005.
\newblock Choosing words in computer-generated weather forecasts.
\newblock \emph{Artificial Intelligence}, 167(1-2):137--169.

\bibitem[{Ren et~al.(2020)Ren, Wu, Liu, Zhou, and Ma}]{ren2020retrieve}
Ren, Shuo, Yu~Wu, Shujie Liu, Ming Zhou, and Shuai Ma. 2020.
\newblock A retrieve-and-rewrite initialization method for unsupervised machine
  translation.
\newblock In \emph{Proceedings of the 58th Annual Meeting of the Association
  for Computational Linguistics, {ACL} 2020, Online, July 5-10, 2020}, pages
  3498--3504, Association for Computational Linguistics.

\bibitem[{Rezende, Mohamed, and Wierstra(2014)}]{Rezende2014vae}
Rezende, Danilo~Jimenez, Shakir Mohamed, and Daan Wierstra. 2014.
\newblock Stochastic backpropagation and approximate inference in deep
  generative models.
\newblock In \emph{Proceedings of the 31th International Conference on Machine
  Learning, {ICML} 2014, Beijing, China, 21-26 June 2014}, volume~32 of
  \emph{{JMLR} Workshop and Conference Proceedings}, pages 1278--1286,
  JMLR.org.

\bibitem[{Ribeiro et~al.(2020)Ribeiro, Schmitt, Sch{\"{u}}tze, and
  Gurevych}]{ribeiro2020investigating}
Ribeiro, Leonardo F.~R., Martin Schmitt, Hinrich Sch{\"{u}}tze, and Iryna
  Gurevych. 2020.
\newblock Investigating pretrained language models for graph-to-text
  generation.
\newblock \emph{CoRR}, abs/2007.08426.

\bibitem[{Riley et~al.(2021{\natexlab{a}})Riley, Constant, Guo, Kumar, Uthus,
  and Parekh}]{riley2021textsettr}
Riley, Parker, Noah Constant, Mandy Guo, Girish Kumar, David Uthus, and Zarana
  Parekh. 2021{\natexlab{a}}.
\newblock {T}ext{SETTR}: {F}ew-shot text style extraction and tunable targeted
  restyling.
\newblock In \emph{Proceedings of the 59th Annual Meeting of the Association
  for Computational Linguistics and the 11th International Joint Conference on
  Natural Language Processing (Volume 1: Long Papers)}, pages 3786--3800,
  Association for Computational Linguistics, Online.

\bibitem[{Riley et~al.(2021{\natexlab{b}})Riley, Constant, Guo, Kumar, Uthus,
  and Parekh}]{Riley2021TextSETTRFT}
Riley, Parker, Noah Constant, Mandy Guo, Girish Kumar, David~C. Uthus, and
  Zarana Parekh. 2021{\natexlab{b}}.
\newblock Text{S}ettr: {F}ew-shot text style extraction and tunable targeted
  restyling.
\newblock In \emph{ACL/IJCNLP}.

\bibitem[{Roller et~al.(2021)Roller, Dinan, Goyal, Ju, Williamson, Liu, Xu,
  Ott, Smith, Boureau, and Weston}]{roller2020recipes}
Roller, Stephen, Emily Dinan, Naman Goyal, Da~Ju, Mary Williamson, Yinhan Liu,
  Jing Xu, Myle Ott, Eric~Michael Smith, Y{-}Lan Boureau, and Jason Weston.
  2021.
\newblock Recipes for building an open-domain chatbot.
\newblock In \emph{Proceedings of the 16th Conference of the European Chapter
  of the Association for Computational Linguistics: Main Volume, {EACL} 2021,
  Online, April 19 - 23, 2021}, pages 300--325, Association for Computational
  Linguistics.

\bibitem[{Romanov et~al.(2019)Romanov, Rumshisky, Rogers, and
  Donahue}]{romanov-etal-2019-adversarial}
Romanov, Alexey, Anna Rumshisky, Anna Rogers, and David Donahue. 2019.
\newblock Adversarial decomposition of text representation.
\newblock In \emph{Proceedings of the 2019 Conference of the North {A}merican
  Chapter of the Association for Computational Linguistics: Human Language
  Technologies, Volume 1 (Long and Short Papers)}, pages 815--825, Association
  for Computational Linguistics, Minneapolis, Minnesota.

\bibitem[{Ruder, Dosovitskiy, and Brox(2016)}]{ruder2016artistic}
Ruder, Manuel, Alexey Dosovitskiy, and Thomas Brox. 2016.
\newblock Artistic style transfer for videos.
\newblock In \emph{Pattern Recognition - 38th German Conference, {GCPR} 2016,
  Hannover, Germany, September 12-15, 2016, Proceedings}, volume 9796 of
  \emph{Lecture Notes in Computer Science}, pages 26--36, Springer.

\bibitem[{Rush, Chopra, and Weston(2015)}]{rush2015neural}
Rush, Alexander~M., Sumit Chopra, and Jason Weston. 2015.
\newblock A neural attention model for abstractive sentence summarization.
\newblock In \emph{Proceedings of the 2015 Conference on Empirical Methods in
  Natural Language Processing, {EMNLP} 2015, Lisbon, Portugal, September 17-21,
  2015}, pages 379--389, The Association for Computational Linguistics.

\bibitem[{Russell, Dewey, and Tegmark(2015)}]{russell2015research}
Russell, Stuart~J., Daniel Dewey, and Max Tegmark. 2015.
\newblock Research priorities for robust and beneficial artificial
  intelligence.
\newblock \emph{AI Magazine}, 36(4):105--114.

\bibitem[{Sancheti et~al.(2020)Sancheti, Krishna, Srinivasan, and
  Natarajan}]{sancheti2020reinforced}
Sancheti, Abhilasha, Kundan Krishna, Balaji~Vasan Srinivasan, and Anandhavelu
  Natarajan. 2020.
\newblock Reinforced rewards framework for text style transfer.
\newblock In \emph{European Conference on Information Retrieval}, pages
  545--560, Springer.

\bibitem[{dos Santos, Melnyk, and
  Padhi(2018)}]{nogueira-dos-santos-etal-2018-fighting}
dos Santos, C{\'{\i}}cero~Nogueira, Igor Melnyk, and Inkit Padhi. 2018.
\newblock Fighting offensive language on social media with unsupervised text
  style transfer.
\newblock In \emph{Proceedings of the 56th Annual Meeting of the Association
  for Computational Linguistics, {ACL} 2018, Melbourne, Australia, July 15-20,
  2018, Volume 2: Short Papers}, pages 189--194, Association for Computational
  Linguistics.

\bibitem[{Scao and Rush(2021)}]{scao2021how}
Scao, Teven~Le and Alexander~M. Rush. 2021.
\newblock How many data points is a prompt worth?
\newblock In \emph{Proceedings of the 2021 Conference of the North American
  Chapter of the Association for Computational Linguistics: Human Language
  Technologies, {NAACL-HLT} 2021, Online, June 6-11, 2021}, pages 2627--2636,
  Association for Computational Linguistics.

\bibitem[{Schler et~al.(2006)Schler, Koppel, Argamon, and
  Pennebaker}]{schler2006effects}
Schler, Jonathan, Moshe Koppel, Shlomo Argamon, and James~W. Pennebaker. 2006.
\newblock Effects of age and gender on blogging.
\newblock In \emph{Computational Approaches to Analyzing Weblogs, Papers from
  the 2006 {AAAI} Spring Symposium, Technical Report SS-06-03, Stanford,
  California, USA, March 27-29, 2006}, pages 199--205, {AAAI}.

\bibitem[{See, Liu, and Manning(2017)}]{see2017get}
See, Abigail, Peter~J. Liu, and Christopher~D. Manning. 2017.
\newblock Get to the point: {S}ummarization with pointer-generator networks.
\newblock In \emph{Proceedings of the 55th Annual Meeting of the Association
  for Computational Linguistics, {ACL} 2017, Vancouver, Canada, July 30 -
  August 4, Volume 1: Long Papers}, pages 1073--1083, Association for
  Computational Linguistics.

\bibitem[{Sennrich, Haddow, and
  Birch(2016{\natexlab{a}})}]{sennrich2016controlling}
Sennrich, Rico, Barry Haddow, and Alexandra Birch. 2016{\natexlab{a}}.
\newblock Controlling politeness in neural machine translation via side
  constraints.
\newblock In \emph{{NAACL} {HLT} 2016, The 2016 Conference of the North
  American Chapter of the Association for Computational Linguistics: Human
  Language Technologies, San Diego California, USA, June 12-17, 2016}, pages
  35--40, The Association for Computational Linguistics.

\bibitem[{Sennrich, Haddow, and
  Birch(2016{\natexlab{b}})}]{sennrich-etal-2016-edinburgh}
Sennrich, Rico, Barry Haddow, and Alexandra Birch. 2016{\natexlab{b}}.
\newblock {E}dinburgh neural machine translation systems for {WMT} 16.
\newblock In \emph{Proceedings of the First Conference on Machine Translation:
  Volume 2, Shared Task Papers}, pages 371--376, Association for Computational
  Linguistics, Berlin, Germany.

\bibitem[{Shang et~al.(2019)Shang, Li, Fu, Bing, Zhao, Shi, and
  Yan}]{shang-etal-2019-semi}
Shang, Mingyue, Piji Li, Zhenxin Fu, Lidong Bing, Dongyan Zhao, Shuming Shi,
  and Rui Yan. 2019.
\newblock Semi-supervised text style transfer: {C}ross projection in latent
  space.
\newblock In \emph{Proceedings of the 2019 Conference on Empirical Methods in
  Natural Language Processing and the 9th International Joint Conference on
  Natural Language Processing (EMNLP-IJCNLP)}, pages 4937--4946, Association
  for Computational Linguistics, Hong Kong, China.

\bibitem[{Shao et~al.(2018)Shao, Ciampaglia, Varol, Yang, Flammini, and
  Menczer}]{shao2018spread}
Shao, Chengcheng, Giovanni~Luca Ciampaglia, Onur Varol, Kai-Cheng Yang,
  Alessandro Flammini, and Filippo Menczer. 2018.
\newblock The spread of low-credibility content by social bots.
\newblock \emph{Nature communications}, 9(1):1--9.

\bibitem[{Sharma et~al.(2021)Sharma, Lin, Miner, Atkins, and
  Althoff}]{sharma2021towards}
Sharma, Ashish, Inna~W. Lin, Adam~S. Miner, David~C. Atkins, and Tim Althoff.
  2021.
\newblock Towards facilitating empathic conversations in online mental health
  support: {A} reinforcement learning approach.
\newblock \emph{CoRR}, abs/2101.07714.

\bibitem[{Sheikha and Inkpen(2011)}]{sheikha2011formal}
Sheikha, Fadi~Abu and Diana Inkpen. 2011.
\newblock Generation of formal and informal sentences.
\newblock In \emph{{ENLG} 2011 - Proceedings of the 13th European Workshop on
  Natural Language Generation, 28-30 September 2011, Nancy, France}, pages
  187--193, The Association for Computer Linguistics.

\bibitem[{Shen et~al.(2017)Shen, Lei, Barzilay, and Jaakkola}]{shen2017style}
Shen, Tianxiao, Tao Lei, Regina Barzilay, and Tommi Jaakkola. 2017.
\newblock Style transfer from non-parallel text by cross-alignment.
\newblock In \emph{Advances in neural information processing systems}, pages
  6830--6841.

\bibitem[{Shetty and Adibi(2004)}]{shetty2004enron}
Shetty, Jitesh and Jafar Adibi. 2004.
\newblock The enron email dataset database schema and brief statistical report.
\newblock \emph{Information sciences institute technical report, University of
  Southern California}, 4(1):120--128.

\bibitem[{Shetty, Schiele, and Fritz(2018)}]{shetty2018author}
Shetty, Rakshith, Bernt Schiele, and Mario Fritz. 2018.
\newblock {A4NT:} author attribute anonymity by adversarial training of neural
  machine translation.
\newblock In \emph{27th {USENIX} Security Symposium, {USENIX} Security 2018,
  Baltimore, MD, USA, August 15-17, 2018}, pages 1633--1650, {USENIX}
  Association.

\bibitem[{Shuster et~al.(2020)Shuster, Humeau, Bordes, and
  Weston}]{shuster2020image}
Shuster, Kurt, Samuel Humeau, Antoine Bordes, and Jason Weston. 2020.
\newblock Image-{C}hat: {E}ngaging grounded conversations.
\newblock In \emph{Proceedings of the 58th Annual Meeting of the Association
  for Computational Linguistics, {ACL} 2020, Online, July 5-10, 2020}, pages
  2414--2429, Association for Computational Linguistics.

\bibitem[{Song et~al.(2019)Song, Tan, Qin, Lu, and Liu}]{song2019mass}
Song, Kaitao, Xu~Tan, Tao Qin, Jianfeng Lu, and Tie{-}Yan Liu. 2019.
\newblock {MASS:} masked sequence to sequence pre-training for language
  generation.
\newblock In \emph{Proceedings of the 36th International Conference on Machine
  Learning, {ICML} 2019, 9-15 June 2019, Long Beach, California, {USA}},
  volume~97 of \emph{Proceedings of Machine Learning Research}, pages
  5926--5936, {PMLR}.

\bibitem[{Sripada et~al.(2004)Sripada, Reiter, Davy, and
  Nilssen}]{sripada2004lessons}
Sripada, Somayajulu, Ehud Reiter, Ian Davy, and Kristian Nilssen. 2004.
\newblock Lessons from deploying {NLG} technology for marine weather forecast
  text generation.
\newblock In \emph{Proceedings of the 16th Eureopean Conference on Artificial
  Intelligence, ECAI'2004, including Prestigious Applicants of Intelligent
  Systems, {PAIS} 2004, Valencia, Spain, August 22-27, 2004}, pages 760--764,
  {IOS} Press.

\bibitem[{Stamatatos et~al.(1997)Stamatatos, Michos, Fakotakis, and
  Kokkinakis}]{stamatatos1997user}
Stamatatos, Efstathios, S~Michos, Nikos Fakotakis, and George Kokkinakis. 1997.
\newblock A user-assisted business letter generator dealing with text's
  stylistic variations.
\newblock In \emph{Proceedings Ninth IEEE International Conference on Tools
  with Artificial Intelligence}, pages 182--189, IEEE.

\bibitem[{Starr(2019)}]{starr2019counterfactuals}
Starr, William. 2019.
\newblock Counterfactuals.
\newblock \emph{The Stanford Encyclopedia of Philosophy}.

\bibitem[{Sudhakar, Upadhyay, and
  Maheswaran(2019)}]{Sudhakar2019TransformingDR}
Sudhakar, Akhilesh, Bhargav Upadhyay, and Arjun Maheswaran. 2019.
\newblock "transforming" delete, retrieve, generate approach for controlled
  text style transfer.
\newblock In \emph{Proceedings of the 2019 Conference on Empirical Methods in
  Natural Language Processing and the 9th International Joint Conference on
  Natural Language Processing, {EMNLP-IJCNLP} 2019, Hong Kong, China, November
  3-7, 2019}, pages 3267--3277, Association for Computational Linguistics.

\bibitem[{Sutskever, Vinyals, and Le(2014)}]{sutskever2014sequence}
Sutskever, Ilya, Oriol Vinyals, and Quoc~V Le. 2014.
\newblock Sequence to sequence learning with neural networks.
\newblock In \emph{Advances in neural information processing systems}, pages
  3104--3112.

\bibitem[{Syed et~al.(2020)Syed, Verma, Srinivasan, Natarajan, and
  Varma}]{Syed2020AdaptingLM}
Syed, Bakhtiyar, Gaurav Verma, Balaji~Vasan Srinivasan, Anandhavelu Natarajan,
  and Vasudeva Varma. 2020.
\newblock Adapting language models for non-parallel author-stylized rewriting.
\newblock In \emph{AAAI}.

\bibitem[{Tan and Goonawardene(2017)}]{tan2017internet}
Tan, Sharon Swee-Lin and Nadee Goonawardene. 2017.
\newblock Internet health information seeking and the patient-physician
  relationship: {A} systematic review.
\newblock \emph{Journal of medical Internet research}, 19(1):e9.

\bibitem[{Tannen(1990)}]{tannen1990gender}
Tannen, Deborah. 1990.
\newblock Gender differences in topical coherence: {C}reating involvement in
  best friends' talk.
\newblock \emph{Discourse Processes}, 13(1):73--90.

\bibitem[{Tian, Hu, and Yu(2018)}]{tian2018structured}
Tian, Youzhi, Zhiting Hu, and Zhou Yu. 2018.
\newblock Structured content preservation for unsupervised text style transfer.
\newblock \emph{CoRR}, abs/1810.06526.

\bibitem[{Tikhonov et~al.(2019)Tikhonov, Shibaev, Nagaev, Nugmanova, and
  Yamshchikov}]{tikhonov-etal-2019-style}
Tikhonov, Alexey, Viacheslav Shibaev, Aleksander Nagaev, Aigul Nugmanova, and
  Ivan~P. Yamshchikov. 2019.
\newblock Style transfer for texts: {R}etrain, report errors, compare with
  rewrites.
\newblock In \emph{Proceedings of the 2019 Conference on Empirical Methods in
  Natural Language Processing and the 9th International Joint Conference on
  Natural Language Processing (EMNLP-IJCNLP)}, pages 3936--3945, Association
  for Computational Linguistics, Hong Kong, China.

\bibitem[{Tikhonov and Yamshchikov(2018)}]{tikhonov2018wrong}
Tikhonov, Alexey and Ivan~P. Yamshchikov. 2018.
\newblock What is wrong with style transfer for texts?
\newblock \emph{CoRR}, abs/1808.04365.

\bibitem[{Tran, Zhang, and Soleymani(2020)}]{tran2020towards}
Tran, Minh, Yipeng Zhang, and Mohammad Soleymani. 2020.
\newblock Towards a friendly online community: {A}n unsupervised style transfer
  framework for profanity redaction.
\newblock \emph{CoRR}, abs/2011.00403.

\bibitem[{Trudgill(1972)}]{trudgill1972sex}
Trudgill, Peter. 1972.
\newblock Sex, covert prestige and linguistic change in the urban british
  english of norwich.
\newblock \emph{Language in society}, 1(2):179--195.

\bibitem[{Tse et~al.(2015)Tse, Schrader, Ghosh, Liao, and
  Lundie}]{tse2015bibliometric}
Tse, Jonathan, Dawn~E. Schrader, Dipayan~P. Ghosh, Tony~C. Liao, and David
  Lundie. 2015.
\newblock A bibliometric analysis of privacy and ethics in \emph{IEEE Security
  and Privacy}.
\newblock \emph{Ethics and Information Technology}, 17(2):153--163.

\bibitem[{Uszkoreit et~al.(2010)Uszkoreit, Ponte, Popat, and
  Dubiner}]{uszkoreit2010large}
Uszkoreit, Jakob, Jay Ponte, Ashok~C. Popat, and Moshe Dubiner. 2010.
\newblock Large scale parallel document mining for machine translation.
\newblock In \emph{{COLING} 2010, 23rd International Conference on
  Computational Linguistics, Proceedings of the Conference, 23-27 August 2010,
  Beijing, China}, pages 1101--1109, Tsinghua University Press.

\bibitem[{Vaswani et~al.(2017)Vaswani, Shazeer, Parmar, Uszkoreit, Jones,
  Gomez, Kaiser, and Polosukhin}]{vaswani2017attention}
Vaswani, Ashish, Noam Shazeer, Niki Parmar, Jakob Uszkoreit, Llion Jones,
  Aidan~N. Gomez, Lukasz Kaiser, and Illia Polosukhin. 2017.
\newblock Attention is all you need.
\newblock In \emph{Advances in Neural Information Processing Systems 30: Annual
  Conference on Neural Information Processing Systems 2017, December 4-9, 2017,
  Long Beach, CA, {USA}}, pages 5998--6008.

\bibitem[{Vincent et~al.(2010)Vincent, Larochelle, Lajoie, Bengio, and
  Manzagol}]{vincent2010stacked}
Vincent, Pascal, Hugo Larochelle, Isabelle Lajoie, Yoshua Bengio, and
  Pierre{-}Antoine Manzagol. 2010.
\newblock Stacked denoising autoencoders: {L}earning useful representations in
  a deep network with a local denoising criterion.
\newblock \emph{Journal of Machine Learning Research}, 11:3371--3408.

\bibitem[{Voigt et~al.(2018)Voigt, Jurgens, Prabhakaran, Jurafsky, and
  Tsvetkov}]{voigt2018rtgender}
Voigt, Rob, David Jurgens, Vinodkumar Prabhakaran, Dan Jurafsky, and Yulia
  Tsvetkov. 2018.
\newblock {RtGender}: {A} corpus for studying differential responses to gender.
\newblock In \emph{Proceedings of the Eleventh International Conference on
  Language Resources and Evaluation, {LREC} 2018, Miyazaki, Japan, May 7-12,
  2018}, European Language Resources Association {(ELRA)}.

\bibitem[{Wang, Quan, and Wang(2019)}]{wang2019biset}
Wang, Kai, Xiaojun Quan, and Rui Wang. 2019.
\newblock {BiSET: B}i-directional selective encoding with template for
  abstractive summarization.
\newblock In \emph{Proceedings of the 57th Conference of the Association for
  Computational Linguistics, {ACL} 2019, Florence, Italy, July 28- August 2,
  2019, Volume 1: Long Papers}, pages 2153--2162, Association for Computational
  Linguistics.

\bibitem[{Wang, Hua, and Wan(2019)}]{wang2019controllable}
Wang, Ke, Hang Hua, and Xiaojun Wan. 2019.
\newblock Controllable unsupervised text attribute transfer via editing
  entangled latent representation.
\newblock In \emph{Advances in Neural Information Processing Systems 32: Annual
  Conference on Neural Information Processing Systems 2019, NeurIPS 2019,
  December 8-14, 2019, Vancouver, BC, Canada}, pages 11034--11044.

\bibitem[{Wang et~al.(2019)Wang, Wu, Mou, Li, and
  Chao}]{wang-etal-2019-harnessing}
Wang, Yunli, Yu~Wu, Lili Mou, Zhoujun Li, and Wenhan Chao. 2019.
\newblock Harnessing pre-trained neural networks with rules for formality style
  transfer.
\newblock In \emph{Proceedings of the 2019 Conference on Empirical Methods in
  Natural Language Processing and the 9th International Joint Conference on
  Natural Language Processing (EMNLP-IJCNLP)}, pages 3573--3578, Association
  for Computational Linguistics, Hong Kong, China.

\bibitem[{Waseem et~al.(2017)Waseem, Davidson, Warmsley, and
  Weber}]{waseem2017understanding}
Waseem, Zeerak, Thomas Davidson, Dana Warmsley, and Ingmar Weber. 2017.
\newblock Understanding abuse: {A} typology of abusive language detection
  subtasks.
\newblock In \emph{Proceedings of the First Workshop on Abusive Language
  Online, ALW@ACL 2017, Vancouver, BC, Canada, August 4, 2017}, pages 78--84,
  Association for Computational Linguistics.

\bibitem[{Weng, Chung, and Szolovits(2019)}]{weng2019unsupervised}
Weng, Wei{-}Hung, Yu{-}An Chung, and Peter Szolovits. 2019.
\newblock Unsupervised clinical language translation.
\newblock In \emph{Proceedings of the 25th {ACM} {SIGKDD} International
  Conference on Knowledge Discovery {\&} Data Mining, {KDD} 2019, Anchorage,
  AK, USA, August 4-8, 2019}, pages 3121--3131, {ACM}.

\bibitem[{Weston, Dinan, and Miller(2018)}]{weston2018retrieve}
Weston, Jason, Emily Dinan, and Alexander~H. Miller. 2018.
\newblock Retrieve and refine: {I}mproved sequence generation models for
  dialogue.
\newblock In \emph{Proceedings of the 2nd International Workshop on
  Search-Oriented Conversational AI, SCAI@EMNLP 2018, Brussels, Belgium,
  October 31, 2018}, pages 87--92, Association for Computational Linguistics.

\bibitem[{Williams(1992)}]{williams1992simple}
Williams, Ronald~J. 1992.
\newblock Simple statistical gradient-following algorithms for connectionist
  reinforcement learning.
\newblock \emph{Machine learning}, 8(3-4):229--256.

\bibitem[{Wiseman, Shieber, and Rush(2017)}]{wiseman2017challenges}
Wiseman, Sam, Stuart~M. Shieber, and Alexander~M. Rush. 2017.
\newblock Challenges in data-to-document generation.
\newblock In \emph{Proceedings of the 2017 Conference on Empirical Methods in
  Natural Language Processing, {EMNLP} 2017, Copenhagen, Denmark, September
  9-11, 2017}, pages 2253--2263, Association for Computational Linguistics.

\bibitem[{Wu, Wang, and Wang(2019)}]{wu2019extract}
Wu, Jiawei, Xin Wang, and William~Yang Wang. 2019.
\newblock Extract and edit: {A}n alternative to back-translation for
  unsupervised neural machine translation.
\newblock In \emph{Proceedings of the 2019 Conference of the North American
  Chapter of the Association for Computational Linguistics: Human Language
  Technologies, {NAACL-HLT} 2019, Minneapolis, MN, USA, June 2-7, 2019, Volume
  1 (Long and Short Papers)}, pages 1173--1183, Association for Computational
  Linguistics.

\bibitem[{Wu et~al.(2019)Wu, Zhang, Zang, Han, and Hu}]{wu2019mask}
Wu, Xing, Tao Zhang, Liangjun Zang, Jizhong Han, and Songlin Hu. 2019.
\newblock "mask and infill" : {A}pplying masked language model to sentiment
  transfer.
\newblock \emph{CoRR}, abs/1908.08039.

\bibitem[{Wu, Wang, and Liu(2020)}]{Wu2020ADF}
Wu, Yu, Yunli Wang, and Shujie Liu. 2020.
\newblock A dataset for low-resource stylized sequence-to-sequence generation.
\newblock In \emph{The Thirty-Fourth {AAAI} Conference on Artificial
  Intelligence, {AAAI} 2020, The Thirty-Second Innovative Applications of
  Artificial Intelligence Conference, {IAAI} 2020, The Tenth {AAAI} Symposium
  on Educational Advances in Artificial Intelligence, {EAAI} 2020, New York,
  NY, USA, February 7-12, 2020}, pages 9290--9297, {AAAI} Press.

\bibitem[{Xing et~al.(2020)Xing, Jin, Jin, Wang, Zhang, and
  Huang}]{xing2020tasty}
Xing, Xiaoyu, Zhijing Jin, Di~Jin, Bingning Wang, Qi~Zhang, and Xuanjing Huang.
  2020.
\newblock Tasty burgers, soggy fries: {P}robing aspect robustness in
  aspect-based sentiment analysis.
\newblock In \emph{Proceedings of the 2020 Conference on Empirical Methods in
  Natural Language Processing, {EMNLP} 2020, Online, November 16-20, 2020},
  pages 3594--3605, Association for Computational Linguistics.

\bibitem[{Xu et~al.(2018)Xu, Sun, Zeng, Zhang, Ren, Wang, and
  Li}]{xu-etal-2018-unpaired}
Xu, Jingjing, Xu~Sun, Qi~Zeng, Xiaodong Zhang, Xuancheng Ren, Houfeng Wang, and
  Wenjie Li. 2018.
\newblock Unpaired sentiment-to-sentiment translation: {A} cycled reinforcement
  learning approach.
\newblock In \emph{Proceedings of the 56th Annual Meeting of the Association
  for Computational Linguistics (Volume 1: Long Papers)}, pages 979--988,
  Association for Computational Linguistics, Melbourne, Australia.

\bibitem[{Xu, Cheung, and Cao(2020)}]{xu2020variational}
Xu, Peng, Jackie Chi~Kit Cheung, and Yanshuai Cao. 2020.
\newblock On variational learning of controllable representations for text
  without supervision.
\newblock In \emph{Proceedings of the 37th International Conference on Machine
  Learning, {ICML} 2020, 13-18 July 2020, Virtual Event}, volume 119 of
  \emph{Proceedings of Machine Learning Research}, pages 10534--10543, {PMLR}.

\bibitem[{Xu et~al.(2020)Xu, Tao, Cheng, Tan, and Zhang}]{xu2020towardsfeature}
Xu, Qiuling, Guanhong Tao, Siyuan Cheng, Lin Tan, and Xiangyu Zhang. 2020.
\newblock Towards feature space adversarial attack.
\newblock \emph{CoRR}, abs/2004.12385.

\bibitem[{Xu, Ge, and Wei(2019)}]{xu2019formality}
Xu, Ruochen, Tao Ge, and Furu Wei. 2019.
\newblock Formality style transfer with hybrid textual annotations.
\newblock \emph{CoRR}, abs/1903.06353.

\bibitem[{Xu et~al.(2012)Xu, Ritter, Dolan, Grishman, and
  Cherry}]{xu2012paraphrasing}
Xu, Wei, Alan Ritter, Bill Dolan, Ralph Grishman, and Colin Cherry. 2012.
\newblock Paraphrasing for style.
\newblock In \emph{{COLING} 2012, 24th International Conference on
  Computational Linguistics, Proceedings of the Conference: Technical Papers,
  8-15 December 2012, Mumbai, India}, pages 2899--2914, Indian Institute of
  Technology Bombay.

\bibitem[{Yamshchikov et~al.(2021)Yamshchikov, Shibaev, Khlebnikov, and
  Tikhonov}]{yamshchikov2020style}
Yamshchikov, Ivan~P., Viacheslav Shibaev, Nikolay Khlebnikov, and Alexey
  Tikhonov. 2021.
\newblock Style-transfer and paraphrase: {L}ooking for a sensible semantic
  similarity metric.
\newblock In \emph{Thirty-Fifth {AAAI} Conference on Artificial Intelligence,
  {AAAI} 2021, Thirty-Third Conference on Innovative Applications of Artificial
  Intelligence, {IAAI} 2021, The Eleventh Symposium on Educational Advances in
  Artificial Intelligence, {EAAI} 2021, Virtual Event, February 2-9, 2021},
  pages 14213--14220, {AAAI} Press.

\bibitem[{Yamshchikov et~al.(2019)Yamshchikov, Shibaev, Nagaev, Jost, and
  Tikhonov}]{yamshchikov-etal-2019-decomposing}
Yamshchikov, Ivan~P., Viacheslav Shibaev, Aleksander Nagaev, J{\"u}rgen Jost,
  and Alexey Tikhonov. 2019.
\newblock Decomposing textual information for style transfer.
\newblock In \emph{Proceedings of the 3rd Workshop on Neural Generation and
  Translation}, pages 128--137, Association for Computational Linguistics, Hong
  Kong.

\bibitem[{Yang et~al.(2018)Yang, Hu, Dyer, Xing, and
  Berg{-}Kirkpatrick}]{yang2018unsupervised}
Yang, Zichao, Zhiting Hu, Chris Dyer, Eric~P. Xing, and Taylor
  Berg{-}Kirkpatrick. 2018.
\newblock Unsupervised text style transfer using language models as
  discriminators.
\newblock In \emph{Advances in Neural Information Processing Systems 31: Annual
  Conference on Neural Information Processing Systems 2018, NeurIPS 2018, 3-8
  December 2018, Montr{\'{e}}al, Canada}, pages 7298--7309.

\bibitem[{Yi et~al.(2020)Yi, Liu, Li, and Sun}]{Yi2020TextST}
Yi, Xiaoyuan, Zhenghao Liu, Wenhao Li, and Maosong Sun. 2020.
\newblock Text style transfer via learning style instance supported latent
  space.
\newblock In \emph{Proceedings of the Twenty-Ninth International Joint
  Conference on Artificial Intelligence, {IJCAI} 2020}, pages 3801--3807,
  ijcai.org.

\bibitem[{Yuan et~al.(2021)Yuan, Cheng, Zhang, Hao, Gan, and
  Carin}]{yuan2021improving}
Yuan, Siyang, Pengyu Cheng, Ruiyi Zhang, Weituo Hao, Zhe Gan, and Lawrence
  Carin. 2021.
\newblock Improving zero-shot voice style transfer via disentangled
  representation learning.
\newblock \emph{CoRR}, abs/2103.09420.

\bibitem[{Zeng, Shoeybi, and Liu(2020)}]{zeng2020style}
Zeng, Kuo{-}Hao, Mohammad Shoeybi, and Ming{-}Yu Liu. 2020.
\newblock Style example-guided text generation using generative adversarial
  transformers.
\newblock \emph{CoRR}, abs/2003.00674.

\bibitem[{Zhang et~al.(2018{\natexlab{a}})Zhang, Utiyama, Sumita, Neubig, and
  Nakamura}]{zhang2018guiding}
Zhang, Jingyi, Masao Utiyama, Eiichiro Sumita, Graham Neubig, and Satoshi
  Nakamura. 2018{\natexlab{a}}.
\newblock Guiding neural machine translation with retrieved translation pieces.
\newblock In \emph{Proceedings of the 2018 Conference of the North American
  Chapter of the Association for Computational Linguistics: Human Language
  Technologies, {NAACL-HLT} 2018, New Orleans, Louisiana, USA, June 1-6, 2018,
  Volume 1 (Long Papers)}, pages 1325--1335, Association for Computational
  Linguistics.

\bibitem[{Zhang et~al.(2018{\natexlab{b}})Zhang, Dinan, Urbanek, Szlam, Kiela,
  and Weston}]{zhang2018personalizing}
Zhang, Saizheng, Emily Dinan, Jack Urbanek, Arthur Szlam, Douwe Kiela, and
  Jason Weston. 2018{\natexlab{b}}.
\newblock Personalizing dialogue agents: {I} have a dog, do you have pets too?
\newblock In \emph{Proceedings of the 56th Annual Meeting of the Association
  for Computational Linguistics, {ACL} 2018, Melbourne, Australia, July 15-20,
  2018, Volume 1: Long Papers}, pages 2204--2213, Association for Computational
  Linguistics.

\bibitem[{Zhang et~al.(2020)Zhang, Kishore, Wu, Weinberger, and
  Artzi}]{Zhang2020BERTScoreET}
Zhang, Tianyi, Varsha Kishore, Felix Wu, Kilian~Q. Weinberger, and Yoav Artzi.
  2020.
\newblock {BERTScore: E}valuating text generation with {BERT}.
\newblock In \emph{8th International Conference on Learning Representations,
  {ICLR} 2020, Addis Ababa, Ethiopia, April 26-30, 2020}, OpenReview.net.

\bibitem[{Zhang and Liu(2013)}]{zhang2013writer}
Zhang, Xu{-}Yao and Cheng{-}Lin Liu. 2013.
\newblock Writer adaptation with style transfer mapping.
\newblock \emph{IEEE Transactions on Pattern Analysis and Machine
  Intelligence}, 35(7):1773--1787.

\bibitem[{Zhang, Ge, and Sun(2020)}]{zhang-etal-2020-parallel}
Zhang, Yi, Tao Ge, and Xu~Sun. 2020.
\newblock Parallel data augmentation for formality style transfer.
\newblock In \emph{Proceedings of the 58th Annual Meeting of the Association
  for Computational Linguistics}, pages 3221--3228, Association for
  Computational Linguistics, Online.

\bibitem[{Zhang et~al.(2018{\natexlab{c}})Zhang, Xu, Yang, and
  Sun}]{zhang2018learning}
Zhang, Yi, Jingjing Xu, Pengcheng Yang, and Xu~Sun. 2018{\natexlab{c}}.
\newblock Learning sentiment memories for sentiment modification without
  parallel data.
\newblock In \emph{Proceedings of the 2018 Conference on Empirical Methods in
  Natural Language Processing, Brussels, Belgium, October 31 - November 4,
  2018}, pages 1103--1108, Association for Computational Linguistics.

\bibitem[{Zhang et~al.(2018{\natexlab{d}})Zhang, Ren, Liu, Wang, Chen, Li,
  Zhou, and Chen}]{zhang2018style}
Zhang, Zhirui, Shuo Ren, Shujie Liu, Jianyong Wang, Peng Chen, Mu~Li, Ming
  Zhou, and Enhong Chen. 2018{\natexlab{d}}.
\newblock Style transfer as unsupervised machine translation.
\newblock \emph{CoRR}, abs/1808.07894.

\bibitem[{Zhao et~al.(2018)Zhao, Kim, Zhang, Rush, and
  LeCun}]{Zhao2018AdversariallyRA}
Zhao, Junbo~Jake, Yoon Kim, Kelly Zhang, Alexander~M. Rush, and Yann LeCun.
  2018.
\newblock Adversarially regularized autoencoders.
\newblock In \emph{Proceedings of the 35th International Conference on Machine
  Learning, {ICML} 2018, Stockholmsm{\"{a}}ssan, Stockholm, Sweden, July 10-15,
  2018}, volume~80 of \emph{Proceedings of Machine Learning Research}, pages
  5897--5906, {PMLR}.

\bibitem[{Zheng et~al.(2019)Zheng, Chalasani, Ghosal, Lutz, and
  Smolic}]{zheng2019stada}
Zheng, Xu, Tejo Chalasani, Koustav Ghosal, Sebastian Lutz, and Aljosa Smolic.
  2019.
\newblock {STaDA: S}tyle transfer as data augmentation.
\newblock In \emph{Proceedings of the 14th International Joint Conference on
  Computer Vision, Imaging and Computer Graphics Theory and Applications,
  {VISIGRAPP} 2019, Volume 4: VISAPP, Prague, Czech Republic, February 25-27,
  2019}, pages 107--114, SciTePress.

\bibitem[{Zhu et~al.(2017)Zhu, Park, Isola, and Efros}]{zhu2017unpaired}
Zhu, Jun{-}Yan, Taesung Park, Phillip Isola, and Alexei~A. Efros. 2017.
\newblock Unpaired image-to-image translation using cycle-consistent
  adversarial networks.
\newblock In \emph{{IEEE} International Conference on Computer Vision, {ICCV}
  2017, Venice, Italy, October 22-29, 2017}, pages 2242--2251, {IEEE} Computer
  Society.

\bibitem[{Zhu, Bernhard, and Gurevych(2010)}]{zhu2010monolingual}
Zhu, Zhemin, Delphine Bernhard, and Iryna Gurevych. 2010.
\newblock A monolingual tree-based translation model for sentence
  simplification.
\newblock In \emph{{COLING} 2010, 23rd International Conference on
  Computational Linguistics, Proceedings of the Conference, 23-27 August 2010,
  Beijing, China}, pages 1353--1361, Tsinghua University Press.

\bibitem[{Zue and Glass(2000)}]{zue2000conversational}
Zue, Victor~W and James~R Glass. 2000.
\newblock Conversational interfaces: {A}dvances and challenges.
\newblock \emph{Proceedings of the IEEE}, 88(8):1166--1180.

\end{thebibliography}

\end{document}